%% file: iclr2026_conference.tex
\documentclass{article} 
\usepackage{iclr2026_conference, times}

\input{math_commands.tex}

\usepackage{hyperref}
\usepackage{url}
\usepackage{enumitem}
\usepackage{multirow}
\usepackage[ruled, vlined]{algorithm2e}
\usepackage{graphicx}
\usepackage{booktabs}
\usepackage{xcolor}
\usepackage[breakable]{tcolorbox}
\tcbuselibrary{skins}
\usepackage{listings}
\usepackage{wrapfig}
\usepackage{textcomp}
\usepackage[strings]{underscore}

\newcommand{\mname}{Reasoning BO}

\title{Reasoning BO: Enhancing Bayesian Optimization with the
  Long-Context Reasoning
Power of LLMs}
\author{
  Zhuo Yang\textsuperscript{1,2} \quad
  Daolang Wang\textsuperscript{3} \quad
  Lingli Ge\textsuperscript{1,4} \quad
  Beilun Wang\textsuperscript{5} \quad
  Tianfan Fu\textsuperscript{6,1} \quad
  Yuqiang Li\textsuperscript{1} \\ \\
  \textsuperscript{1} Shanghai Artificial Intelligence Laboratory \quad
  \textsuperscript{2} Xidian University \quad
  \textsuperscript{3} North University of China \\
  \textsuperscript{4} Fudan University \quad
  \textsuperscript{5} Southeast University \quad
  \textsuperscript{6} Nanjing University \\
}

\iclrfinalcopy 

\begin{document}
\maketitle

\begin{abstract}
  Many real-world scientific and industrial applications require the
  optimization
  of expensive black-box functions. Bayesian Optimization (BO) provides an
  effective framework for such problems.
  However, traditional BO methods are prone to get trapped in local optima
  and often lack interpretable insights.
  To address this issue, this paper designs {\mname}, a novel framework that
  leverages reasoning models to guide the sampling process in BO
  while incorporating
  multi-agent systems and knowledge graphs for online knowledge accumulation.
  We systematically evaluate our approach across 10 diverse tasks encompassing
  synthetic mathematical functions and complex real-world applications.
  The framework demonstrates its capability to progressively refine
  sampling strategies through real-time insights and hypothesis evolution,
  effectively identifying higher-performing regions of the search space for
  focused exploration. This process highlights the powerful reasoning and
  context-learning abilities of LLMs in optimization scenarios. For example,
  in the Direct Arylation task (a chemical reaction yield optimization problem),
  our method increased the yield to 60.7\%, whereas traditional BO achieved
  only a 25.2\% yield. Furthermore, our investigation reveals that smaller
  LLMs, after post-training, can attain comparable performance to
  their larger counterparts.
\end{abstract}

\section{Introduction}
Real-world scientific optimization problems often involve complex,
expensive-to-evaluate
functions across diverse domains, from chemical synthesis
\citep{Startt2024-cx, Liang2024-mk,DBLP:journals/corr/abs-2503-05607}
to hyperparameter tuning
\citep{DBLP:conf/nips/SnoekLA12,DBLP:conf/aaai/FeurerSH15,Onorato2024}.
Bayesian Optimization provides an efficient framework for such black-box
problems by constructing probabilistic surrogate models and intelligently
guiding experiments through acquisition functions~\citep{7352306}. While BO
outperforms traditional methods in sample efficiency, it faces key
limitations: susceptibility to local optima~\citep{DBLP:conf/ijcai/CisseECGC24}, sensitivity to initial sampling~\citep{Huang202212}, and lack of interpretable scientific insights~\citep{González-Duque2024}.
These challenges motivate our integration of reasoning capabilities with BO.

Recent research has explored integrating LLMs to create hybrid intelligent
optimization frameworks that overcome BO's inherent
limitations. These frameworks leverage three key LLM capabilities: First, their
cross-domain knowledge provides global heuristics to avoid
local optima~\citep{DBLP:journals/corr/abs-2501-16224}. Second, domain
priors like chemical reaction rules can be naturally injected through
language to reduce sensitivity to initial sampling. Most importantly, LLMs
generate scientifically plausible mechanistic hypotheses and iteratively
refine them through experiments \citep{DBLP:conf/ijcai/CisseECGC24}. However,
current LLM-BO integration approaches suffer from two critical limitations:
(1) potentially hallucinated optimization suggestions from LLMs that
may violate domain
constraints and compromise safety, and (2) inability to dynamically incorporate
experimental discoveries into subsequent optimization cycles. These shortcomings
necessitate the development of memory-augmented closed-loop
optimization systems.

To address these challenges, we present the \mname\ framework (illustrated in
Figure~\ref{fig:flowchart}), which incorporates three mechanisms for
trustworthy optimization. First, we embed a Reasoning Model within
the BO loop: (a) users describe experiments in natural language via
the Experiment Compass to define the search space; (b) the BO
algorithm proposes candidate points, which are then evaluated by the
LLM—leveraging domain priors, historical data, and knowledge
graphs—to generate scientific hypotheses and assign confidence scores
to each candidate; (c) candidates are filtered based on confidence
and consistency with prior results to ensure scientific plausibility.
Second, we construct a dynamic knowledge management system that
integrates structured domain rules in knowledge graphs and
unstructured literature in vector databases, enabling both expert
knowledge injection and real-time assimilation of new findings.
Third, we fine-tune lightweight models using reinforcement learning
on reasoning trajectories to further enhance optimization performance.

For ease of exposition, we summarize our main contributions as
\begin{itemize}[leftmargin=5pt]
  \item Incorporation of a reasoning model that leverages LLMs'
    inference abilities to automatically generate and evolve scientific
    hypotheses, ensuring scientific plausibility via confidence-based
    filtering.

  \item Development of a full-cycle knowledge management system
    enabling dynamic updating of domain expertise throughout optimization,
    featuring flexible knowledge architectures with customizable storage
    formats and open agent interfaces for extensibility.

  \item Demonstrated superior performance across diverse optimization tasks,
    particularly in chemical yield prediction and high-dimensional benchmarks.
    In our Direct Arylation benchmark, Reasoning BO achieves a 23.3\% higher
    final yield (94.39\% vs 76.60\%) and 44.6\% higher initial performance
    (66.08\% vs 21.62\%) compared to Vanilla BO, showcasing both superior
    initialization and continuous optimization capabilities.
\end{itemize}

\section{Related Work}

\subsection{Bayesian Optimization}
Bayesian Optimization (BO) is a widely used framework for black-box
function optimization, which relies on probabilistic surrogate models
that are iteratively refined using historical
observations~\citep{Rasmussen2004}. Typically, BO consists of two
main components: a surrogate model—most commonly a Gaussian Process
(GP) that models the objective as $f(\mathbf{x}) \sim
\mathcal{N}(\mu(\mathbf{x}),
\sigma^{2}(\mathbf{x}))$~\citep{Rasmussen2004}—and an acquisition
function that guides the search by balancing exploration and
exploitation. Among various acquisition strategies, Expected
Improvement (EI) is particularly popular due to its solid theoretical
foundation~\citep{Mockus1978, zhan_expected_2020}. Despite its theoretical
sample efficiency, BO faces notable challenges in practical,
high-cost evaluation scenarios. First, the computational complexity
of surrogate model training increases rapidly with the number of
evaluations~\citep{LAN2022100970}. Second, standard acquisition
functions often struggle to dynamically adjust the
exploration-exploitation trade-off, which can result in convergence
to sub-optimal local optima~\citep{DBLP:conf/nips/ErikssonPGTP19}. To
address these limitations, our approach leverages the global
heuristic perspective and reasoning capabilities of large language
models (LLMs), which can partially mitigate the
bottlenecks and enhance the overall efficiency.

\subsection{Language-Guided Bayesian Optimization}

By leveraging LLMs' contextual
understanding\citep{DBLP:journals/corr/abs-2206-07682,
DBLP:conf/emnlp/Dong0DZMLXX0C0S24}
and few-shot generation capabilities \citep{DBLP:conf/nips/BrownMRSKDNSSAA20},
recent advances demonstrate their potential to effectively address black-box
optimization challenges when integrated with BO \citep{DBLP:conf/iclr/LiuASS24}.
Current mainstream work in this area can be primarily categorized into two
types: (1) LLM-assisted experimental design, which incorporates reasoning
capabilities into the development of candidates sampling strategies, and (2)
knowledge transfer techniques, which leverage domain-specific insights or
user belief to guide the optimization process. In the first category,
SLLMBO \citep{DBLP:journals/corr/abs-2410-20302} employs a hybrid
sampling strategy
combining LLMs with Tree-structured Parzen Estimator (TPE)
\citep{DBLP:journals/eswa/ShakyaDSA24}
for efficient hyperparameter optimization. HypBO
\citep{DBLP:conf/ijcai/CisseECGC24}
and BORA \citep{DBLP:journals/corr/abs-2501-16224} further accelerate Bayesian
searches by leveraging expert hypotheses to rapidly identify promising regions.
In the second category, MI-SMBO \citep{DBLP:conf/aaai/FeurerSH15}
initializes Bayesian Optimization using configurations suggested by
meta-learning,
while BOChemian \citep{rankovic2023bochemian} transforms textual chemical
procedures into an informative feature space suitable for optimization.

While current approaches have made significant progress, they still encounter
challenges such as potential hallucinations in LLM-generated suggestions that
could affect optimization accuracy
\citep{DBLP:journals/corr/abs-2401-11817,DBLP:journals/csur/JiLFYSXIBMF23},
over-reliance on manual prompt templates limiting adaptability, and
insufficient mechanisms for persisting knowledge. Additionally, these
methods may face issues related to data leakage during benchmark testing. To
address these issues, our framework integrates an evolvable domain knowledge
graph for continuous learning \citep{DBLP:conf/acl/ZhaoLJQB23}, a multi-agent
system to enhance adaptability, and RAG mechanisms to improve knowledge reuse
efficiency and correct factual errors\citep{DBLP:conf/emnlp/0001PCKW21}.

\subsection{Reinforcement Learning-Enhanced Reasoning in LLMs}

Recent advances in reinforcement learning have significantly enhanced
the reasoning capabilities of large language models through policy
optimization techniques. Policy gradient methods
\citep{DBLP:journals/ijon/XuXZL24, DBLP:journals/corr/abs-2108-02665}
and Proximal Policy Optimization (PPO)
\citep{DBLP:journals/corr/SchulmanWDRK17} form the theoretical
foundation for training LLMs to generate high-quality reasoning
trajectories. Building upon these approaches, Group Relative Policy
Optimization (GRPO) \citep{DBLP:journals/corr/abs-2402-03300} extends
PPO through group relative advantage estimation, enabling more stable
policy updates in complex reasoning scenarios. The application of RL
to reasoning has shown promise across diverse domains, including
rule-based reasoning \citep{xie2025logicrlunleashingllmreasoning,
tinyzero, zeng2025simplerlzooinvestigatingtamingzero}, visual
reasoning with out-of-distribution generalization
\citep{chen2025r1v}, and medical reasoning through verified question
answering \citep{DBLP:journals/corr/abs-2502-19655}. These works
demonstrate that Reinforcement Learning from Human Feedback (RLHF)
\citep{DBLP:conf/nips/Ouyang0JAWMZASR22} can effectively optimize
models to produce more reliable and structured reasoning outputs,
providing the foundation for integrating reasoning capabilities into
optimization frameworks.

\begin{figure}[h]
  \centering
  \includegraphics[width=\linewidth]{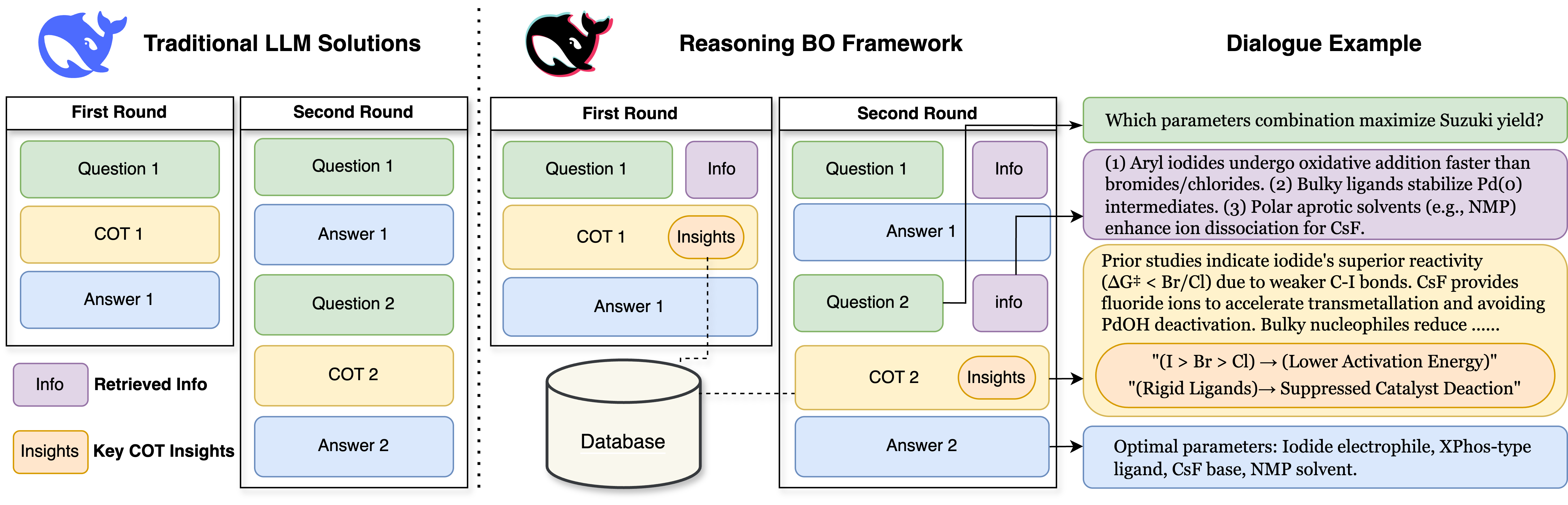}
  \caption{Comparison of classic methods versus our algorithm in utilizing
    Chain-of-Thought (CoT) data. ``Info'' represents database
    information retrieved
    based on question keywords, ``Insights'' represents critical insights
  extracted from Chain-of-Thought data analysis.}
  \label{fig:multi-round}
\end{figure}

\section{Methodology}

\subsection{Motivation}

The integration of LLMs with traditional BO
addresses three fundamental limitations: (1) ineffective utilization
of domain-specific prior knowledge, (2) lack of interpretability in
mathematical optimization, and (3) weak cross-domain adaptability~\citep{González-Duque_2024}.
Classical BO defines search spaces as rigid parametric constructs,
targeting either single-objective optima~\citep{feliot_bayesian_2017}
or Pareto fronts~\citep{DBLP:conf/iclr/NavonSFC21,
DBLP:conf/icdm/RuchteG21,letham2018constrainedbayesianoptimizationnoisy} in multi-objective scenarios. Such
formulations, even with extensions like Constrained BayesianOptimization~\citep{gelbart2014bayesianoptimizationunknownconstraints,DBLP:journals/access/AminiVM25,10.5555/3044805.3044997}, are inadequate for domains like chemistry where constraints are implicit and difficult to formulate as explicit conditions~\citep{DBLP:journals/corr/abs-2203-17241}. Moreover, as
illustrated in Figure~\ref{fig:multi-round}, traditional
approaches do not retain CoT data within the context window. In
subsequent rounds of interaction, CoT data from previous rounds is
not referenced, a design choice intended to prevent lengthy reasoning
processes from consuming valuable context space and introducing
noise. However, in experimental sciences, such reasoning trails may
contain critical insights. To address this, our framework extracts
structured notes from each round of CoT data and stores them. During
subsequent queries to the language model, relevant
information is retrieved from the database using keywords extracted
from the query. This mechanism ensures more effective utilization of
CoT data compared to conventional methods.

\begin{figure}
  \centering
  \includegraphics[width=\linewidth]{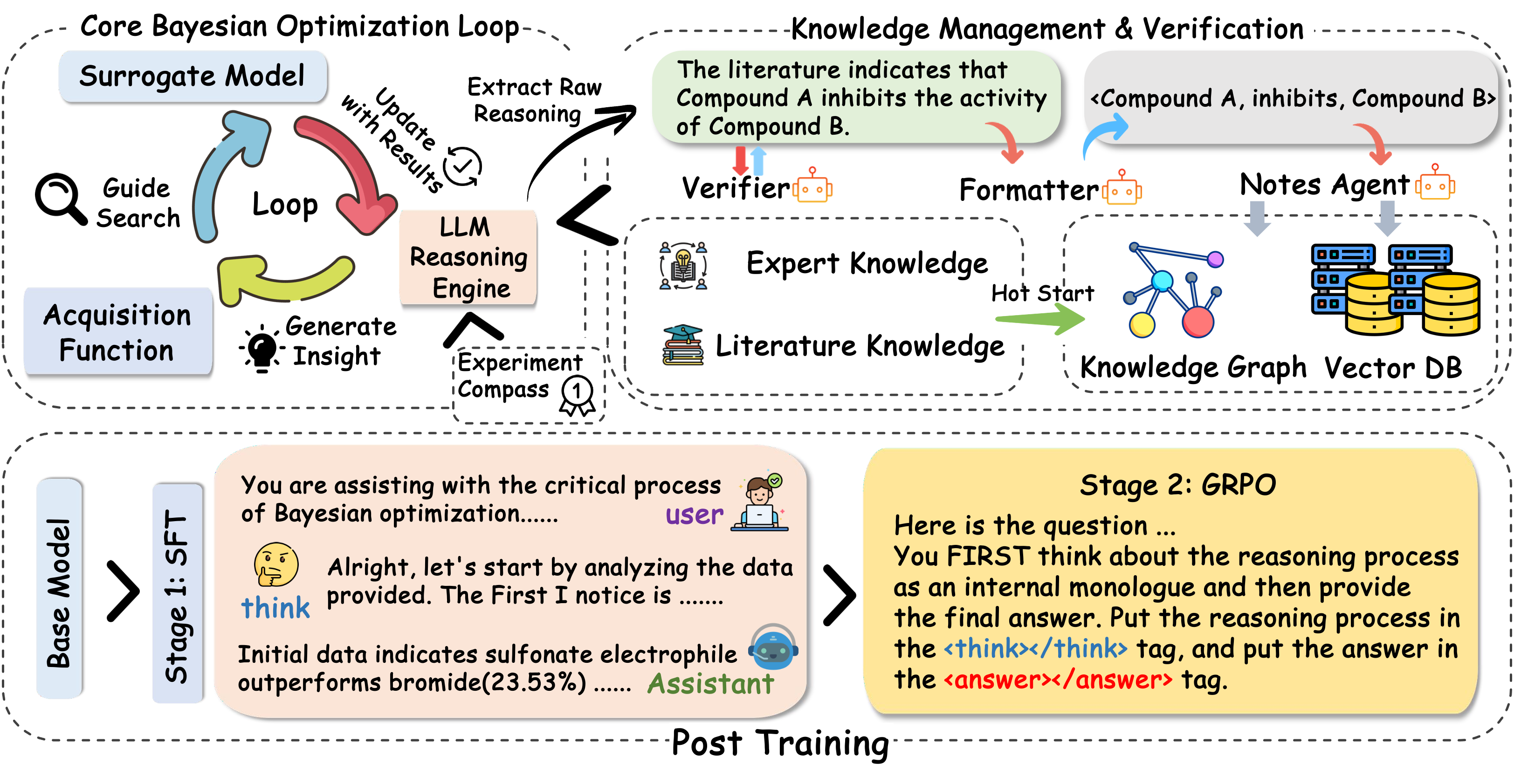}
  \caption{Overview of \mname 's algorithmic framework showcasing the
    interaction
  between Bayesian optimization, reasoning models, and knowledge graphs.}
  \label{fig:flowchart}
\end{figure}

\mname\ addresses these challenges through three core technical
components illustrated in Figure~\ref{fig:flowchart}: (1)
reasoning-enhanced BO that incorporates natural language
specifications and domain knowledge, (2) multi-agent knowledge
management for dynamic information extraction and storage, and (3)
post-training strategies for model enhancement. This framework
operates as an end-to-end system where users provide experimental
objectives through natural language, and the system delivers
optimization results with interpretable insights.

\subsection{Reasoning-Enhanced Bayesian Optimization}
\label{sec:reasoning_enhanced_BO}

Our \mname\ framework integrates reasoning models with BO
through a knowledge-guided optimization process. First, users provide an
Experiment Compass that leverages the language model's semantic
understanding to initialize the search space. The model then
generates an experimental
overview and warm-starts the initial data points. During each optimization
round, a dual-channel knowledge management system (vector database and
knowledge graph) dynamically updates domain knowledge.

The \mname\ algorithm (Algorithm~\ref{alg:reasoning-bo}) implements
reasoning-enhanced
Bayesian optimization through three key phases: (1) Initialization where the
LLM generates initial samples and constructs the Gaussian Process surrogate
model; (2) Iterative optimization rounds where BO first proposes five
candidate points,
then the LLM analyzes these candidates by generating hypotheses based
on historical
insights and experimental data, assigns confidence scores to each hypothesis,
and selects the three candidates with highest confidence scores; (3)
Continuous updates to both the knowledge
base and surrogate model parameters, ensuring the optimization balances
domain knowledge with data-driven principles.

Our sampling strategy avoids direct generation of candidate points by the
LLM to prevent test data contamination from its pretraining corpus and
potential hallucinations that may introduce noise. Instead, we adopt a BO-guided
approach where the optimization process directs the LLM's hypothesis
generation and reasoning. This design
leverages the LLM's strengths in knowledge integration and interpretable
reasoning while delegating numerical optimization to the Bayesian framework. {\textbf{To better illustate \mname\ , we provide a complete
    end-to-end case study in Appendix~\ref{sec:case_study}, using the Buchwald-Hartwig benchmark~\citep{shields_bayesian_2021}.}

\begin{algorithm}
  [htbp]
  \caption{\mname}
  \label{alg:reasoning-bo} \LinesNumbered

  \KwIn{Experiment Compass $C$, Number of samples $n$ in one trial,
  Maximum number of iterations $i_{\text{max}}$}
  \KwOut{All Results $R$, Insight History $H$, and final Summary and Conclusion}

  Initialize experiment with Experiment Compass $C$, set maximum iteration
  count $i_{\text{max}}$; Initialize historical Insights $H = \{\}$,
  Results $R = \{\}$, and Candidate Points $P = \{\}$\;

  Generate experiment overview $O$ using LLM based on $C$, extract key
  entities $E$ from $O$ and store them into vector database $V$ and
  knowledge graph $G$\;

  Reasoning LLM generates initial Reasoning Data $D_{0}$ and Insight
  $I_{0}$ based on $O$ and $C$\; Select high-confidence candidate points $P
  _{0}\subseteq I_{0}$\;

  \For{$t = 1$ \KwTo $i_{\text{max}}$}{
    BO acquisition function proposes 5 candidate points $C_{t}$\;

    Construct input for Reasoning Model by aggregating $H$, $C$, $R$,
    and $C_{t}$\;
    Extract keywords from previous insight $I_{t-1}$ and query $V$
    and $G$ for relevant knowledge\;

    Reasoning Model analyzes each candidate in $C_{t}$, generates
    hypotheses and assigns confidence scores,
    outputs new Reasoning Data $D_{t}$ and Insight $I_{t}$\;

    Select $n=3$ candidates $P_{t} \subseteq C_{t}$ with highest
    confidence scores from $I_{t}$\;

    Conduct BO experiments to obtain results $R_{t}$ using selected
    candidates $P_{t}$\;

    Update $R \leftarrow R \cup R_{t}$, $P \leftarrow P \cup P_{t}$,
    and $H \leftarrow H \cup I_{t}$\;

    Extract notes $N_{t}$ from $D_{t}$ and store them into $V$ and $G$\;
  }

  Generate Conclusion and Summary based on $R$ and $H$\;
\end{algorithm}

\subsection{Multi Agents System for Dynamic
Knowledge Precipitation}
Our dynamic knowledge precipitation mechanism uses a multi-agent
framework for efficient extraction, verification, and integration of
knowledge. Verifier agent extracts variable relationships and
domain knowledge from LLM-generated reasoning data, identifying key
entities, their interactions, and causal dependencies within the
reasoning trajectories. Formatter agent then structures the verified
knowledge for storage by parsing reasoning trajectories into verified
\texttt{<entity, relation, entity>} triples, ensuring consistency and
compatibility with both structured storage formats.
Finally, Notes Agent integrates this structured knowledge into the
vector database and knowledge graph,
where triples are incrementally incorporated into the knowledge base.
Additionally, the system incorporates expert and prior knowledge to
enhance domain-specific comprehensiveness. Verified knowledge is
systematically stored in a dual-channel knowledge management system:
vector databases store high-dimensional semantic features for
efficient similarity searches, while knowledge graphs maintain
interpretable networks of conceptual relationships. During
optimization, the system triggers knowledge retrieval through keyword
queries: first traversing the knowledge graph for structured knowledge
paths, then querying the vector database for contextual information.


\subsection{Post Training Strategies}

To enhance the reasoning capabilities of smaller language models, we implement
a two-stage post-training strategy followed by a knowledge integration phase:

\textbf{1. Cold Start with Supervised Fine-Tuning:} In this stage, we leverage
Low-Rank Adaptation (LoRA) \citep{DBLP:conf/iclr/HuSWALWWC22} to fine-tune
the model using domain-specific reasoning data (technical details in
Appendix~\ref{sec:lora_details}). Our training dataset is constructed from
reasoning trajectories collected during optimization experiments, structured
in a three-component format: \texttt{<user>} contains the optimization prompt
and experimental context, \texttt{<think>} includes the internal reasoning
process generated by reasoning models, and \texttt{<assistant>} provides the
final structured response with candidate recommendations.

\textbf{2. Reinforcement Learning from Human Feedback (RLHF):} We employ Group
Relative Policy Optimization (GRPO) using the GSM8K mathematical
reasoning dataset
to develop robust reasoning patterns. GSM8K consists of 8.5K grade school math
word problems that require multi-step reasoning, where each problem contains a
question and a step-by-step solution demonstrating the reasoning process.
Interestingly, although GSM8K focuses on mathematical reasoning
rather than scientific optimization, we find that training on this
dataset effectively activates the multi-step inference capabilities
essential for optimization tasks. Our
experiments demonstrate
that mathematical reasoning training can transfer to scientific
optimization domains,
enabling smaller models to develop the structured thinking patterns
necessary for
effective Bayesian optimization guidance.

Beyond these two stages, during the knowledge integration phase, the
model dynamically
integrates information from the knowledge graph and vector database (established
in Sections 3.2-3.3). Each reasoning step triggers a dual retrieval process:
first querying the knowledge graph for structured relationships, then
searching the vector space for contextual patterns.

\section{Experiments}

\subsection{Experimental Setup}
\label{sec:experiment_setup}

\textbf{Real-world BO benchmarks}. We evaluate three real-world BO
benchmarks in our main experiments:
(1) \textbf{Suzuki-Miyaura Reaction:} Maximizing product yield by optimizing
electrophile-nucleophile pairs, ligand, base, and solvent under
catalyst compatibility
constraints \citep{perera_platform_2018}; (2) \textbf{Direct Arylation
Reaction:} Maximizing isolated yield through optimal selection of ligand, base,
solvent, concentration and temperature
\citep{mcnally_discovery_2011}; (3) \textbf{Lunar
Lander:} Optimizing control policies for spacecraft landing with mixed
continuous-discrete variables, balancing fuel efficiency and landing precision
under physical constraints \citep{DBLP:conf/nips/MausKEPCG24}.

\textbf{Synthetic Function Benchmarks}. Following
\citep{DBLP:journals/corr/abs-2501-16224,
DBLP:journals/corr/abs-2410-20302, DBLP:journals/corr/abs-2311-14645},
we use synthetic functions to systematically evaluate optimization
algorithms' ability to handle different landscape characteristics: (1)
\textbf{Ackley:} A d-dimensional multimodal function featuring nearly flat outer
regions and a steep central valley. (2) \textbf{Rosenbrock:} A
non-convex d-dimensional
function with narrow parabolic valley structure; (3) \textbf{Hartmann:} A highly
multimodal function with numerous local minima separated by high
energy barriers,
exhibiting strong parameter coupling effects. To ensure that LLMs
cannot identify the optimal solutions by recognizing the
benchmark names, we replaced the function names in the experiment compass
with generic terms such as ``mathematical function''. See
Appendix~\ref{sec:benchmarks} for
complete benchmark specifications.

\textbf{Implementation Details of \mname}. We implement our \mname\ algorithm using QWQ-Plus \citep{qwq32b}, a medium-sized reasoning model from the Qwen series that achieves competitive performance against state-of-the-art reasoning models like DeepSeek-R1. For less complex tasks, such as note extraction and verification, we employ the computationally efficient Qwen2.5-Instruct-14B \citep{DBLP:journals/corr/abs-2501-15383} to reduce overhead. As detailed in Appendix~\ref{sec:api_models}, the capabilities of reasoning models can vary significantly. For BO
acquisition function, we adopt qLogEI, a numerically stable batch
variant that is well-suited
for high-dimensional optimization
tasks~\citep{DBLP:conf/nips/AmentDEBB23} (see Appendix
Section~\ref{sec:logei_family} for details).

\textbf{Baselines}. (1) \textbf{Random Search}: Unbiased exploration baseline.
(2) \textbf{Vanilla BO}: Example of vanilla BO using qLogEI acquisition
function~\citep{DBLP:conf/nips/AmentDEBB23}. (3) \textbf{Analytic EI}: BO
algorithm using Analytic LogEI acquisition function
\citep{frazier_value_2011, frazier_knowledge-gradient_2009}.
(4) \textbf{CMA-ES}: Covariance Matrix Adaptation Evolution Strategy,
a derivative-free evolutionary optimization method
\citep{DBLP:journals/corr/Hansen16a}. Detailed parameter
configurations are provided in Appendix~\ref{sec:baseline_configs}.

\textbf{Experimental Protocol}. Optimization performance is evaluated
based on the best observed objective value. Following the protocol
established in BORA~\citep{DBLP:journals/corr/abs-2501-16224}, we
initialize each experiment and evaluate 3 candidate points in each subsequent round. To balance real-world budget
constraints, the maximum number of experiments is set to 30. Each experiment
is repeated 10 times with fixed random seeds, and the average results are
reported. The source code is publicly available at
\url{https://anonymous.4open.science/r/Reasoning-BO/}.

\noindent
\textbf{Evaluation Metrics.} We employ multiple quantitative metrics to
comprehensively assess optimization performance, as summarized in
Table~\ref{tab:merged_table}. The Coefficient of Variation (CV)
measures relative dispersion,
while standard deviation (Std) quantifies absolute variability. Log
AUC evaluates
cumulative performance by integrating the optimization trajectory. Log
Regret captures the cumulative deviation from optimal values. IMP@$\alpha$
assesses initial performance over the first $\alpha$ iterations, and
CVaR@n evaluates
tail-end performance at confidence level n. These metrics collectively
provide insights into optimization stability, convergence speed, and
robustness across different phases of the search process.

\subsection{Results}
\label{sec:results}

\begin{figure}[h!]
  \centering
  \includegraphics[width=\textwidth]{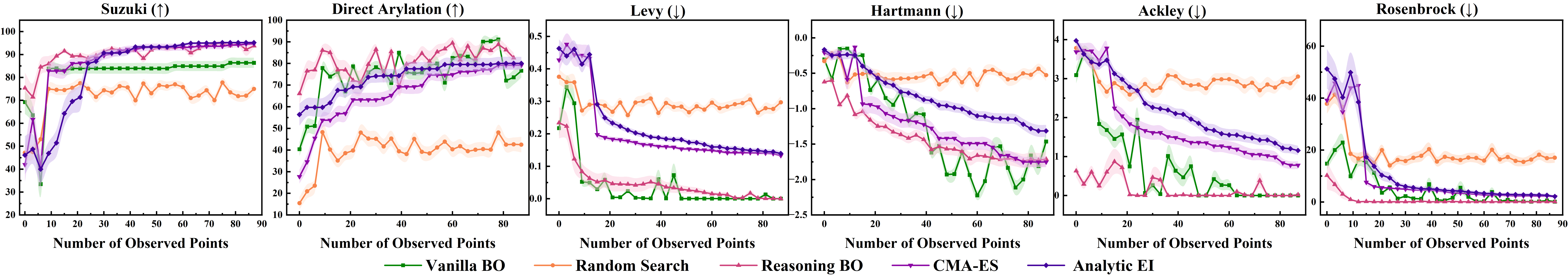}
  \caption{
    Reasoning BO vs Baselines on six experiments. The y-axis
    indicates the best observed objective value.
    Solid lines represent average values, while
    shaded areas depict the standard error of the mean.
  }
  \label{fig:reasoning-bo-vs-baselines}
\end{figure}

\begin{table}[h]
  \centering
  \footnotesize \resizebox{\textwidth}{!}{
    \begin{tabular}{l@{\hskip 4pt}|@{\hskip 4pt}l@{\hskip
        4pt}|@{\hskip 4pt}cccc@{\hskip 4pt}|@{\hskip 4pt}ccc@{\hskip
      4pt}|@{\hskip 4pt}ccc}
      \toprule \textbf{Benchmark}                         &
      \textbf{Method} & CV $\downarrow$ & Std $\downarrow$ & Log
      Regret $\downarrow$ & Log AUC $\uparrow$ & CVaR@0.1 $\uparrow$
      & CVaR@0.3 $\uparrow$ & CVaR@0.5 $\uparrow$ & IMP@1 $\uparrow$
      & IMP@3 $\uparrow$ & IMP@5 $\uparrow$ \\
      \midrule \multirow{5}{*}{Direct Arylation}          & Random
      Search   & 0.67            & 26.49            & 7.14
      & 7.01               & 2.67                & 9.53
      & 17.22               & 28.74            & 16.93            &
      27.57            \\
      & Vanilla BO      & 0.35            & 22.75            & 6.64
      & 7.53               & 19.76               & 40.63
      & 52.67               & 43.62            & 45.17            &
      55.91            \\
      & CMA-ES          & 0.28            & 15.29            & 6.41
      & 7.41               & 32.73               & 46.34
      & 52.08               & 25.20            & 32.73            &
      39.32            \\
      & Analytic EI     & \textbf{0.16}   & \textbf{8.69}    & 6.11
      & 7.50               & \textbf{53.35}      & 58.03
      & 62.40               & 51.23            & 53.35            &
      54.22            \\
      & \mname          & 0.24            & 13.77            &
      \textbf{6.03}           & \textbf{7.56}      & 44.18
      & \textbf{60.76}      & \textbf{67.71}      & \textbf{60.07}
      & \textbf{66.64}   & \textbf{71.22}   \\
      \midrule \multirow{5}{*}{Suzuki}                    & Random
      Search   & 0.34            & 21.45            & 6.65
      & 7.50               & 22.98               & 39.32
      & 50.27               & 52.43            & 48.43            &
      59.43            \\
      & Vanilla BO      & 0.20            & 12.21            & 5.42
      & 7.59               & 49.97               & 67.26
      & 71.02               & 62.32            & 49.97            &
      60.33            \\
      & CMA-ES          & 0.23            & 15.41            & 6.00
      & 7.67               & 43.79               & 66.20
      & 73.13               & 38.16            & 43.79            &
      56.52            \\
      & Analytic EI     & 0.29            & 19.84            & 5.93
      & 7.63               & 33.98               & 53.30
      & 65.35               & 41.91            & 40.94            &
      42.57            \\
      & \mname          & \textbf{0.14}   & \textbf{7.71}    &
      \textbf{4.91}           & \textbf{7.69}      & \textbf{64.76}
      & \textbf{74.66}      & \textbf{78.02}      & \textbf{67.85}
      & \textbf{69.52}   & \textbf{73.39}   \\
      \midrule \multirow{5}{*}{CPA-Catalyzed Thiol-Imine} & Random
      Search   & 0.45            & 1.40             & 5.54
      & 4.00               & 0.51                & 1.15
      & 1.60                & 0.86             & 0.85             &
      1.27             \\
      & Vanilla BO      & 0.27            & \textbf{1.21}    & 5.58
      & 4.21               & 1.06                & 1.72
      & 2.16                & 1.45             & 1.53             &
      1.70             \\
      & CMA-ES          & 0.33            & 1.43             &
      \textbf{5.46}           & 4.32               & 0.91
      & 1.70                & 2.21                & 0.77
      & 0.91             & 1.27             \\
      & Analytic EI     & 0.27            & 1.29             & 5.49
      & 4.33               & 1.02                & 1.90
      & 2.46                & 0.90             & 1.05             &
      1.34             \\
      & \mname          & \textbf{0.21}   & 1.22             & 5.52
      & \textbf{4.50}      & \textbf{1.57}       & \textbf{2.35}
      & \textbf{2.88}       & \textbf{1.54}    & \textbf{1.74}    &
      \textbf{2.07}    \\
      \bottomrule
    \end{tabular}
  }
  \caption{Quantitative performance comparison of optimization
    methods across three benchmarks. Bold values indicate the best
  quantitative result in each column.}
  \label{tab:merged_table}
\end{table}

\textbf{Empirical Insights}

(1) \textbf{Performance:} As shown in Figure~\ref{fig:reasoning-bo-vs-baselines}, \mname\ robustly and efficiently outperforms baselines across diverse benchmarks, including both real-world chemical optimization tasks and synthetic mathematical functions. This highlights the algorithm's effectiveness in navigating intricate search spaces.

(2) \textbf{Initialization Strategy:} \mname\ employs an advanced initialization strategy that leverages expert priors and domain-specific knowledge to systematically select critical starting points. For example, it initializes near the known optimum of the Levy function ([1, ..., 1]) to promote rapid convergence and identifies favorable starting points in the Rosenbrock function's narrow valley to accelerate progress. For functions with asymmetric bounds like Ackley, its adaptive learning mechanism effectively mitigates the impact of suboptimal initial points.

\subsection{Extended Experimental Analysis}

\begin{figure}[htbp]
  \centering
  \includegraphics[width=0.6\textwidth]{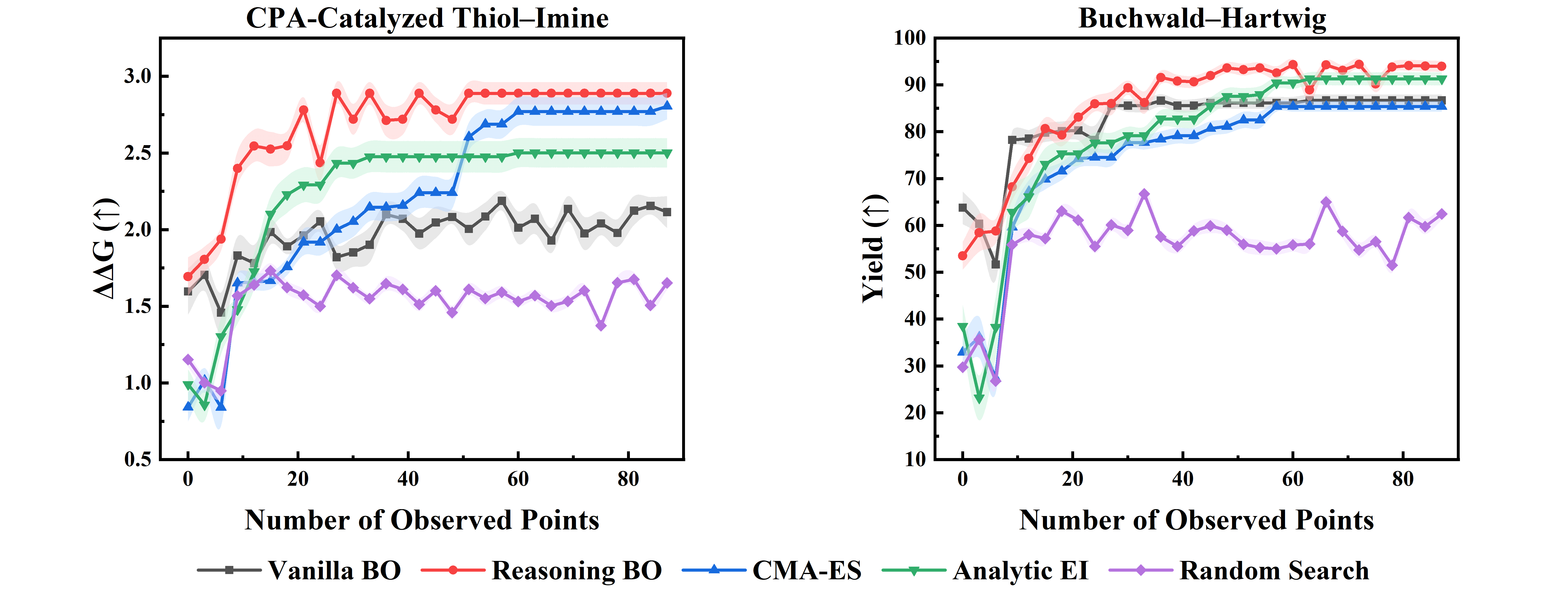}
  \caption{Chemical reaction benchmarks: CPA-Catalyzed Thiol–Imine
  and Buchwald-Hartwig amination.}
  \label{fig:chem_results}
\end{figure}

This subsection analyzes our model's performance on additional
chemical datasets and high-dimensional benchmarks.
Figure~\ref{fig:chem_results} presents optimization trajectories
across two chemical reactions. In the CPA-Catalyzed Thiol–Imine
reaction, our framework achieves 0.20 kcal/mol improvement during
initialization and 0.67 kcal/mol improvement during convergence
compared to vanilla BO. For Buchwald-Hartwig amination, we achieve
10.37\% yield improvement (95.06\% vs. 84.68\%).

\begin{figure}[htbp]
  \centering
  \includegraphics[width=0.6\textwidth]{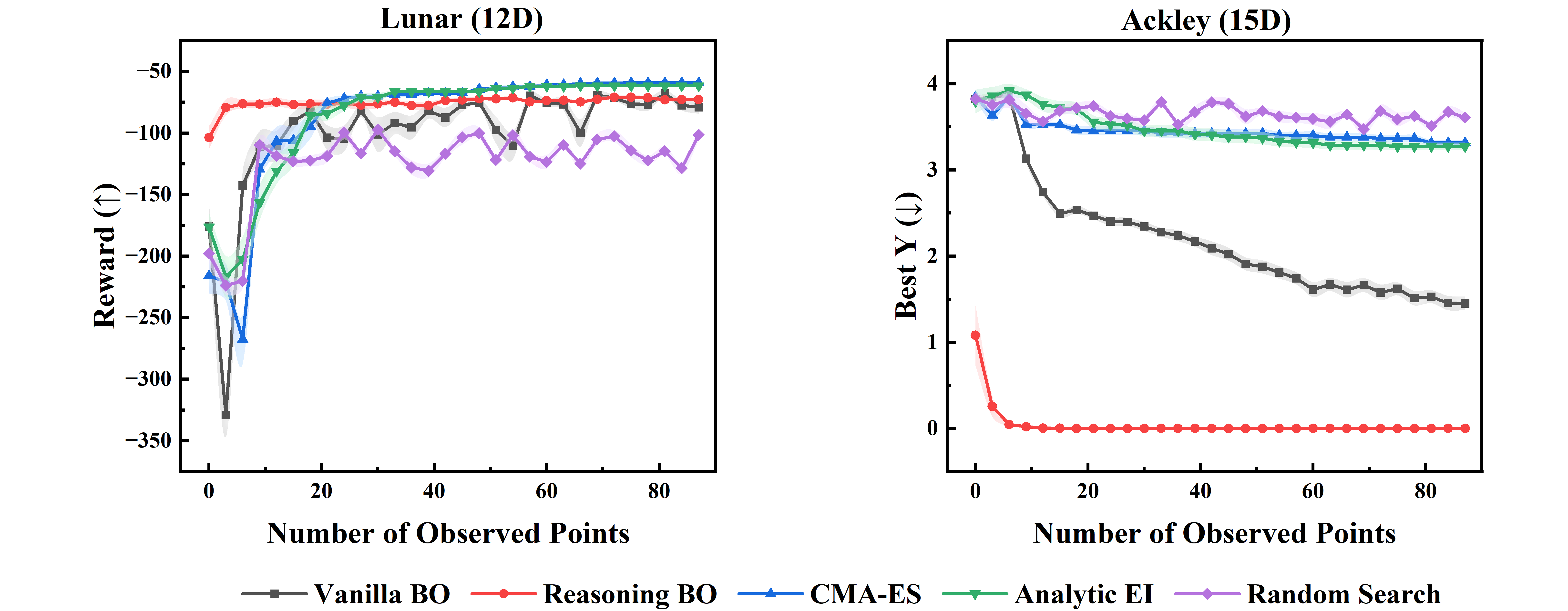}
  \caption{High-dimensional benchmarks: Lunar Lander (12D) and Ackley (15D).}
  \label{fig:highdim}
\end{figure}

Our approach also demonstrates notable improvements on
high-dimensional benchmarks (Figure~\ref{fig:highdim}). On Lunar
Lander (12D), we achieve 71.02 higher initial reward compared to
vanilla BO. For Ackley (15D), our algorithm shows -2.34 difference
during initialization and -1.46 final difference at 90 observations.

\subsection{Ablation Studies}

We conduct ablation studies to investigate the impact of
different components in our framework. These studies include initialization
vs optimization analysis, structured notes analysis, and reinforcement learning
model comparisons.

\textbf{Initialization vs Optimization Analysis.} To systematically
analyze the contributions of our algorithm's initialization
strategy and continuous optimization capability, we conduct comprehensive
ablation studies comparing different initialization-optimization
combinations. Figure~\ref{fig:ablations_init} presents the comparative results
across multiple benchmarks.
\textit{\textbf{Empirical Insights:}} The analysis
reveals that Reasoning BO
initialization consistently identifies better starting points
compared to Sobol sampling in Vanilla BO, achieving 44.6\% higher
initial performance in Direct Arylation (66.08\% vs 21.62\%). The RBO
optimization phase consistently outperforms VBO regardless of
initialization method, validating our algorithm's superior
optimization capability. The RBO$\rightarrow$VBO configuration
demonstrates that good initialization alone can help Vanilla BO
surpass its full version, though with less efficiency than full RBO.

\begin{figure}[htbp]
  \centering
  \includegraphics[width=\linewidth]{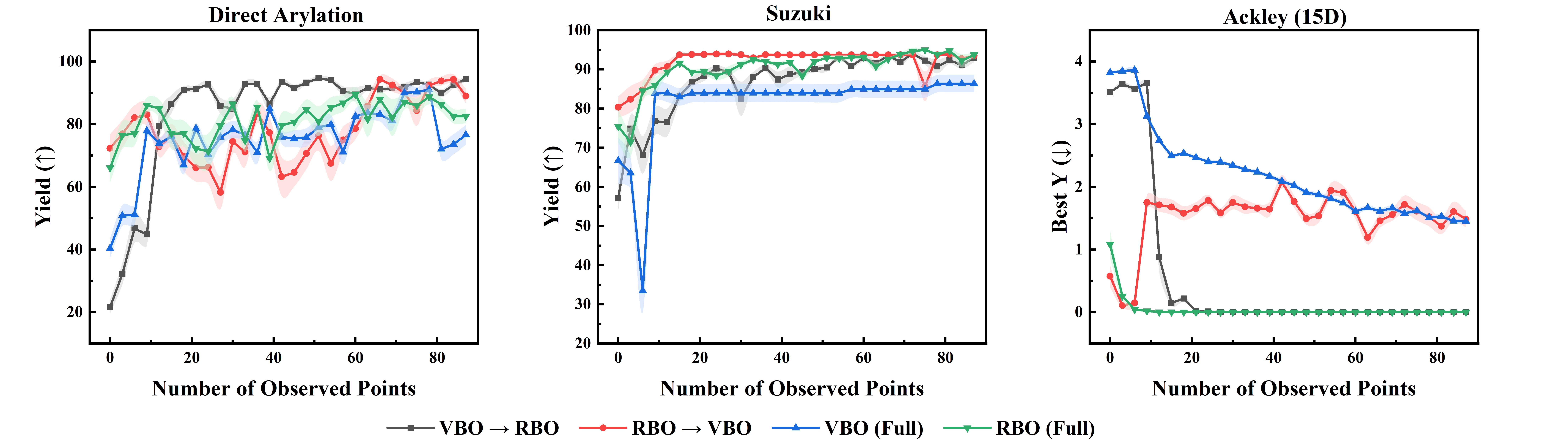}
  \caption{Ablation study comparing different initialization and
    optimization strategies. Four configurations are evaluated: full
    Reasoning BO (RBO), full Vanilla BO (VBO), RBO initialization followed
    by VBO optimization (RBO$\rightarrow$VBO), and VBO initialization followed
  by RBO optimization (VBO$\rightarrow$RBO).}
  \label{fig:ablations_init}
\end{figure}

\textbf{Structured Notes Analysis.} We investigate the impact of
incorporating structured notes in our framework. While database mechanisms
introduce minimal latency in real-world applications, we previously conducted
experiments without database integration for rapid validation. Here we introduce
structured notes as an ablation study to evaluate the contribution of
knowledge-guided reasoning. The results are visualized in
Figure~\ref{fig:ablations}.
\textit{\textbf{Empirical Insights:}} The ablation study reveals
domain-dependent
effects of incorporating notes. Chemical applications show consistent
improvements:
Suzuki coupling achieves 2\% better initial performance, while Direct Arylation
demonstrates significantly better optimization outcomes during
mid-to-late stages.
For the mathematical Hartmann function, where prior knowledge is limited, notes
constrain exploration and slow optimization progress. These findings
suggest that
structured notes provide substantial benefits in knowledge-rich domains but may
hinder performance in purely synthetic optimization scenarios.

\begin{figure}[htbp]
  \centering
  \includegraphics[width=0.8\textwidth]{
    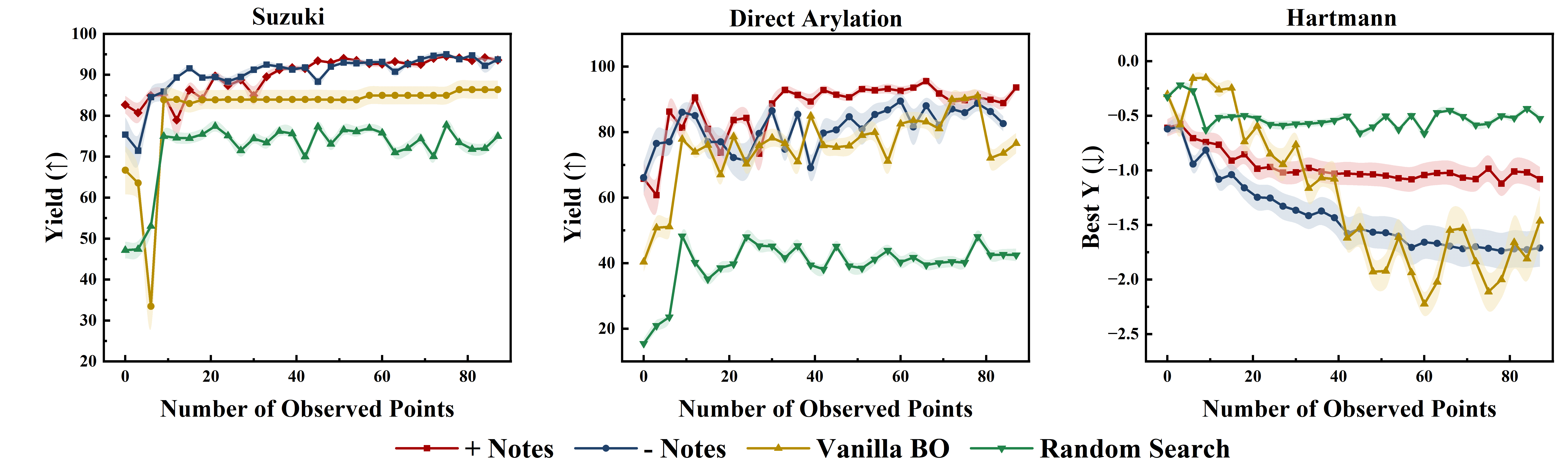
  }
  \caption{Ablation study comparing \mname\ with and without structured notes
    on three optimization benchmarks: (a) Suzuki coupling, (b) Direct Arylation
    reactions, and (c) 6D Hartmann function, alongside random search and vanilla
  BO baselines.}
  \label{fig:ablations}
\end{figure}

\textbf{Reinforcement Learning Enhanced Models.}
We compare the capabilities of our RL-enhanced
models against baseline. Figure~\ref{fig:rl_comparison} shows the
performance comparison between our RL-14b model, Instruct-14b,
and Distilled-14b variants, along with larger models for scalability assessment.
\textit{\textbf{Empirical Insights:}} The RL-enhanced models demonstrate
significant improvements over baseline approaches, with the RL-14b
model showing superior performance across multiple benchmarks.
The comparison with larger models validates the scalability of our
approach, while the distilled variants maintain competitive
performance with reduced computational requirements.

\begin{figure}[htbp]
  \centering
  \includegraphics[width=0.6\textwidth]{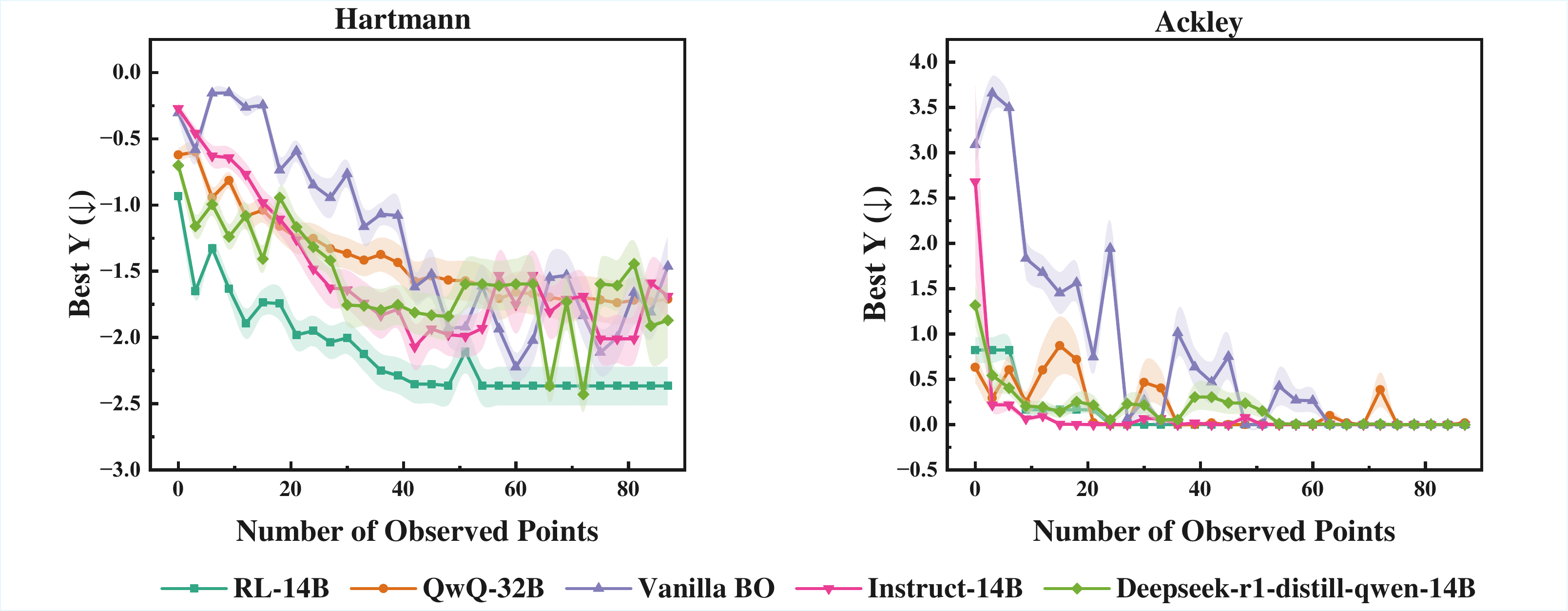}
  \caption{Performance comparison of RL-enhanced models versus baseline
  approaches with similar parameter counts.}
  \label{fig:rl_comparison}
\end{figure}

\section{Conclusions}
This paper has presented \mname, a novel framework that
systematically integrates
knowledge graphs, multi-agent systems, and reinforcement learning
into the Bayesian
optimization paradigm, establishing an intelligent optimization system for
scientific discovery with superior cost-effectiveness. The framework
innovatively
combines the reasoning capabilities of LLMs with the knowledge
storage properties of knowledge graphs and vector databases, creating a
dynamic knowledge fusion mechanism that effectively injects domain knowledge
into the optimization process. The multi-agent architecture ensures flexibility
and adaptability, providing a unified and customizable solution for
user-friendly
interactions. 
Future research directions
include developing
global attention mechanisms to overcome context length constraints, creating
novel knowledge distillation paradigms to enhance knowledge extraction
efficiency, and exploring synergistic optimization paths between reinforcement
learning algorithms and generated data, further unlocking the reasoning
potential of LLMs in Bayesian optimization.

\newpage

\bibliography{iclr2026_conference}
\bibliographystyle{iclr2026_conference}

\appendix

\newpage


\large{\textbf{Appendix}}

\section{Extended Results}
\subsection{Impact of Different API Models}
\label{sec:api_models}

We evaluate the performance of three API models (DeepSeek, qwq-plus, and
deepseek-distilled-qwen-7b) across three distinct datasets to assess
their reasoning capabilities within our framework.
Figure~\ref{fig:api_comparison} presents
the comparative results on Suzuki coupling, Direct Arylation, and 6D
Hartmann function benchmarks.

\vspace{-0.5em}

\textit{\textbf{Empirical Insights:}} On
chemical datasets, both qwq-plus and
DeepSeek substantially outperform the baseline BO,
achieving comparable results that highlight their strong reasoning
capabilities. For the Hartmann function benchmark, DeepSeek initially leads
but shows limited exploration in later stages, while qwq-plus maintains more
consistent performance. The distilled model demonstrates slight advantages
over the baseline, suggesting that model capacity directly influences
optimization effectiveness.

\begin{figure}[htbp]
  \centering
  \includegraphics[width=\textwidth]{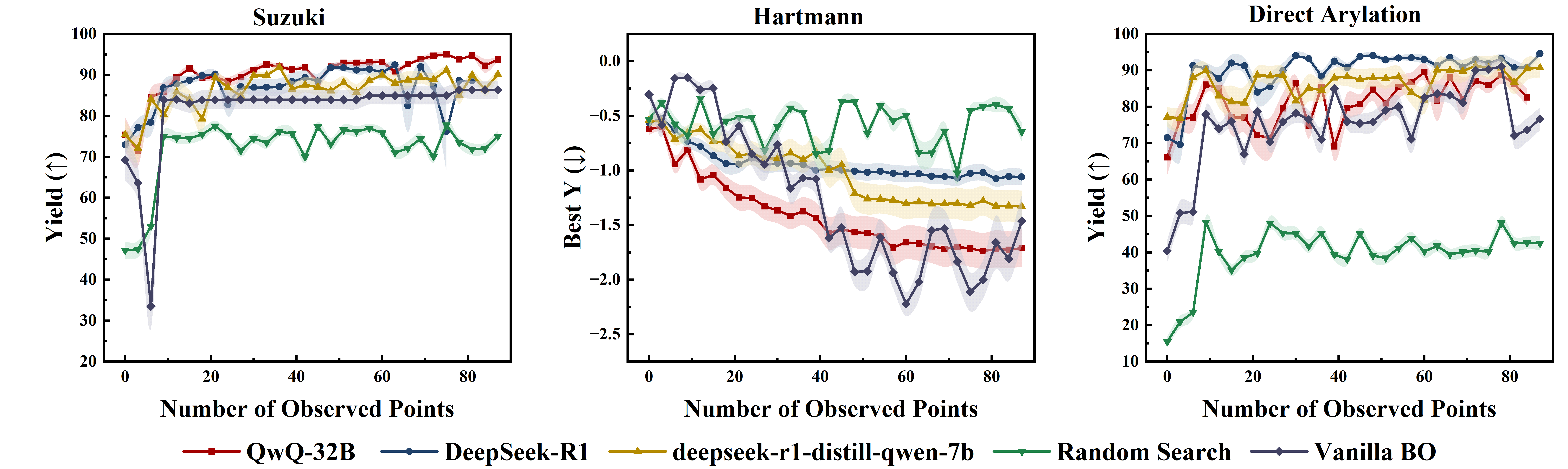}
  \caption{Performance comparison of different API models on three
  optimization tasks.}
  \label{fig:api_comparison}
\end{figure}

\subsection{Error Bar Analysis}

\begin{figure}[htbp]
  \centering
  \includegraphics[width=0.9\textwidth]{
    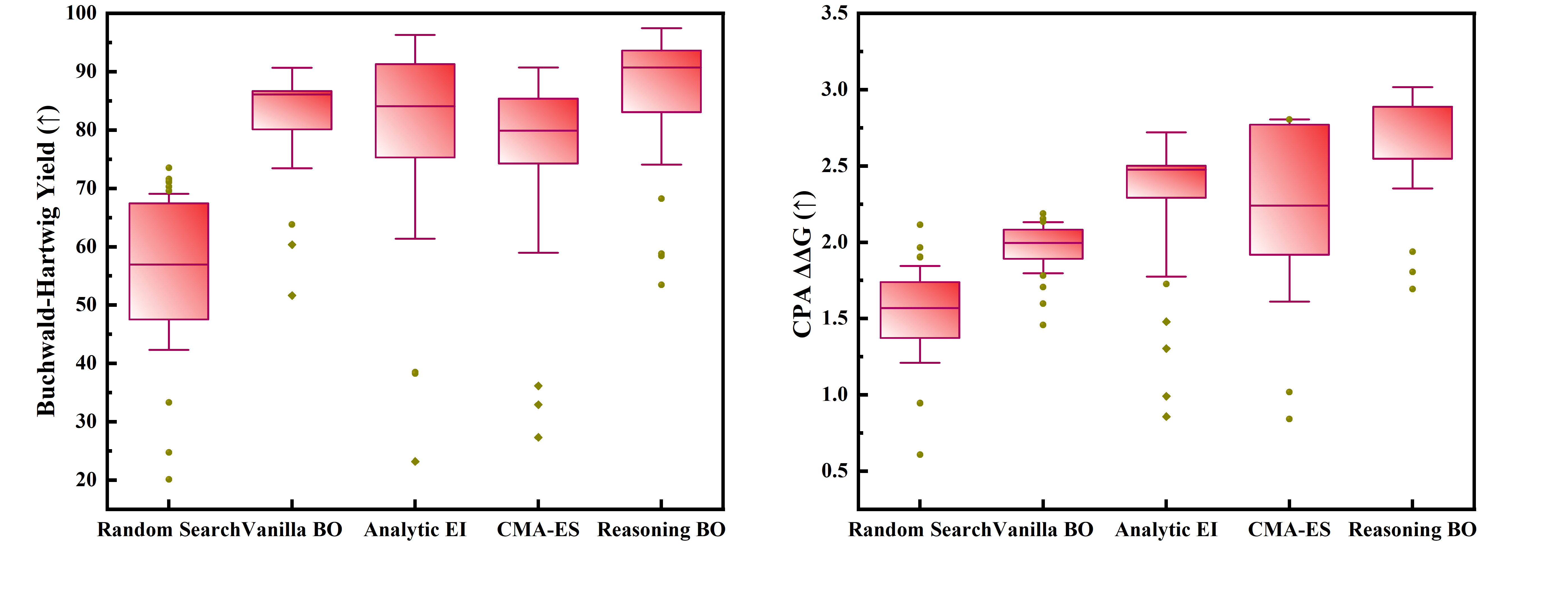
  }
  \caption{Error bar analysis of optimization performance on two chemical
    benchmarks: CPA-Catalyzed Thiol-Imine and Buchwald-Hartwig amination.
    The figure depicts the mean and standard deviation of the optimization
    results across multiple trials, highlighting the consistency and robustness
  of our framework compared to baseline methods.}
  \label{fig:errorbar_chem}
\end{figure}

For comprehensive error bar analyses of chemical and high-dimensional
benchmarks,
refer to Figures~\ref{fig:errorbar_chem}
and~\ref{fig:errorbar_highdim} respectively.

\begin{figure}[htbp]
  \centering
  \includegraphics[width=0.9\textwidth]{
    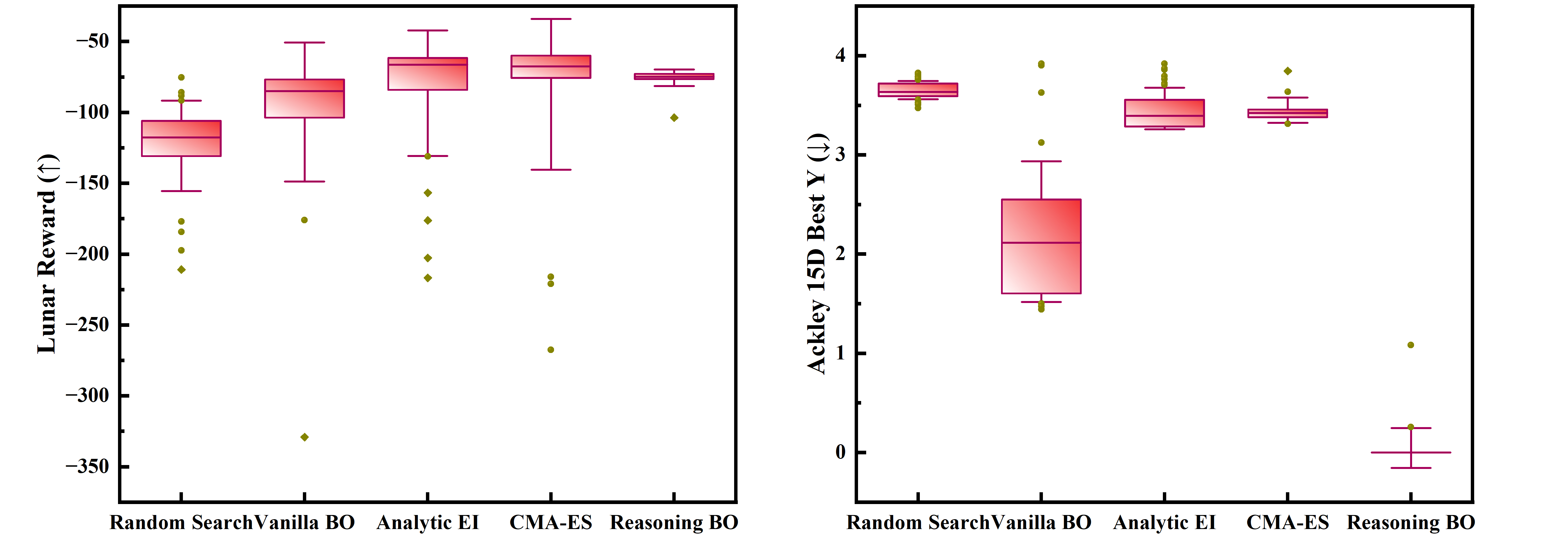
  }
  \caption{Error bar analysis of optimization performance on high-dimensional
    benchmarks: Lunar Lander (12D) and Ackley (15D). The figure illustrates
    the mean and standard deviation across multiple trials, demonstrating the
    stability and effectiveness of our framework in handling high-dimensional
  problems.}
  \label{fig:errorbar_highdim}
\end{figure}

\section{Technical Implementation Details}
\label{sec:lora_details}

\subsection{Low-Rank Adaptation (LoRA) Implementation}

LoRA enhances fine-tuning efficiency by freezing the pre-trained model weights
to prevent catastrophic forgetting, while introducing trainable rank
decomposition matrices $B$ and $A$ into the multi-head self-attention layers
of the transformer architecture. These matrices are sensitive to task-specific
adaptations and allow the model to learn domain-specific terminology and
relationships effectively. During forward propagation, the input $x$ is
processed as follows:

\begin{equation}
  \varDelta W = BA, \quad h = W_{0}x + \varDelta Wx = W_{0}x + BAx,
\end{equation}

where $W_{0}$ represents the original weight matrix of the pre-trained model,
and $\varDelta W$ denotes the incremental adjustment introduced by LoRA.
This approach ensures that the model's general linguistic capabilities
remain intact while adapting to the target domain. The rank decomposition
significantly reduces the number of trainable parameters while maintaining
effective adaptation to domain-specific optimization tasks.

\subsection{Baseline Configuration Details}
\label{sec:baseline_configs}

This section provides detailed parameter configurations for CMA-ES baseline
method used in our experiments.

\textbf{CMA-ES Parameters:} The Covariance Matrix Adaptation Evolution Strategy
is configured with the following parameters based on our implementation:
population size (popsize) = 5, maximum iterations (maxiter) = 6, initial step
size $\sigma_0 = 0.1$, with variable bounds set to [0, 1] for normalized
search space. The algorithm is initialized with 3 Sobol-generated points
and uses Upper Confidence Bound (UCB) acquisition function with $\beta = 0.1$
for Gaussian Process-guided exploration. Each experiment is repeated 10 times
for statistical robustness, with 30 total evaluation trials per run.

\section{Case Study}
\label{sec:case_study}

\subsection{1. Experiment Compass Input}

The algorithm begins with a user-prepared Experiment Compass, which serves as
the structured input containing all necessary experimental parameters
and constraints.
This JSON-formatted document specifies the reaction components, optimization
objectives, and parameter bounds for the Buchwald-Hartwig amination case
study. The complete Experiment Compass configuration used in our
evaluation is detailed below.

\input{other/buchwald_compass_config.tex}

\subsection{2. Overview Generation}

The framework processes the Experiment Compass through a large language
model (LLM) to generate a comprehensive overview of the Buchwald-Hartwig
amination reaction. This overview synthesizes key experimental parameters,
chemical constraints, and optimization objectives into a structured knowledge
representation that guides subsequent algorithmic decisions. The
generated overview is presented below.

\input{other/buchwald_overview.tex}

\subsection{3. Candidate Initialization} 
        
The initialization phase leverages the
reasoning capabilities of LLMs to generate promising candidate points within
the constrained parameter space defined in the Experiment Compass. The LLM receives structured prompts
containing domain knowledge and optimization objectives, as detailed below.
        
\input{other/buchwald_init_prompt.tex}

The model then synthesizes
these inputs to produce initialization insights, recommending candidate points that are strategically positioned in regions
likely to yield high performance. An example of the generated insights is shown below.

\input{other/buchwald_init_insight.tex}

These intelligent initializations serve as
the starting points for subsequent Bayesian optimization. Following the
execution of initial experiments based on these recommendations, the framework
incorporates both the newly acquired experimental data and the initialization
points to generate refined suggestions for the next iteration, which are
then fed into the optimization loop for further refinement.

\textbf{4. Optimization Loop}

The framework enters an iterative optimization cycle. In each loop, the LLM receives a detailed prompt containing all historical experimental data, previously generated insights, new candidate points from the optimizer, and specific instructions for the current task. We provide two examples of such prompts, for the Buchwald-Hartwig and Suzuki coupling reactions, respectively:
\input{other/buchwald_hartwig_optimization_prompt.tex}
\input{other/optimization_loop.tex}

Based on such a prompt, the LLM analyzes the provided data and generates new hypotheses to guide the subsequent round of experiments. An example of the LLM's output, containing its reasoning and recommended parameter sets, is shown below:
\input{other/buchwald_loop_insight.tex}

These recommendations are then evaluated through Bayesian
optimization, and the results are fed back into the system for the
next iteration, continuing until predefined limits are reached.

\subsection{5. Summary and Conclusion}

In the final stage of the optimization framework, the LLM synthesizes all
accumulated insights and experimental data to generate comprehensive
summaries and conclusions. The system's final summary of the
Buchwald-Hartwig amination optimization is presented below, highlighting
key findings and performance metrics. The concluding analysis
provides actionable recommendations and broader implications for
reaction optimization, demonstrating the framework's ability to
extract meaningful chemical
insights from complex experimental data.

\input{other/buchwald_summary.tex}

\input{other/buchwald_conclusion.tex}

\section{Benchmarks}
\label{sec:benchmarks}

\subsection{Suzuki Reaction (5D)}
The Suzuki-Miyaura cross-coupling reaction is a widely used strategy in
organic synthesis for forming biaryl motifs by coupling an aryl
halide (electrophile)
with a boronic acid derivative (nucleophile), catalyzed by palladium complexes.
This benchmark simulates a combinatorial exploration of substrate pairs and
reaction conditions in a data-driven optimization task.

\begin{itemize}
  \item \textbf{Objective}: Maximize the product yield (\%) of the
    Suzuki reaction
    through selection of optimal electrophile-nucleophile pairs, ligand,
    base, and solvent under catalyst compatibility constraints.

  \item \textbf{Optimization Variables}: The input variables include
    the electrophile
    SMILES, nucleophile SMILES, ligand SMILES, base SMILES, and solvent SMILES,
    which are described in detail in Table~\ref{tab:suzuki_vars}.
\end{itemize}

\begin{table}[htbp]
  \centering
  \begin{tabular}{ll}
    \toprule Variable Name        & Description
    \\
    \midrule Electrophile\_SMILES & Aryl halide substrate represented
    as SMILES. Four halide derivatives tested. \\
    Nucleophile\_SMILES           & Boronic acid substrate in SMILES
    form. Three boronate variants included.     \\
    Ligand\_SMILES                & Phosphine ligands used to form
    Pd(0) complexes. Eleven ligands included.     \\
    Base\_SMILES                  & Base required for the
    transmetallation step. Seven options provided.         \\
    Solvent\_SMILES               & Solvent medium (SMILES); affects
    solubility and catalyst stability.          \\
    \bottomrule
  \end{tabular}
  \caption{Optimization variables for the Suzuki reaction benchmark}
  \label{tab:suzuki_vars}
\end{table}

\begin{itemize}
  \item \textbf{Target}: Simulated or experimentally derived product yield
    (\%) of the coupling reaction.

  \item \textbf{Constraints}:
    \begin{itemize}
      \item All reagent combinations must form stable palladium complexes.

      \item Conditions must support the full catalytic cycle including
        oxidative addition, transmetallation, and reductive elimination.
    \end{itemize}

  \item \textbf{Challenges}:
    \begin{itemize}
      \item Discrete Combinatorial Landscape: The entire search space
        is composed of categorical variables with strong dependency
        interactions, making it highly multimodal and sparse.

      \item Chemical Compatibility Filtering: Some electrophile--ligand--base
        combinations are chemically incompatible and lead to
        decomposition or catalyst poisoning, creating discontinuities
        in the yield landscape.

      \item Ligand Diversity: The ligands span a broad range of steric
        and electronic profiles, affecting both oxidative addition rates
        and Pd complex stability.
    \end{itemize}
\end{itemize}

This benchmark is designed to assess the optimizer's ability to navigate highly
categorical chemical reaction spaces, identify synergistic effects, and avoid
known failure modes due to chemical incompatibility---all essential
in real-world
medicinal chemistry and process development workflows.

\subsection{Direct Arylation Reaction (5D)}

Direct Arylation Reaction Optimization (5D) enables C--H activation without the
need for pre-functionalized substrates, offering a highly atom-economical
route for biaryl bond formation. However, this reaction is notoriously
sensitive to reaction conditions, including ligand/base/solvent
synergy and subtle
effects from concentration and temperature. This benchmark simulates a
complex reaction landscape composed entirely of categorical and discrete
variables, designed to optimize reaction yield under experimentally feasible
constraints.

\begin{table}[htbp]
  \centering
  \begin{tabular}{lp{9cm}}
    \toprule Variable Name & Description
    \\
    \midrule Base\_SMILES  & Base used in the reaction, including its
    counterion (e.g., cesium or potassium salts). \\
    Ligand\_SMILES         & Phosphine ligand coordinating the Pd
    center. Spans bulky to $\pi$-extended ligands.    \\
    Solvent\_SMILES        & Solvent system influencing reactivity
    and selectivity. Four options included.          \\
    Concentration          & Molar concentration (mol/L) of the
    reactants. Discrete experimental levels.            \\
    Temp\_C                & Reaction temperature (°C). Three
    practical heating levels are considered.              \\
    \bottomrule
  \end{tabular}
  \caption{Optimization variables for the Direct Arylation reaction
  benchmark}
  \label{tab:arylation_vars}
\end{table}

\begin{itemize}
  \item \textbf{Objective}: Maximize the isolated product yield (\%) of a direct
    arylation reaction by selecting an optimal combination of ligand,
    base, solvent, concentration, and temperature.

  \item \textbf{Optimization Variables}: As detailed in
    Table~\ref{tab:arylation_vars},
    we optimize five key variables: Base\_SMILES, Ligand\_SMILES,
    Solvent\_SMILES,
    Concentration, and Temp\_C.

  \item \textbf{Target}: Reaction\_Yield (\%) - isolated or calculated product
    yield under each condition set.

  \item \textbf{Constraints}:
    \begin{itemize}
      \item All SMILES strings must correspond to chemically feasible
        components.

      \item Temperature and concentration values are restricted to predefined,
        experimentally validated levels.
    \end{itemize}

  \item \textbf{Challenges}:
    \begin{itemize}
      \item Categorical-Only Input Space: All variables are discrete, forcing
        the optimizer to navigate a fully combinatorial landscape.

      \item Multi-Factor Interactions: Ligand--base--solvent
        compatibility plays a critical role; some combinations lead to
        zero or negligible reactivity.

      \item Kinetic--Thermodynamic Balance: Subtle shifts in temperature
        or dilution can invert reaction selectivity or alter decomposition
        profiles.
    \end{itemize}
\end{itemize}

This benchmark serves as a stringent test for discrete Bayesian optimization
frameworks, requiring reasoning over rich chemical priors and non-obvious
synergistic effects. It reflects a realistic synthetic design problem
where trial
numbers must be minimized, and each reaction carries real-world material and
time cost.

\subsection{Buchwald-Hartwig Reaction (4D)}
The Buchwald-Hartwig amination is a palladium-catalyzed cross-coupling
reaction that forms C-N bonds, widely used in pharmaceutical
synthesis and materials
science \citep{shields_bayesian_2021}. This benchmark evaluates optimization
algorithms on a 4-dimensional categorical space representing key
reaction components.

\begin{table}[htbp]
  \centering
  \begin{tabular}{lp{9cm}}
    \toprule Variable       & Description
    \\
    \midrule Ligand\_SMILES & Phosphine-based ligands modulating
    catalyst activity. Includes monophosphines and bulky phosphines. \\
    Additive\_SMILES        & Reaction modifiers affecting
    intermediate stability and side reactions.                             \\
    Base\_SMILES            & Reagents supporting deprotonation and
    catalyst turnover.                                            \\
    ArylHalide\_SMILES      & Electrophilic component determining
    oxidative addition efficiency.                                  \\
    \bottomrule
  \end{tabular}
  \caption{Optimization variables for the Buchwald-Hartwig reaction
  benchmark}
  \label{tab:buchwald_vars}
\end{table}

\begin{itemize}
  \item \textbf{Objective}: Maximize the isolated yield (\%) of the desired
    amine product by optimizing ligand, additive, base, and aryl halide
    combinations while maintaining catalytic cycle stability.

  \item \textbf{Optimization Variables}: As detailed in
    Table~\ref{tab:buchwald_vars},
    we optimize four categorical variables: Ligand\_SMILES, Additive\_SMILES,
    Base\_SMILES, and ArylHalide\_SMILES.

  \item \textbf{Target}: Reaction\_Yield (\%) - isolated product yield under
    each condition set.

  \item \textbf{Challenges}:
    \begin{itemize}
      \item Fully Categorical Space: Requires navigation of discrete
        combinations with strong multi-modality.

      \item Catalyst Deactivation: Some combinations may lead to
        inactive catalytic systems.

      \item Complex Ligand Effects: Steric and electronic properties
        dramatically influence reactivity.
    \end{itemize}
\end{itemize}

This benchmark provides a realistic test for discrete optimization in chemical
reaction spaces, requiring algorithms to recognize synergistic patterns
while avoiding invalid regions - a critical capability for pharmaceutical
process development.

\subsection{Chiral Phosphoric Acid–Catalyzed Thiol–Imine Addition Reactions
(3D)}
The chiral phosphoric acid (CPA)–catalyzed addition of thiols to N-acylimines
is a powerful strategy for constructing chiral thioaminal motifs, widely
used in asymmetric catalysis and pharmaceutical synthesis. This benchmark task
focuses on optimizing reaction conditions to achieve high enantioselectivity
by maximizing the free energy difference ($\Delta\Delta G$, kcal/mol)
between competing transition states leading to each enantiomer.

\begin{table}[htbp]
  \centering
  \begin{tabular}{lp{9cm}}
    \toprule Variable & Description
    \\
    \midrule Catalyst & Chiral phosphoric acids derived from the
    BINOL scaffold, with diverse steric and electronic profiles
    through variation at the 3,3'-substitution sites. \\
    Imine             & N-acyl imine substrates derived from various
    aryl aldehydes and amines, serving as the electrophilic partner.
    \\
    Thiol             & A range of nucleophilic thiol reagents whose
    structure impacts both reactivity and side reaction propensity.
    \\
    \bottomrule
  \end{tabular}
  \caption{Optimization variables for the Chiral Phosphoric Acid–Catalyzed
  Thiol–Imine Addition Reaction benchmark}
  \label{tab:cpa_vars}
\end{table}

\begin{itemize}
  \item \textbf{Objective}: Maximize enantioselectivity ($\Delta\Delta G$)
    while preserving mechanistic integrity during the asymmetric transformation.

  \item \textbf{Optimization Variables}: As detailed in
    Table~\ref{tab:cpa_vars},
    we optimize three categorical variables: Catalyst, Imine, and Thiol.

  \item \textbf{Target Metric}: Free energy difference ($\Delta\Delta
    G$, kcal/mol),
    quantifying enantioselectivity. Higher values indicate stronger
    stereoselective induction.

  \item \textbf{Experimental Constraints}:
    \begin{itemize}
      \item Reactions must maintain CPA catalytic activity and stereocontrol
        under organic solvent conditions.

      \item Avoid side reactions such as dehydration, polymerization, or
        oxidation by conducting the reaction under mild conditions.

      \item Catalysts must be synthetically accessible and conformationally
        stable.
    \end{itemize}

  \item \textbf{Optimization Challenges}:
    \begin{itemize}
      \item High-dimensional categorical space: Purely discrete
        variables with non-linear and multimodal response surfaces.

      \item Stereoselectivity sensitivity: Small structural changes in
        the CPA catalyst can significantly affect $\Delta\Delta G$
        and product selectivity.

      \item Catalyst–substrate synergy: Effective combinations depend
        on subtle non-covalent interactions and spatial
        complementarity.
    \end{itemize}
\end{itemize}

This benchmark provides a realistic challenge for evaluating optimization
algorithms' ability to navigate complex discrete spaces and identify synergistic
combinations that enhance enantioselectivity while avoiding failed reactions.

\subsection{Lunar Lander Task (12D)}
The Lunar Lander is a classic control problem where the objective is to design
a policy that safely lands a spacecraft on the moon's surface while minimizing
fuel consumption and distance from the target landing zone. This benchmark
evaluates optimization algorithms on a 12-dimensional space combining
continuous state variables and discrete actions.

\begin{table}[htbp]
  \centering
  \begin{tabular}{lll}
    \toprule Variable             & Description
    & Bound         \\
    \midrule horizontal\_position & Normalized x-coordinate of lander
    & [-1.5, 1.5]   \\
    vertical\_position            & Normalized y-coordinate of lander
    & [-1.5, 1.5]   \\
    horizontal\_velocity          & Normalized x-velocity
    & [-5.0, 5.0]   \\
    vertical\_velocity            & Normalized y-velocity
    & [-5.0, 5.0]   \\
    angle                         & Orientation in radians
    & [-3.14, 3.14] \\
    angular\_velocity             & Rotation rate
    & [-5.0, 5.0]   \\
    left\_leg\_contact            & Boolean ground contact
    & [0, 1]        \\
    right\_leg\_contact           & Boolean ground contact
    & [0, 1]        \\
    no\_action                    & No engine firing
    & [0, 1]        \\
    fire\_left\_engine            & Left rotation thrust
    & [0, 1]        \\
    fire\_main\_engine            & Vertical deceleration
    & [0, 1]        \\
    fire\_right\_engine           & Right rotation thrust
    & [0, 1]        \\
    \bottomrule
  \end{tabular}
  \caption{Optimization variables for the Lunar Lander benchmark}
  \label{tab:lunar_vars}
\end{table}

\begin{itemize}
  \item \textbf{Objective}: Maximize the average landing reward over 50 terrain
    variations by optimizing the control policy's response to the 8D state
    space and 4D action space.

  \item \textbf{Optimization Variables}: As shown in Table~\ref{tab:lunar_vars},
    we optimize 12 variables including position, velocity, orientation
    states, and discrete engine control actions.

  \item \textbf{Target}: The average landing reward is calculated based on
    continuous rewards during descent and final outcomes. Each step's
    reward depends on the lander's position, velocity, and orientation
    relative to the landing pad. Leg contacts add +10 points each, while
    engine firings deduct points (-0.3 per main engine frame, -0.03 per side
    engine frame). The episode concludes with +100 for safe landing or -100
    for crashing, summed with all step rewards.

  \item \textbf{Constraints}:
    \begin{itemize}
      \item State variables must remain within physical bounds

      \item Only one action can be active at any time

      \item Successful landing requires coming to rest within target zone
    \end{itemize}

  \item \textbf{Challenges}:
    \begin{itemize}
      \item High-dimensional mixed space with continuous and discrete variables

      \item Delayed reward signals requiring long-term planning

      \item Precise control needed during final descent phase

      \item Trade-off between fuel efficiency and landing accuracy
    \end{itemize}
\end{itemize}

This benchmark tests an optimizer's ability to handle hybrid spaces and
delayed rewards in a physics-based environment. The problem
originates from OpenAI's
Gymnasium implementation
(\href{https://gymnasium.farama.org/environments/box2d/lunar_lander/}{Lunar
Lander environment}).

\subsection{Levy Function (5D)}

The Levy function \citep{laguna_experimental_2005} is a challenging synthetic
benchmark for global optimization algorithms, particularly in high-dimensional
spaces. The 5D implementation has the form:
\[
  f(\mathbf{x}) = sin^{2}(\pi w_{1}) + \sum_{i=1}^{4}(w_{i}- 1)^{2}[1 + 10\sin
  ^{2}(\pi w_{i}+ 1)] + (w_{5}- 1)^{2}[1 + \sin^{2}(2 \pi w_{5})]
\]
where $w_{i}= (1 + \frac{x_{i}- 1}{4})$ and
$\mathbf{x}= [x_{1}, x_{2}, x_{3}, x_{4}, x_{5}]$ is the input
vector. The function
has a global minimum of 0 at $x^{*}=[1,1,1,1,1]$ within the search space $x_{i}
\in [-10, 10]$. Its highly multimodal, non-convex landscape features numerous
local minima, complex variable interactions, and a rugged oscillatory surface
that becomes exponentially more challenging with increasing dimensionality,
making it an excellent test for an optimizer's ability to escape local
optima and navigate complex high-dimensional spaces with multiple basins of
attraction.

\subsection{Hartmann Function (6D)}

The 6-dimensional Hartmann function \citep{picheny:hal-00658212} is a
challenging synthetic benchmark for global optimization algorithms, particularly
in high-dimensional spaces. The function takes the form:
\[
  f(\mathbf{x}) = -\sum_{i=1}^{4}\alpha_{i}\exp\left(-\sum_{j=1}^{6}A_{ij}(
  x_{j}- P_{ij})^{2}\right)
\]
where $\mathbf{x}= [x_{1}, x_{2}, x_{3}, x_{4}, x_{5}, x_{6}]$ is the
input vector
within the hypercube $x_{i}\in (0,1)$. The 6D version contains approximately
$10^{6}$ local minima, with the global minimum surrounded by multiple
high-barrier
local optima. The function exhibits strong nonlinear coupling effects
between parameters, creating a complex energy landscape that tests an
optimizer's
ability to navigate multimodal spaces and escape local traps. The global
minimum value is approximately $-3.32237$ in the 6D case, with
numerous suboptimal
solutions separated by energy barriers $\Delta f \geq 0.5$, making it
particularly challenging for optimization algorithms to locate the true
optimum.

\subsection{Ackley Function (2D)}

The 2-dimensional Ackley function and 15-dimensional \citep{Adorio2005MVFM}
is a widely used benchmark for testing optimization algorithms'
ability to balance
global exploration and local exploitation. The function takes the form:
\[
  f(\mathbf{x}) = -20 \exp\left(-0.2 \sqrt{\frac{1}{2}\sum_{i=1}^{2}x_{i}^{2}}
  \right) - \exp\left(\frac{1}{2}\sum_{i=1}^{2}\cos(2\pi x_{i})\right) + 2
  0 + \exp(1)
\]
where $\mathbf{x}= [x_{1}, x_{2}]$ is the input vector within the search space
$x_{i}\in [-32.768, 32.768]$. The function has a global minimum of 0 at
$x^{*}=[0,0]$. Its landscape features an almost flat outer region that can trap
optimization algorithms in local optima, combined with a narrow funnel-shaped
global optimum region that requires precise local search. The deceptive nature
of the function comes from the interaction between the exponential and cosine
terms, creating a complex multimodal surface that tests an optimizer's
ability to escape flat regions while maintaining precision during final
convergence. The standard 2D version contains numerous shallow local
minima surrounding
the global optimum, making it particularly challenging for algorithms to
navigate between coarse and fine-grained search patterns.

\subsection{Rosenbrock Function (3D)}

The 3-dimensional Rosenbrock function \citep{Zimmermann1979TowardsGO} is a
classic benchmark for evaluating optimization algorithms in non-convex
spaces with challenging geometric properties. The function takes the form:
\[
  f(\mathbf{x}) = \sum_{i=1}^{2}\left[100(x_{i+1}- x_{i}^{2})^{2}+ (1 - x_{i}
  )^{2}\right]
\]
where $\mathbf{x}= [x_{1}, x_{2}, x_{3}]$ is the input vector within the
search space $x_{i}\in [-5, 10]$. The function has a global minimum of 0 at $x
^{*}=[1,1,1]$. Its landscape features a parabolic-shaped valley with a flat
bottom containing the global minimum, surrounded by steep walls that
create challenging
optimization dynamics. The 3D version exhibits the characteristic
difficulties of higher-dimensional Rosenbrock functions, including
gradient direction
oscillations in the valley region and deceptive local optima that can
trap optimization
algorithms. The function tests an optimizer's ability to navigate long, narrow,
and flat-bottomed valleys while maintaining convergence precision,
with the difficulty
increasing exponentially in higher dimensions due to the emergence of $2^{n-1}$
local minima for $n\geq4$.

\section{Baselines}
In our \mname\ approach, we primarily focus on the acquisition function
design by leveraging prior knowledge and LLMs' reasoning capabilities.
Therefore, our baseline selection mainly considers variations of acquisition
functions, particularly the LogEI family. Additionally, we include
Covariance Matrix Adaptation Evolution Strategy (CMA-ES) as a special
numerical optimization baseline. CMA-ES is a stochastic, derivative-free
optimization method for nonlinear or non-convex continuous problems, belonging
to the class of evolutionary algorithms that mimic biological
evolution principles
through mutation and selection mechanisms.

\subsection{LogEI Family}
\label{sec:logei_family}

The LogEI family addresses numerical stability issues in traditional
Expected Improvement (EI) by operating in log-space. While standard
EI implementations
can suffer from numerical underflow when improvement probabilities are small,
LogEI transformations enable stable computation across the full range of
possible inputs \citep{DBLP:conf/nips/AmentDEBB23}. The key transformation
is given by:

\[
  \text{LogEI}_{y^*}(\mathbf{x}) = \texttt{log\_h}((\mu(\mathbf{x}) - y^{*}
  ) / \sigma(\mathbf{x})) + \texttt{log}(\sigma(\mathbf{x}))
\]

where $y^{*}$ is the current best observation value, $\mu(\mathbf{x})$ is
the predicted mean, and $\sigma(\mathbf{x})$ is the predicted deviation.

\subsubsection{Analytic LogEI}
Analytic LogEI provides a numerically robust implementation through piecewise
decomposition:
\[
  \texttt{log\_h}(z) =
  \begin{cases}
    \texttt{log}(\phi(z) + z \Phi(z))
    & z > -1                           \\
    -z^{2}/ 2 - c_{1}+ \texttt{log1mexp}(\log(\texttt{erfcx}(-z /
    \sqrt{2}) |z|) + c_{2}) & - 1 / \sqrt{\epsilon}< z \leq -1 \\
    -z^{2}/ 2 - c_{1}-2 \log(|z|)
    & z \leq - 1 / \sqrt{\epsilon}
  \end{cases}
\]
where $c_{1}= \log(2\pi) / 2$, $c_{2}= \log(\pi / 2) / 2$, $\epsilon$ is the
numerical precision, $\phi$ is the standard normal PDF, $\Phi$ is the
standard normal CDF, \texttt{log1mexp} is a numerically stable
implementation of $\log (1 - \exp(z))$, and \texttt{erfcx} is a numerically
stable implementation of $\exp(z^{2})\text{erfc}(z)$.

\subsubsection{Monte Carlo Parallel LogEI}
For parallel batch optimization, Monte Carlo LogEI extends the
stability benefits
through:
\[
  \text{qLogEI}_{y^*}(\mathbf{X}) \approx \log\text{sumexp}_{i}\left(
    \tau_{\max}
    \log\text{sumexp}_{j}\left(\log\text{softplus}_{\tau_0}(\xi^{i}(\mathbf{x}
  _{j}) - y^{*})\right) / \tau_{\max}\right)
\]
where $i$ indexes Monte Carlo draws from the GP posterior,
$j=1,\dotsc,q$ indexes
candidates in the batch, $\tau_{0}$ and $\tau_{\max}$ are temperature
parameters controlling the approximation quality, and \texttt{logsoftplus}
is a numerically stable implementation of $\log(\log(1 + \exp(z)))$.

\subsection{CMA-ES}
CMA-ES (Covariance Matrix Adaptation Evolution Strategy) is an evolutionary
algorithm for continuous nonlinear optimization that adapts both the mean
and covariance matrix of its search distribution. The method combines
maximum-likelihood
principles with evolution path tracking, making it particularly effective
for complex, non-convex optimization problems
\citep{DBLP:journals/corr/Hansen16a}.

The algorithm updates its parameters through likelihood maximization similar
to expectation-maximization. The mean vector update maximizes:
\[
  m_{k + 1}= \underset{m}{\arg\max}\sum_{i = 1}^{\mu}w_{i}\log p_{\mathcal{N}}
  (x_{i:\lambda}\mid m)
\]
where $\log p_{\mathcal{N}}(x)$ is the log-likelihood from a multivariate
normal distribution:
\[
  \log p_{\mathcal{N}}(x) = -\frac{1}{2}\log \det(2\pi C) - \frac{1}{2}(x -
  m)^{T}C^{-1}(x - m)
\]
The covariance matrix update follows:
\[
  \sum_{i = 1}^{\mu}w_{i}\frac{x_{i:\lambda}-
  m_{k}}{\sigma_{k}}\left( \frac{x_{i:\lambda}-
  m_{k}}{\sigma_{k}}\right)^{T}= \underset{C}{\arg\max}\sum_{i = 1}^{\mu}w_{i}
  \log p_{\mathcal{N}}\left( \frac{x_{i:\lambda}- m_{k}}{\sigma_{k}}\mid C
  \right)
\]
These updates enable CMA-ES to learn a second-order model of the objective
function while requiring only solution rankings rather than derivatives. The
method has demonstrated superior performance to Bayesian optimization in certain
hyperparameter tuning tasks \citep{DBLP:journals/corr/LoshchilovH16}.
The algorithm's
effectiveness stems from several distinctive characteristics: its adaptive
step-size control mechanism prevents premature convergence, while the updates
can be interpreted as natural gradient descent. Furthermore, CMA-ES performs
principal component analysis on successful search steps to guide the
optimization
trajectory, and exhibits notable robustness when handling ill-conditioned
problems.

\section{System Messages and Prompts Templates}

\subsection{Experiment Compass Object}
The Experiment Compass serves as the standardized input template for
our algorithm,
enabling the description of complex objectives in natural language.
It incorporates
prior knowledge via the "description" field, aiding LLMs in generating
overviews and initial experimental points. Additionally, it standardizes
variable names for insights generation, ensuring consistency across multiple
rounds of experiments. The template below illustrates the structure
of the Experiment Compass.

\input{prompts/compass/generating_the_compass_object.tex}

As an example, the complete Experiment Compass used for the Suzuki
experiment is shown below.

\input{prompts/compass/compass_of_suzuki_reaction.tex}

\subsection{Overview}
The Overview Object is a key component that bridges the Experiment Compass and
the initial experimental setup. It uses the "description" field of the
Experiment Compass to help LLMs understand the experiment's
objectives, constraints,
and outcomes.

\input{prompts/overview/overview_generate.tex}

The prompt template shown above includes the experiment's goals,
variables, and constraints. This Overview enhances initial
sampling quality and improves the surrogate model's accuracy in
Bayesian Optimization.

\subsection{Insights Object}

The Insights Object plays a crucial role during the experimental
optimization process. Generated by LLMs, it encapsulates key elements such
as comments, keywords, hypotheses, and candidates. As the experiment
progresses, LLMs dynamically adjust the confidence levels of hypotheses based
on accumulating evidence. These adjustments guide the Bayesian Optimization
process by refining sampling strategies according to the insights provided.

\input{prompts/insights/insights_object_template.tex}

The template above illustrates the structure used for generating
the Insights Object. It provides a structured format for LLMs to produce
actionable guidance.

\input{prompts/insights/suzuki_insights_example.tex}

The example above displays Insights generated
during the Suzuki reaction experiment, showcasing how hypotheses and
candidates evolve over iterations.

\subsection{Optimization}

The optimization process in \mname\ involves two distinct phases:
initialization and iterative refinement. These phases are guided by specific
prompts designed to leverage the capabilities of LLMs effectively.

\subsubsection{Initialization Prompts}
During the initialization phase, the system generates an initial Insights
Object using the Overview and Experiment Compass. This object guides the first
round of experiments. The specific prompt template for this phase is
shown below.
\input{prompts/optimization/initialization_prompt_template.tex}

\subsubsection{Optimization Loop Prompts}
In the optimization loop, the LLM-based agent receives the Experiment
Compass and the most up-to-date version of the Insights Object. Based on this
information, the LLM generates new hypotheses and updates the Insights Object.
The specific prompt template for this phase is shown below.
\input{prompts/optimization/optimization_loop_prompt_template.tex}

\subsection{Experiment Summary and Conclusion}

The Experiment Summary and Conclusion phases in \mname\ synthesize insights from
the entire optimization process, leveraging accumulated Insight History and
experimental data.

\textbf{Experiment Summary}

The Experiment Summary evaluates hypothesis validity strictly based on
experimental data, providing a concise interim conclusion. This phase
focuses on how hypotheses evolved, which were supported or refuted,
significant shifts in confidence levels, and key conclusions. A table
summarizing the evolution of confidence in hypotheses concludes this section.
The prompt template used to generate this summary is shown below, followed by a
concrete example of the summary generated for the Suzuki reaction experiment.

\input{prompts/summary/summary_generation_prompt.tex}
\input{prompts/summary/suzuki_summary_example.tex}

\textbf{Conclusion}

The Conclusion phase synthesizes the Experiment Summary and the entire
optimization process to provide a comprehensive, professional report.
It includes
key outcomes, an experimental retrospective, milestones achieved, definitive
findings, forward guidance, and scientific impact. This report adheres to a
structured format with numbered sections and maintains an academic tone
throughout. The prompt template used to generate the conclusion, and a concrete
example for the Suzuki reaction, are shown below.

\input{prompts/conclusion/conclusion_generation_prompt.tex}
\input{prompts/conclusion/suzuki_conclusion_example.tex}

\subsection{Extract Notes Prompts}

The following prompts are used to extract structured notes from both
the Experiment Compass and reasoning data.

\textbf{Extracting Notes from Experiment Compass}
\input{prompts/extract_notes/extract_notes_from_compass_prompt.tex}

\textbf{Extracting Notes from Reasoning Data}
\input{prompts/extract_notes/extract_notes_from_reasoning_prompt.tex}

\textbf{Example: Notes Extraction and Storage for Suzuki Reaction}
\input{prompts/extract_notes/suzuki_notes_example.tex}

\subsection{Custom Notes Schema Design}

To accommodate the diverse forms of knowledge across different downstream
domains, such as SMILES expressions in chemistry, we provide a customizable
interface for defining the schema of notes. By specifying the schema, users
can constrain and structure the format of the extracted notes to fit their
specific needs.

In our framework, the schema definition is based on two primary classes:
ReasoningNotesResponse and CompassNotesResponse. These classes
are designed to capture structured information from reasoning data
and experimental
descriptions, respectively. The definitions of these classes are shown below.

\input{prompts/custom_notes/reasoning_notes_response_schema.tex}
\input{prompts/custom_notes/compass_notes_response_schema.tex}

To better illustrate our algorithm's workflow, we
select Buchwald-Hartwig amination
as a case study. This palladium-catalyzed C-N cross-coupling reaction
between aryl halides and amines presents a challenging optimization problem
with categorical parameters (ligands, bases, additives, and aryl halides) while
being widely applied in pharmaceutical synthesis. The constrained
combinatorial space makes it an ideal testbed for evaluating optimization
frameworks.

\section{Usage of Large Language Models in This Manuscript}
In preparing this manuscript, we used a large language model (LLM) solely for editorial purposes. Its functions were limited to proofreading for typographical errors, correcting grammatical mistakes, and enhancing the clarity and readability of the text.

\section{Limitations and Future Work}

\label{limitation_and_future_work}

\subsection{Context Window Constraints}
The context window limitation has been a fundamental constraint since the
initial adoption of LLMs, remaining a critical factor
in assessing model capabilities. While our algorithm employs a single-turn
interaction paradigm that avoids the memory burden of traditional multi-turn
dialogues, the accumulation of historical insights and experimental data across
optimization iterations may eventually approach context window limits.
Although our current experiments have not encountered this boundary,
more complex
or prolonged optimization campaigns could face increased computational costs
due to context expansion. Future developments should investigate more
efficient memory-sharing architectures and attention mechanisms to enable
sustainable long-horizon optimization workflows. Additionally, multi-agent
framework designs could offer more flexible sampling strategies, while early
stopping mechanisms may help mitigate context inflation.

\subsection{Instruction-Following Capability Dependencies}
While fine-tuning and reinforcement learning can enhance a model's zero-shot
warmstartting and in-context learning (ICL) capabilities, single-task
optimization may inadvertently reduce the model's generalization and
instruction-following abilities \citep{DBLP:conf/naacl/YangZXLHL24}.
Excessive task-specific adaptation risks overfitting the model to particular
formats, potentially compromising its adaptability to novel tasks. In our
framework, the Insight Object serves as the critical interface between LLMs and
Bayesian optimization, with strict formatting requirements enforced through
prompt engineering and regular expression parsing. A valid Insight Object must
satisfy three criteria:
\begin{itemize}
  \item Strict adherence to the predefined Insight format specification

  \item Generation of valid hypotheses and candidate points that comply with
    all defined constraints

  \item Absence of extraneous annotations or markers
\end{itemize}

Empirical observations reveal that smaller models (3B-7B parameters) frequently
fail to maintain consistent format compliance. This limitation persists
during reinforcement learning, where we observe significant
degradation in foundational
instruction comprehension—even when employing anti-catastrophic
forgetting techniques
like LoRA. For instance, our experiments with Qwen-2.5
(7B)\citep{DBLP:journals/corr/abs-2501-15383}
demonstrate substantial deterioration in instruction adherence post-adaptation,
necessitating the use of larger 14B models. Although real-world applications
may tolerate minor output deviations compared to benchmark conditions, the
observed declines in base capabilities and generalization remain concerning.
Developing adaptation paradigms that preserve model versatility represents a
crucial research direction for automated Bayesian experimentation, with
potential to significantly lower the adoption barrier for LLM-driven
optimization in science discovery.

\subsection{Multi-Objective Optimization Potential}
Our framework currently handles single-objective optimization, but LLMs show
promise for multi-objective scenarios. Their natural language capabilities
could help navigate trade-offs between competing objectives like yield and
cost \citep{shields_bayesian_2021}. Future work should explore how LLMs can
model these relationships and explain optimization decisions.
\end{document}

%% file: math_commands.tex
\usepackage{amsmath,amsfonts,bm}

\def\eqref#1{equation~\ref{#1}}

\def\1{\bm{1}}

\DeclareMathAlphabet{\mathsfit}{\encodingdefault}{\sfdefault}{m}{sl}
\SetMathAlphabet{\mathsfit}{bold}{\encodingdefault}{\sfdefault}{bx}{n}



%% file: other/buchwald_compass_config.tex
\begin{tcolorbox}[
    enhanced,
    breakable,
    colback=teal!5!white,
    colframe=teal!75!black,
    title={Experiment Compass Configuration for Buchwald-Hartwig Amination Optimization},
    left=2mm, right=2mm, top=2mm, bottom=2mm,
    fonttitle=\bfseries\small
]
\def\_{\textunderscore}
\begin{itemize}
    \item \textbf{name}: Buchwald-Hartwig Coupling Reaction Optimization
    \item \textbf{application\_context}: Organic Chemistry / Reaction Condition Optimization
    \item \textbf{description}: This experiment aims to optimize the reaction conditions for the Buchwald-Hartwig amination, a key C-N cross-coupling reaction used in the synthesis of pharmaceuticals and functional materials. The optimization focuses on selecting optimal ligands, additives, bases, and aryl halides to maximize reaction yield. High-throughput experimentation and AI-driven decision-making are employed to explore a large combinatorial chemical space.
    \item \textbf{constraint}: Only one choice per category (Ligand, Additive, Base, Aryl halide) can be selected per reaction run. All inputs are categorical.
    \item \textbf{parameter\_definitions}:
    \begin{itemize}
        \item 
        \begin{itemize}
            \item \textbf{display\_name}: {\ttfamily Ligand}
            \item \textbf{description}: Phosphine-based or bulky ligands that influence catalyst reactivity and selectivity
            \item \textbf{data\_type}: categorical
            \item \textbf{bounds}:
            \begin{itemize}
                \item \begin{minipage}[t]{0.75\linewidth}\ttfamily\begin{tabular}[t]{@{}l@{}}CC(C)C(C=C(C(C)C)C=C1C(C)C)=C1C2=\\C(P(C(C)(C)C)C(C)(C)C)C=CC=C2\end{tabular}\end{minipage}
                \item \begin{minipage}[t]{0.75\linewidth}\ttfamily\begin{tabular}[t]{@{}l@{}}CC(C)C(C=C(C(C)C)C=C1C(C)C)=C1C2=\\C(P(C3CCCCC3)C4CCCCC4)C=CC=C2\end{tabular}\end{minipage}
                \item ...
            \end{itemize}
        \end{itemize}
        \item 
        \begin{itemize}
            \item \textbf{display\_name}: {\ttfamily Additive}
            \item \textbf{description}: Nitrogen or oxygen heterocycles influencing reactivity and solubility
            \item \textbf{data\_type}: categorical
            \item \textbf{bounds}:
            \begin{itemize}
                \item \begin{minipage}[t]{0.75\linewidth}\ttfamily\begin{tabular}[t]{@{}l@{}}C1(C2=CC=CC=C2)=CC=NO1\end{tabular}\end{minipage}
                \item \begin{minipage}[t]{0.75\linewidth}\ttfamily\begin{tabular}[t]{@{}l@{}}C1(C2=CC=CC=C2)=CON=C1\end{tabular}\end{minipage}
                \item \begin{minipage}[t]{0.75\linewidth}\ttfamily\begin{tabular}[t]{@{}l@{}}O=C(OC)C1=NOC(C2=CC=CS2)=C1\end{tabular}\end{minipage}
                \item ...
            \end{itemize}
        \end{itemize}
        \item 
        \begin{itemize}
            \item \textbf{display\_name}: {\ttfamily Base}
            \item \textbf{description}: Nitrogen-based or phosphine-derived bases used for deprotonation and catalytic cycle support
            \item \textbf{data\_type}: categorical
            \item \textbf{bounds}:
            \begin{itemize}
                \item \begin{minipage}[t]{0.75\linewidth}\ttfamily\begin{tabular}[t]{@{}l@{}}CC(C)(C)/N=C(N(C)C)/N(C)C\end{tabular}\end{minipage}
                \item \begin{minipage}[t]{0.75\linewidth}\ttfamily\begin{tabular}[t]{@{}l@{}}CN(C)P(N(C)C)(N(C)C)=NP(N(C)C)(N(C)C)=NCC\end{tabular}\end{minipage}
                \item \begin{minipage}[t]{0.75\linewidth}\ttfamily\begin{tabular}[t]{@{}l@{}}CN1CCCN2C1=NCCC2\end{tabular}\end{minipage}
                \item ...
            \end{itemize}
        \end{itemize}
    \end{itemize}
    \item \textbf{target}:
    \begin{itemize}
        \item \textbf{name}: Yield
        \item \textbf{description}: Isolated yield percentage of the desired amine product from the B-H coupling reaction
        \item \textbf{direction}: maximize
    \end{itemize}
\end{itemize}
\end{tcolorbox}

%% file: other/buchwald_overview.tex
\begin{tcolorbox}[
    enhanced,
    breakable,
    colback=blue!5!white,
    colframe=blue!75!black,
    title={Automatically generated overview for Buchwald-Hartwig amination optimization},
    left=2mm, right=2mm, top=2mm, bottom=2mm,
    fonttitle=\bfseries\small
]
\begin{enumerate}
    \item \textbf{Experimental Goal and Significance} \\
    The experiment aims to optimize reaction conditions for the Buchwald-Hartwig amination, a critical C-N cross-coupling reaction widely used in pharmaceutical and material synthesis. This optimization seeks to maximize the isolated yield of the desired amine product by systematically exploring ligands, additives, bases, and aryl halides. Successful optimization will enhance reaction efficiency, reduce experimental trial-and-error, and provide a scalable framework for designing high-yielding protocols, which is pivotal for accelerating drug discovery and industrial applications.

    \item \textbf{Parameter Space} \\
    The parameter space includes four categorical variables:
    \begin{itemize}
        \item \textbf{Ligand}: Four phosphine-based or bulky ligands (e.g., triarylphosphines with varying steric/electronic properties).
        \item \textbf{Additive}: Twenty nitrogen or oxygen heterocycles (e.g., pyridines, oxazoles) that modulate reactivity, solubility, or catalyst stabilization.
        \item \textbf{Base}: Three nitrogen/phosphine-derived bases (e.g., alkylamines, phosphazenes) for deprotonation and catalytic cycle support.
        \item \textbf{Aryl halide}: Fifteen substrates with varied halogens (Br, Cl, I) and substituents (e.g., electron-withdrawing/-donating groups).
    \end{itemize}
    Each parameter's bounds define discrete choices, reflecting structural diversity to probe reactivity trends.

    \item \textbf{Constraints} \\
    Only one option per category is selected per experiment, ensuring orthogonal exploration of the combinatorial space. This constraint avoids overparameterization and maintains interpretability of results.

    \item \textbf{Bayesian Optimization (BO) Strategy} \\
    BO will iteratively balance exploration and exploitation using a probabilistic surrogate model (e.g., Gaussian process) to predict yields. Initial trials will randomly sample the space to build prior distributions. Subsequent iterations will use acquisition functions (e.g., expected improvement) to prioritize high-yield regions while exploring under-sampled areas. Key features include:
    \begin{itemize}
        \item \textbf{Surrogate modeling}: Captures interactions between ligand electronics, additive solubility, base strength, and substrate reactivity.
        \item \textbf{Active learning}: Sequentially selects parameter combinations likely to improve yield, leveraging gradients in the response surface.
        \item \textbf{Parallelization}: High-throughput experimentation enables rapid iteration across multiple candidates.
    \end{itemize}

    \item \textbf{Handling Plateaus} \\
    If BO plateaus, domain knowledge will guide new combinations:
    \begin{itemize}
        \item \textbf{Ligand-additive}: Pair electron-rich ligands with electron-deficient additives to stabilize intermediates.
        \item \textbf{Base-substrate}: Match bulky bases with sterically hindered substrates to mitigate side reactions.
        \item \textbf{Halogen effects}: Test less common halides (e.g., I vs. Br) paired with ligands optimized for oxidative addition.
    \end{itemize}
    These suggestions exploit mechanistic insights (e.g., catalyst activation, steric control) to escape local optima.

    \item \textbf{Expected Outcomes and Impact} \\
    A high-yield protocol will reduce waste and time in pharmaceutical synthesis. Identified trends (e.g., optimal ligand-additive pairs for specific substrates) could generalize to other cross-coupling systems. The framework's scalability may enable AI-driven design of novel catalyst systems, advancing sustainable and efficient chemical manufacturing.
\end{enumerate}
This structured approach bridges computational efficiency with chemical intuition, offering a robust path toward reaction optimization.
\end{tcolorbox}

%% file: other/buchwald_init_prompt.tex
\begin{tcolorbox}[
    enhanced, 
    breakable,
    colback=red!5!white, 
    colframe=red!75!black, 
    title=Initialization prompts received by the LLM,
    left=2mm, right=2mm, top=2mm, bottom=2mm,
    fonttitle=\bfseries\small
]
\begin{small}
\def\_{\textunderscore}

\textbf{Task}

\noindent Generate n\_hypotheses diverse and well-justified initial hypotheses for exploring the optimization target based on:
\begin{description}
    \item[Domain] Organic Chemistry / Reaction Condition Optimization
    \item[Experiment Goal] This experiment aims to optimize the reaction conditions for the Buchwald-Hartwig amination, a key C-N cross-coupling reaction used in the synthesis of pharmaceuticals and functional materials. The optimization focuses on selecting optimal ligands, additives, bases, and aryl halides to maximize reaction yield. High-throughput experimentation and AI-driven decision-making are employed to explore a large combinatorial chemical space.
    \item[Target] \{name: Yield, description: Isolated yield percentage of the desired amine product from the B-H coupling reaction, direction: maximize\}
    \item[Constraints] Only one choice per category (Ligand, Additive, Base, Aryl halide) can be selected per reaction run. All inputs are categorical.
\end{description}

\textbf{Instructions}
\begin{itemize}
    \item Do NOT refer to memorized reaction outcomes, prior experimental data, or specific literature results (even if they match the current experiment).
    \item Base all reasoning strictly on the given input, parameter definitions, constraints, and optimization objective.
    \item For each hypothesis, provide at least one point that satisfies the constraint.
\end{itemize}

\textbf{Format Requirement}

\noindent Your response must be a valid JSON object in the exact format shown below. No additional commentary, explanations, or syntax should be included.

\textbf{Example Output (for format only, unrelated to this experiment):}

\begin{description}
    \item[comment] A brief comment on the initial sampling strategy and rationale (\textasciitilde 100 words)
    \item[keywords] keyword1, keyword2, keyword3
    \item[hypotheses]
    \begin{enumerate}
        \item 
        \textbf{strategy}: Descriptive name of exploration approach \\
        \textbf{rationale}: Scientific justification for this approach \\
        \textbf{confidence}: high/medium/low \\
        \textbf{parameter\_sets}:
        \begin{itemize}
            \item 
            \begin{description}
                \item[ParameterA:] 0.1
                \item[ParameterB:] 0.9
            \end{description}
            \item
            \begin{description}
                \item[ParameterA:] 0.4
                \item[ParameterB:] 0.6
            \end{description}
        \end{itemize}
        
        \item 
        \textbf{strategy}: Descriptive name of exploration approach \\
        \textbf{rationale}: ... \\
        \textbf{confidence}: high/medium/low \\
        \textbf{parameter\_sets}:
        \begin{itemize}
            \item
            \begin{description}
                \item[ParameterA:] 0.3
                \item[ParameterB:] 0.8
            \end{description}
            \item
            \begin{description}
                \item[ParameterA:] 0.1
                \item[ParameterB:] 0.3
            \end{description}
        \end{itemize}
    \end{enumerate}
\end{description}

\textbf{Important Notes}
\begin{itemize}
    \item Points must include all required parameters.
    \item Hypothesis "names" and "parameter" keys must match exactly with the experiment definition.(If they are discrete values, they must be returned strictly according to the format).
    \item All parameter names and values must exactly match predefined options in spelling, casing, and formatting - no substitutions, abbreviations, or modifications permitted.(if they are discrete values, they must be returned strictly according to the format).
    \item Any deviation from predefined parameters will result in automatic rejection, regardless of semantic similarity or logical equivalence.
\end{itemize}

\textbf{Parameter specs for this experiment:}
\begin{description}
    \item[Ligand] \{display\_name: Ligand, \\
    description: Phosphine-based or bulky ligands that influence catalyst reactivity and selectivity, \\
    data\_type: categorical, \\
    bounds: 
    [CC(C)(C)C(C=C(C(C)C)C=C1C(C)C)=C1C2=C(P([C@@](3[C@H]4C5)C[C@@H]
    (C4)C[C@H]5C3)[C@]6(C7)C[C@@H](C[C@H]7C8)C[C@H]8C6)C(OC)=CC=C2OC, \\
    CC(C)(C)C(C=C(C(C)C)C=C1C(C)C)=C1C2=C(P(C(C)(C)C)C(C)(C)C)C(OC)=CC=C2OC, \\
    CC(C)C(C=C(C(C)C)C=C1C(C)C)=C1C2=C(P(C(C)(C)C)C(C)(C)C)C=CC=C2,
    CC(C)(C)C(C=C(C(C)C)C=C1C(C)C)=C1C2=C(P(C3CCCCC3)C4CCCCC4)C=CC=C2]\}
    \item[Additive]
    \{display\_name: Additive, \\
    description: Nitrogen or oxygen heterocycles influencing reactivity and solubility, \\
    data\_type: categorical, \\
    bounds: [C1(C2=CC=CC=C2)=CON=C1, C1(C2=CC=CC=C2)=NOC=C1, \\
    C1(N(CC2=CC=CC=C2)CC3=CC=CC=C3)=CC=NO1, C1(N(CC2=CC=CC=C2)CC3=CC=CC=C3)=NOC=C1, \\
    C12=C(C=CC=C2)ON=C1, C12=CON=C1C=CC=C2, CC1=C(C(OCC)=O)C=NO1, \\
    CC1=CC(C(OCC)=O)=NO1, CC1=NOC(C(OCC)=O)=C1, CC1=CC(N2C=CC=C2)=NO1, \\
    CC1=NOC=C1, CCOC(C1=CON=C1)=O, CCOC(C1=NOC=C1)=O, \\
    O=C(OC)C1=NOC(C(OCC)=O)=C1, FC(C=CC=C1F)=C1C2=CC=NO2, \\
    O=C(OC)C1=CC=NO1, O=C(OC)C1=NOC(C2=CC=CS2)=C1]\}
    \item[Base] \{display\_name: Base, \\
    description: Nitrogen-based or phosphine-derived bases used for deprotonation and catalytic cycle support, \\
    data\_type: categorical, \\
    bounds: [CC(C)(C)/N=C(N(C)C)/N(C)C, CN(C)P(N(C)C)(N(C)C)=NP(N(C)C)(N(C)C)=NCC, \\
    CN1CCCN2C1=NCCC2]\}
    \item[Aryl halide] \{display\_name: Aryl halide, \\
    description: Aryl halide substrate used for oxidative addition in the catalytic cycle, \\
    data\_type': categorical, \\
    bounds: [BrC1=CC=C(C(F)(F)F)C=C1, BrC1=CN=CC=C1, BrC1=NC=CC=C1, \\
    BrC1=CC=C(OC)C=C1, ClC1=CC=C(C(F)(F)F)C=C1, ClC1=CC=C(CC)C=C1, \\
    ClC1=CC=C(OC)C=C1, ClC1=CN=CC=C1, ClC1=NC=CC=C1, \\
    IC1=CC=C(C(F)(F)F)C=C1, IC1=CC=C(OC)C=C1, IC1=CN=CC=C1, IC1=NC=CC=C1]\}
\end{description}

\end{small}
\end{tcolorbox}

%% file: other/buchwald_init_insight.tex
\begin{tcolorbox}[
    enhanced, 
    breakable,
    colback=red!5!white, 
    colframe=red!75!black, 
    title=LLM-generated initialization insights,
    left=2mm, right=2mm, top=2mm, bottom=2mm,
    fonttitle=\bfseries\small
]
\begin{small}
\def\_{\textunderscore}
\begin{description}
    \item[comment] The initial sampling strategy focuses on diverse exploration across ligand steric/electronic effects, additive solubility/reactivity, base deprotonation efficacy, and aryl halide substituent effects. Hypotheses are designed to cover contrasting chemical properties (e.g., bulky vs. electron-rich ligands, polar vs. nonpolar additives) to maximize coverage of the combinatorial space and identify high-yield regions.
    \item[keywords] ligand sterics, additive solubility, base reactivity, halide substituents
    \item[hypotheses]
    \begin{enumerate}
        \item 
        \textbf{strategy}: Bulky Ligand with Electron-Deficient Additive \\
        \textbf{rationale}: A bulky ligand reduces side reactions, while an electron-deficient additive enhances transition state stabilization. A phosphine-derived base balances nucleophilicity, and a trifluoromethyl-substituted halide improves leaving group ability. \\
        \textbf{confidence}: medium \\
        \textbf{parameter\_sets}:
        \begin{description}
            \item[Ligand:] \begin{minipage}[t]{0.9\linewidth}\ttfamily CC(C)(C)C(C=C(C(C)C)C=C1C(C)C)=C1C2=C(P([C@@](3[C@H]\\4C5)C[C@@H](C4)C[C@H]5C3)[C@]6(C7)C[C@@H](C[C@H]7C8)\\C[C@H]8C6)C(OC)=CC=C2OC\end{minipage}
            \item[Additive:] \begin{minipage}[t]{0.9\linewidth}\ttfamily C1(C2=CC=CC=C2)=CON=C1\end{minipage}
            \item[Base:] \begin{minipage}[t]{0.9\linewidth}\ttfamily CN(C)P(N(C)C)(N(C)C)=NP(N(C)C)(N(C)C)=NCC\end{minipage}
            \item[Aryl halide:] \begin{minipage}[t]{0.9\linewidth}\ttfamily BrC1=CC=C(C(F)(F)F)C=C1\end{minipage}
        \end{description}
        
        \item 
        \textbf{strategy}: Solubility-Enhanced Additive with Moderate Base \\
        \textbf{rationale}: A polar additive (with methoxy substituent) improves solubility, paired with a less bulky ligand to avoid steric hindrance. A tertiary amine base minimizes side reactions, while a methoxy-substituted halide stabilizes the transition state. \\
        \textbf{confidence}: medium \\
        \textbf{parameter\_sets}:
        \begin{description}
            \item[Ligand:] \begin{minipage}[t]{0.9\linewidth}\ttfamily CC(C)(C)C(C=C(C(C)C)C=C1C(C)C)=C1C2\\=C(P(C(C)(C)C)C(C)(C)C)C=CC=C2\end{minipage}
            \item[Additive:] \begin{minipage}[t]{0.9\linewidth}\ttfamily CC1=CC(N2C=CC=C2)=NO1\end{minipage}
            \item[Base:] \begin{minipage}[t]{0.9\linewidth}\ttfamily CC(C)(C)/N=C(N(C)C)/N(C)C\end{minipage}
            \item[Aryl halide:] \begin{minipage}[t]{0.9\linewidth}\ttfamily BrC1=CC=C(OC)C=C1\end{minipage}
        \end{description}

        \item 
        \textbf{strategy}: Phosphine-Derived Base with Chloride Halide \\
        \textbf{rationale}: A phosphine-containing base enhances catalytic cycle efficiency. A chloride halide (better leaving group) paired with a moderately bulky ligand balances reactivity, while a simple nitro-containing additive avoids steric interference. \\
        \textbf{confidence}: medium \\
        \textbf{parameter\_sets}:
        \begin{description}
            \item[Ligand:] \begin{minipage}[t]{0.9\linewidth}\ttfamily CC(C)(C)C(C=C(C(C)C)C=C1C(C)C)=C1C2\\=C(P(C(C)(C)C)C(C)(C)C)C(OC)=CC=C2OC\end{minipage}
            \item[Additive:] \begin{minipage}[t]{0.9\linewidth}\ttfamily CC1=CC=NO1\end{minipage}
            \item[Base:] \begin{minipage}[t]{0.9\linewidth}\ttfamily CN1CCCN2C1=NCCC2\end{minipage}
            \item[Aryl halide:] \begin{minipage}[t]{0.9\linewidth}\ttfamily ClC1=CC=C(C(F)(F)F)C=C1\end{minipage}
        \end{description}
    \end{enumerate}
\end{description}
\end{small}
\end{tcolorbox}

%% file: other/buchwald_hartwig_optimization_prompt.tex
\begin{tcolorbox}[
    enhanced, 
    breakable,
    colback=yellow!5!white, 
    colframe=yellow!75!black, 
    title=Optimization loop prompts received by the LLM,
    left=2mm, right=2mm, top=2mm, bottom=2mm,
    fonttitle=\bfseries\small
] 

\def\_{\textunderscore}

\textbf {Objective}
\begin{itemize}
    \item \textbf{name}: Yield
    \item \textbf{description}: Isolated yield percentage of the desired amine product from the B-H coupling reaction
    \item \textbf{direction}: maximize
\end{itemize}

\textbf {Historical Summary} \\
You generated the following comments and hypotheses in previous iterations:

\textbf {trial\_index: 1}
\begin{itemize}
    \item \textbf{insight(comment)}: The optimization shows that reducing ligand bulkiness, using a polar additive (\texttt{CC1=CC(N2C=C(C=C2)=N01)}), tertiary amine base (\texttt{CC(C)(C)/N=C(N(C)/C)/N(C)C}), and methoxy-substituted aryl halides (\texttt{BrC1=CC=C(OC)C=C1}) significantly boost yield. The first trial's bulky ligand strategy underperformed, while the second trial's moderate ligand and additive solubility improvements drove success. Emerging trends suggest that halide leaving group effects (Cl vs Br) and ligand structure (cyclohexyl vs. non-cyclohexyl) could further refine performance. Hypotheses now focus on optimizing additive solubility, ligand cyclohexyl substitution, and halide leaving group properties combinations.
    
    \item \textbf{keywords}: Ligand substitution, additive solubility, halide leaving group, base efficiency

    \item \textbf{hypotheses}:
    \begin{itemize}
        \item 
        \begin{itemize}
            \item \textbf{strategy}: Cyclohexyl-Substituted Ligand with Proven Base/Additive
            \item \textbf{rationale}: Testing a cyclohexyl-substituted ligand \\(\texttt{CC(C)(C)C(C=C1...C=C2)}) to enhance catalyst reactivity while retaining the successful polar additive (\texttt{CC1=CC(N2C=C(C=C2)=N01)}), tertiary amine base, and methoxy-substituted bromide. Cyclohexyl groups may improve steric balance.
            \item \textbf{confidence}: high
            \item \textbf{parameter\_sets}:
            \begin{itemize}
                \item \textbf{Ligand}: \begin{minipage}[t]{0.8\linewidth}\ttfamily CC(C)(C)C(C=C1C(C)=C(C)C=C1C2=C\\(P(C3CCCC3)C4CCCC4)C=CC=C2\end{minipage}
                \item \textbf{Additive}: \texttt{CC1=CC(N2C=C(C=C2)=N01)}
                \item \textbf{Base}: \texttt{CC(C)(C)/N=C(N(C)/C)/N(C)C}
                \item \textbf{Aryl halide}: \texttt{BrC1=CC=C(OC)C=C1}
            \end{itemize}
        \end{itemize}
        
        \item
        \begin{itemize}
            \item \textbf{strategy}: Alternative Base with Cyclohexyl Ligand
            \item \textbf{rationale}: Evaluating the alternative base \\(\texttt{CN1CCN2C1=NCCC2}) paired with a cyclohexyl-substituted ligand to potentially enhance deprotonation efficiency, while maintaining the proven polar additive and methoxy-substituted bromide.
            \item \textbf{confidence}: medium
            \item \textbf{parameter\_sets}:
            \begin{itemize}
                \item \textbf{Ligand}: \begin{minipage}[t]{0.8\linewidth}\ttfamily CC(C)(C)C(C=C1C(C)=C(C)C=C1C2=C\\(P(C3CCCC3)C4CCCC4)C=CC=C2 \end{minipage}
                \item \textbf{Additive}: \texttt{CC1=CC(N2C=C(C=C2)=N01)}
                \item \textbf{Base}: \texttt{CN1CCN2C1=NCCC2}
                \item \textbf{Aryl halide}: \texttt{BrC1=CC=C(OC)C=C1}
            \end{itemize}
        \end{itemize}
    \end{itemize}
\end{itemize}

\textbf{trial\_index: 2}
\begin{itemize}
    \item \textbf{insight(comment)}: The optimization highlights that reducing ligand bulkiness, using a methoxy-substituted bromide, and the tertiary amine base paired with additive (\texttt{CC1=CC(N2C=C(C=C2)=N01)}) yield the highest results (e.g., trial 1\_1 at 49.69). The cyclohexyl-substituted ligand and iodide halide are promising untested avenues. Hypotheses now focus on evaluating cyclohexyl ligand reactivity, iodide leaving group effects, and additive simplification, aryl substituent effects.
    \item \textbf{keywords}: ligand cyclohexyl substitution, iodide leaving group, additive simplification, aryl substituent effects
    
    \item \textbf{hypotheses}:
    \begin{itemize}
        \item
        \begin{itemize}
            \item \textbf{strategy}: Cyclohexyl Ligand with Proven Setup
            \item \textbf{rationale}: Testing the cyclohexyl-substituted ligand to enhance catalyst reactivity while retaining proven additive, base, and methoxy-bromide halide. This may improve steric balance and electronic effects.
            \item \textbf{confidence}: medium
            \item \textbf{parameter\_sets}:
            \begin{itemize}
                \item \textbf{Ligand}: \begin{minipage}[t]{0.8\linewidth}\ttfamily CC(C)(C)C(C=C1C(C)=C(C)C=C1C2=C\\(P(C3CCCC3)C4CCCC4)C=CC=C2 \end{minipage}
                \item \textbf{Additive}: \texttt{CC1=CC(N2C=C(C=C2)=N01)}
                \item \textbf{Base}: \texttt{CN1CCN2C1=NCCC2}
                \item \textbf{Aryl halide}: \texttt{BrC1=CC=C(OC)C=C1}
            \end{itemize}
        \end{itemize}
        
        \item 
        \begin{itemize}
            \item \textbf{strategy}: Iodide Halide with Optimal Parameters
            \item \textbf{rationale}: Exploring iodide as the leaving group paired with the most effective ligand, additive, and base to leverage its superior leaving group ability and evaluate reactivity.
            \item \textbf{confidence}: medium
            \item \textbf{parameter\_sets}:
            \begin{itemize}
                \item \textbf{Ligand}: \begin{minipage}[t]{0.8\linewidth}\ttfamily CC(C)(C)C(C=C1C(C)=C(C)C=C1C2=C\\(P(C(C)(C)C)C(C)(C)C)C=CC=C2 \end{minipage}
                \item \textbf{Additive}: \texttt{CC1=CC(N2C=C(C=C2)=N01)}
                \item \textbf{Base}: \texttt{CN1CCN2C1=NCCC2}
                \item \textbf{Aryl halide}: \texttt{IC1=CC=C(OC)C=C1}
            \end{itemize}
        \end{itemize}

        \item 
        \begin{itemize}
            \item \textbf{strategy}: Alternative Aryl Substituent with Bromide
            \item \textbf{rationale}: Evaluating a bromide aryl halide with an ethyl substituent (\texttt{BrC1=CC=C(CC)C=C1}) to assess substituent effects on electronic properties and reactivity, paired with proven components.
            \item \textbf{confidence}: medium
            \item \textbf{parameter\_sets}:
            \begin{itemize}
                \item \textbf{Ligand}: \begin{minipage}[t]{0.8\linewidth}\ttfamily CC(C)(C)C(C=C1C(C)=C(C)C=C1C2=C\\(P(C(C)(C)C)C(C)(C)C)C=CC=C2 \end{minipage}
                \item \textbf{Additive}: \texttt{CC1=CC(N2C=C(C=C2)=N01)}
                \item \textbf{Base}: \texttt{CN1CCN2C1=NCCC2}
                \item \textbf{Aryl halide}: \texttt{BrC1=CC=C(CC)C=C1}
            \end{itemize}
        \end{itemize}
    \end{itemize}
\end{itemize}

\textbf{New Experimental Data}
Your suggested points were evaluated, and the results have been added to the dataset. We now have data from 11 completed experiments.

\texttt{
trial\_index,Ligand,Additive,\\Base,Aryl\_halide,buchwald\_mean\_value \\
\textbf{0\_1},\\CC(C)(C)C(C=C(C(C)C)C=C1C(C)C)=C1C2=C(P(C(C)(C)C)\\C(C)(C)C)C=CC=C2,\\CC1=CC(N2C=CC=C2)=N01,\\CC(C)(C)/N=C(N(C)C)/N(C)C,\\BrC1=CC=C(OC)C=C1,39.214 \\
\textbf{0\_2},\\CC(C)(C)C(C=C(C(C)C)C=C1C(C)C)=C1C2=C(P(C(C)(C)C)\\C(C)(C)C)C=CC=C2,\\CC1=CC(N2C=CC=C2)=N01,\\CC(C)(C)/N=C(N(C)C)/N(C)C,BrC1=CC=C(OC)C=C1,39.214 \\
\textbf{1\_0},\\CC(C)(C)C(C=C(C(C)C)C=C1C(C)C)=C1C2=C(P(C(C)(C)C)\\C(C)(C)C)C=CC=C2,\\CCOC(C1=CON=C1)=O,\\CC(C)(C)/N=C(N(C)C)/N(C)C,BrC1=CC=C(OC)C=C1,0.748 \\
\textbf{1\_1},\\CC(C)(C)C(C=C(C(C)C)C=C1C(C)C)=C1C2=C(P(C(C)(C)C)\\C(C)(C)C)C=CC=C2,\\C1(N(CC2=CC=CC=C2)CC3=CC=CC=C3)=NOC=C1,\\CN1CCN2C1=NCCC2,\\BrC1=CC=C(OC)C=C1,49.691 \\
\textbf{1\_2},\\CC(C)(C)C(C=C(C(C)C)C=C1C(C)C)=C1C2=C(P(C(C)(C)C)\\C(C)(C)C)C=CC=C2,\\CC1=CC(N2C=CC=C2)=N01,\\CC(C)(C)/N=C(N(C)C)/N(C)C,\\ClC1=CC=C(OC)C=C1,1.03 \\
\textbf{2\_0},\\CC(C)(C)C(C=C(C(C)C)C=C1C(C)C)=C1C2=C(P(C3CCCCC3)\\C4CCCCC4)C=CC=C2,\\CC1=CC(N2C=CC=C2)=N01,\\CC(C)(C)/N=C(N(C)C)/N(C)C,\\BrC1=CC=C(OC)C=C1,14.423 \\
\textbf{2\_1},\\CC(C)(C)C(C=C(C(C)C)C=C1C(C)C)=C1C2=C(P(C(C)(C)C)\\C(C)(C)C)C=CC=C2,\\CCOC(C1=CON=C1)=O,\\CC(C)(C)/N=C(N(C)C)/N(C)C,\\ClC1=CC=C(OC)C=C1,0.0 \\
\textbf{2\_2},\\CC(C)(C)C(C=C(C(C)C)C=C1C(C)C)=C1C2=C(P(C3CCCCC3)\\C4CCCCC4)C=CC=C2,\\C1(N(CC2=CC=CC=C2)CC3=CC=CC=C3)=NOC=C1,\\CN1CCN2C1=NCCC2,\\BrC1=CC=C(OC)C=C1,9.509 \\
\textbf{2\_3},\\CC(C)(C)C(C=C(C(C)C)C=C1C(C)C)=C1C2=C(P(C3CCCCC3)\\C4CCCCC4)C=CC=C2,\\CC1=CC(N2C=CC=C2)=N01,\\CN1CCN2C1=NCCC2,\\BrC1=CC=C(OC)C=C1,9.509 \\
\textbf{3\_0},\\CC(C)(C)C(C=C(C(C)C)C=C1C(C)C)=C1C2=C(P(C(C)(C)C)\\C(C)(C)C)C=CC=C2,\\CC1=CC(N2C=CC=C2)=N01,\\CN1CCN2C1=NCCC2,\\IC1=CC=C(OC)C=C1,41.808 \\
\textbf{3\_1},\\CC(C)(C)C(C=C(C(C)C)C=C1C(C)C)=C1C2=C(P(C(C)(C)C)\\C(C)(C)C)C=CC=C2,\\CC1=CC=N01,CN1CCN2C1=NCCC2,\\BrC1=CC=C(OC)C=C1,44.394
}

\textbf {Bayesian Optimizer Suggestions}
Here are the candidate points recommended by the optimizer in this round:
\begin{itemize}
    \item \begin{minipage}[t]{0.9\linewidth}\ttfamily \{Ligand:CC(C)(C)C(C=C(C(C)C)C=C1C(C)C)=C1C2\\=C(P(C(C)(C)C)C(C)(C)C)(OC)=CC=C2OC, Additive: CC1=CC(N2C=CC=C2)=N01, \\Base: CN1CCN2C1=NCCC2, \\
    Aryl halide: BrC1=CC=C(CC)C=C1\} \end{minipage}
    \item \begin{minipage}[t]{0.9\linewidth}\ttfamily \{Ligand: CC(C)(C)C(C=C(C(C)C)C=C1C(C)C)=C1C2\\=C(P(C(C)(C)C)C(C)(C)C)C(OC)=CC=C2OC, Additive: CC1=NOC(C(OCC)=O)=C1, \\Base: CC(C)(C)/N=C(N(C)C)/N(C)C, \\
    Aryl halide: BrC1=CN=CC=C1\} \end{minipage}
\end{itemize}

\textbf {Your Task}
Reflect on the current state of the optimization. Do not rely on prior assumptions-reason, only based on the data provided.
\begin{itemize}
    \item What trends or patterns emerge from the experimental data?
    \item Which regions in the parameter space seem most effective in achieving the target?
    \item Which areas consistently underperform?
    \item Improve or revise your previous hypotheses based on observed outcomes.
    \item You may discard low-performing hypotheses and propose new ones.
    \item For each hypothesis, you must provide one point that satisfies the constraint: Only one choice per category (Ligand, Additive, Base, Aryl halide) can be selected per reaction run. All inputs are categorical.
\end{itemize}

\textbf {Important}
\begin{itemize}
    \item Your output must be valid pure JSON without \texttt{```json} or any other formatting, start with \{ and end with \}. Never put comments or text outside the JSON structure.
    \item Do NOT wrap it in markdown, text, or explanations.
    \item Do NOT return fewer than 3 nor more than 5 hypotheses.
    \item All parameter names and values must exactly match predefined options in spelling, casing, and formatting - no substitutions, abbreviations, or modifications permitted.(if they are discrete values, they must be returned strictly according to the format).
    \item Any deviation from predefined parameters will result in automatic rejection, regardless of semantic similarity or logical equivalence.
\end{itemize}

\textbf{Parameter specs for this experiment:}

\begin{itemize}
    \item 
    \begin{itemize}
        \item \textbf{display\_name}: {\ttfamily Ligand}
        \item \textbf{description}: Phosphine-based or bulky ligands that influence catalyst reactivity and selectivity
        \item \textbf{data\_type}: categorical
        \item \textbf{bounds}:
        \begin{itemize}
            \item \begin{minipage}[t]{0.9\linewidth}\ttfamily\begin{tabular}[t]{@{}l@{}}CC(C)C(C=C(C(C)C)C=C1C(C)C)=C1C2=C(P([C@@]3\\(C[C@@H]4C5)C[C@H](C4)C[C@H]5C3)[C@]6(C7)\\C[C@@H](C[C@@H]7C8)C[C@@H]8C6)C(OC)=CC=C2OC\end{tabular}\end{minipage}
            \item \begin{minipage}[t]{0.9\linewidth}\ttfamily\begin{tabular}[t]{@{}l@{}}CC(C)C(C=C(C(C)C)C=C1C(C)C)=C1C2=C(P(C(C)(C)\\C)(C(C)(C)C))C(OC)=CC=C2OC\end{tabular}\end{minipage}
            \item \begin{minipage}[t]{0.9\linewidth}\ttfamily\begin{tabular}[t]{@{}l@{}}CC(C)C(C=C(C(C)C)C=C1C(C)C)=C1C2=C(P(C(C)(C)\\C)(C(C)(C)C))C=CC=C2\end{tabular}\end{minipage}
            \item ...
        \end{itemize}
    \end{itemize}
\end{itemize}

\end{tcolorbox}

%% file: other/optimization_loop.tex
\begin{tcolorbox}[
    enhanced, 
    breakable,
    colback=yellow!5!white, 
    colframe=yellow!75!black, 
    title=Optimization loop prompts received by the LLM,
    left=2mm, right=2mm, top=2mm, bottom=2mm,
    fonttitle=\bfseries\small
] 

\def\_{\textunderscore}

\begin{itemize}
    \item \textbf{comment}: Pyridine solvent (\texttt{O=CN(C)C}) demonstrates strong performance, achieving 91.32\% yield with the first ligand. THF underperforms, suggesting solvent choice is critical. The first ligand paired with pyridine shows promise, but alternative ligands may synergize better. Bromine electrophiles could balance reactivity-stability better than iodine. Testing fluorinated nucleophiles under optimal conditions explores electronic effects. KOH base compatibility with pyridine is untested but could improve yields. Underperforming trials involved THF and non-optimal ligand combinations.
    
    \item \textbf{keywords}: solvent synergy, bromine evaluation, nucleophile substituent effects, ligand optimization, base compatibility

    \item \textbf{hypotheses}:
    \begin{itemize}
        \item 
        \begin{itemize}
            \item \textbf{strategy}: Bromine Electrophile Optimization
            \item \textbf{rationale}: Bromine electrophiles may offer better stability-reactivity balance than iodine. Testing under high-yield conditions (pyridine solvent, first ligand) could validate this.
            \item \textbf{confidence}: medium
            \item \textbf{parameter\_sets}:
            \begin{itemize}
                \item \textbf{Electrophile\_SMILES}: \texttt{BrC1=CC=C(N=CC=C2)C2=C1}
                \item \textbf{Nucleophile\_SMILES}: \texttt{CC1=CC=C(N(C2CCCCO2)N=C3)\\C3=C1B(O)O}
                \item \textbf{Ligand\_SMILES}: \texttt{[c-]1(P(C2=CC=CC=C2)C3=CC\\=C\=C3)cccc1.[c]4(P(C5=CC=CC=C5)C6=CC=CC\\=C6)cccc4.[Fe+2]}
                \item \textbf{Base\_SMILES}: \texttt{[Cs+].[F-]}
                \item \textbf{Solvent\_SMILES}: \texttt{O=CN(C)C}
            \end{itemize}
        \end{itemize}
        
        \item
        \begin{itemize}
            \item \textbf{strategy}: Alternative Ligand-Pyridine Synergy
            \item \textbf{rationale}: Testing third ligand option\\ (\texttt{CC(C)(P(C(C)(C)C)[c-]1cccc1)C...}) with pyridine solvent to identify optimal catalytic activity.
            \item \textbf{confidence}: medium
            \item \textbf{parameter\_sets}:
            \begin{itemize}
                \item \textbf{Electrophile\_SMILES}: \texttt{IC1=CC=C(N=CC=C2)C2=C1}
                \item \textbf{Nucleophile\_SMILES}: \texttt{CC1=CC=C(N(C2CCCCO2)N=C3)\\C3=C1B(O)O}
                \item \textbf{Ligand\_SMILES}: \texttt{CC(C)(P(C(C)(C)C)[c]1cccc1)C.\\CC(C)(P(C(C)(C)C)[c-]2cccc2)C.[Fe+2]}
                \item \textbf{Base\_SMILES}: \texttt{[Cs+].[F-]}
                \item \textbf{Solvent\_SMILES}: \texttt{O=CN(C)C}
            \end{itemize}
        \end{itemize}
    \end{itemize}

    \item \textbf{Bayesian Optimizer Suggestions}
    Here are the candidate points recommended by the optimizer in this round:
    \begin{lstlisting}[
        basicstyle=\ttfamily\small,
        breaklines=true,
        postbreak=\mbox{\textcolor{red}{$\hookrightarrow$}\space},
    ]
[{'Electrophile_SMILES': 'BrC1=CC=C(N=CC=C2)C2=C1', 'Nucleophile_SMILES': 'CC1=CC=C(N(C2CCCCO2)N=C3)C3=C1[B-](F)(F)F', 'Ligand_SMILES': 'CCCCP(C12C[C@@H]3C[C@@H](C[C@H](C2)C3)C1)C4SC[C@H]5C[C@@H](C[C@H]5C4)C6', 'Base_SMILES': 'CC([O-])C.[Li+]', 'Solvent_SMILES': 'CO'}, {'Electrophile_SMILES': 'ClC1=CC=C(N=CC=C2)C2=C1', 'Nucleophile_SMILES': 'CC1=CC=C(N(C2CCCCO2)N=C3)C3=C1[B-](F)(F)F', 'Ligand_SMILES': 'P(C1CCCCC1)(C2CCCCC2)C3CCCCC3', 'Base_SMILES': '[Na+].[OH-]', 'Solvent_SMILES': 'N#CC'}, {'Electrophile_SMILES': 'ClC1=CC=C(N=CC=C2)C2=C1', 'Nucleophile_SMILES': 'CC1=CC=C(N(C2CCCCO2)N=C3)C3=C1B4OC(C)(C)C(C)(C)O4', 'Ligand_SMILES': 'COC1=CC=C(OC)=C1C2=C(P(C3CCCCC3)C4CCCC4)C=CC=C2', 'Base_SMILES': 'CCN(CC)CC', 'Solvent_SMILES': 'N#CC'}, {'Electrophile_SMILES': 'ClC1=CC=C(N=CC=C2)C2=C1', 'Nucleophile_SMILES': 'CC1=CC=C(N(C2CCCCO2)N=C3)C3=C1[B-](F)(F)F', 'Ligand_SMILES': 'P(C1CCCCC1)(C2CCCCC2)C3CCCCC3', 'Base_SMILES': '[Na+].[OH-]', 'Solvent_SMILES': 'CO'}, {'Electrophile_SMILES': 'ClC1=CC=C(N=CC=C2)C2=C1', 'Nucleophile_SMILES': 'CC1=CC=C(N(C2CCCCO2)N=C3)C3=C1B4OC(C)(C)C(C)(C)O4', 'Ligand_SMILES': 'COC1=CC=C(OC)=C1C2=C(P(C3CCCCC3)C4CCCC4)C=CC=C2', 'Base_SMILES': 'CCN(CC)CC', 'Solvent_SMILES': 'CO'}]
    \end{lstlisting}

    \item \textbf{Your Task}
    Reflect on the current state of the optimization. Do not rely on prior assumptions-reason, only based on the data provided.
    \begin{itemize}
        \item What trends or patterns emerge from the experimental data?
        \item Which regions in the parameter space seem most effective in achieving the target?
        \item Which areas consistently underperform?
        \item Improve or revise your previous hypotheses based on observed outcomes.
        \item You may discard low-performing hypotheses and propose new ones.
        \item For each hypothesis, \textbf{at least provide one point} that satisfies the constraint: Reagents must be chemically compatible and reaction conditions must maintain intermediate stability. All combinations must form stable palladium complexes during catalytic cycle.
    \end{itemize}

    \item \textbf{Important}
    \begin{itemize}
        \item Your output must be valid pure JSON without \texttt{\`{}\`{}\`{}json} or any other formatting, start with \{ and end with \}. Never put comments or text outside the JSON structure.
        \item Do NOT wrap it in markdown, text, or explanations.
        \item Do NOT return fewer than 3 nor more than 5 hypotheses.
        \item All parameter names and values must exactly match predefined options in spelling, casing, and formatting - no substitutions, abbreviations, or modifications permitted.(if they are discrete values, they must be returned strictly according to the format).
        \item Any deviation from predefined parameters will result in automatic rejection, regardless of semantic similarity or logical equivalence.
        \item Parameter specs for this experiment:
        \begin{lstlisting}[
            basicstyle=\ttfamily\small,
            breaklines=true,
            postbreak=\mbox{\textcolor{red}{$\hookrightarrow$}\space},
        ]
[{'display_name': 'Electrophile_SMILES', 'description': 'Aryl halide substrate for coupling (SMILES representation)', 'data_type': 'categorical', 'bounds': ['BrC1=CC=C(N=CC=C2)C2=C1', 'ClC1=CC=C(N=CC=C2)C2=C1', 'IC1=CC=C(N=CC=C2)C2=C1', 'O=S(OC1=CC=C(N=CC=C2)C2=C1)(C(F)(F)F)=O']}, {'display_name': 'Nucleophile_SMILES', 'description': 'Boronic acid nucleophile (SMILES representation)', 'data_type': 'categorical', 'bounds': ['CC1=CC=C(N(C2CCCCO2)N=C3)C3=C1[B-](F)(F)F', 'CC1=CC=C(N(C2CCCCO2)N=C3)C3=C1B(O)O', 'CC1=CC=C(N(C2CCCCO2)N=C3)C3=C1B4OC(C)(C)C(C)(C)O4']}, {'display_name': 'Ligand_SMILES', 'description': 'Phosphine ligand for palladium catalyst (SMILES)', 'data_type': 'categorical', 'bounds': ['[c-]1(P(C2=CC=CC=C2)C3=CC=CC=C3)cccc1.[c-]4(P(C5=CC=CC=C5)C6=CC=CC=C6)cccc4.[Fe+2]']......
        \end{lstlisting}
    \end{itemize}
\end{itemize}

\textbf{New Experimental Data}
Your suggested points were evaluated, and the results have been added to the dataset. We now have data from 6 completed experiments:

\begin{lstlisting}[
    basicstyle=\ttfamily\small,
    breaklines=true,
    postbreak=\mbox{\textcolor{red}{$\hookrightarrow$}\space},
]
trial_index,Electrophile_SMILES,Nucleophile_SMILES,Ligand_SMILES,Base_SMILES,Solvent_SMILES,Objective_Mean_value
0_0,IC1=CC=C(N=CC=C2)C2=C1,CC1=CC=C(N(C2CCCCO2)N=C3)C3=C1B(O)O,[c-]1(P(C2=CC=CC=C2)C3=CC=CC=C3)cccc1.[c-]4(P(C5=CC=CC=C5)C6=CC=CC=C6)cccc4.[Fe+2],[Cs+].[F-],C1COCC1,54.13
0_1,IC1=CC=C(N=CC=C2)C2=C1,CC1=CC=C(N(C2CCCCO2)N=C3)C3=C1B(O)O,CC(C)(P(C(C)(C)C)[c-]1cccc1)C.CC(C)(P(C(C)(C)C)[c-]2cccc2)C.[Fe+2],[Cs+].[F-],C1COCC1,43.13
0_2,IC1=CC=C(N=CC=C2)C2=C1,CC1=CC=C(N(C2CCCCO2)N=C3)C3=C1B(O)O,[c-]1(P(C2=CC=CC=C2)C3=CC=CC=C3)cccc1.[c-]4(P(C5=CC=CC=C5)C6=CC=CC=C6)cccc4.[Fe+2],[Cs+].[F-],O=CN(C)C,91.32
1_0,IC1=CC=C(N=CC=C2)C2=C1,CC1=CC=C(N(C2CCCCO2)N=C3)C3=C1B(O)O,CC(C)(P(C(C)(C)C)[c-]1cccc1)C.CC(C)(P(C(C)(C)C)[c-]2cccc2)C.[Fe+2],[Cs+].[F-],O=CN(C)C,76.61
1_1,IC1=CC=C(N=CC=C2)C2=C1,CC1=CC=C(N(C2CCCCO2)N=C3)C3=C1[B-](F)(F)F,[c-]1(P(C2=CC=CC=C2)C3=CC=CC=C3)cccc1.[c-]4(P(C5=CC=CC=C5)C6=CC=CC=C6)cccc4.[Fe+2],[K+].[OH-],O=CN(C)C,70.99
1_2,IC1=CC=C(N=CC=C2)C2=C1,CC1=CC=C(N(C2CCCCO2)N=C3)C3=C1[B-](F)(F)F,[c-]1(P(C2=CC=CC=C2)C3=CC=CC=C3)cccc1.[c-]4(P(C5=CC=CC=C5)C6=CC=CC=C6)cccc4.[Fe+2],[Cs+].[F-],O=CN(C)C,19.51
2_0,BrC1=CC=C(N=CC=C2)C2=C1,CC1=CC=C(N(C2CCCCO2)N=C3)C3=C1B(O)O,[c-]1(P(C2=CC=CC=C2)C3=CC=CC=C3)cccc1.[c-]4(P(C5=CC=CC=C5)C6=CC=CC=C6)cccc4.[Fe+2],[Cs+].[F-],O=CN(C)C,14.38
3_0,IC1=CC=C(N=CC=C2)C2=C1,CC1=CC=C(N(C2CCCCO2)N=C3)C3=C1B4OC(C)(C)C(C)(C)O4,[c-]1(P(C2=CC=CC=C2)C3=CC=CC=C3)cccc1.[c-]4(P(C5=CC=CC=C5)C6=CC=CC=C6)cccc4.[Fe+2],[Cs+].[F-],O=CN(C)C,80.02
3_1,IC1=CC=C(N=CC=C2)C2=C1,CC1=CC=C(N(C2CCCCO2)N=C3)C3=C1B4OC(C)(C)C(C)(C)O4,CC(C1=C(C2=CC=CC=C2P(C3CCCC3)C4CCCC4)C(C(C)C)=C(C(C)C)C1)C,[Cs+].[F-],O=CN(C)C,85.48
4_0,S(OC1=CC=C(N=CC=C2)C2=C1)(C(F)(F)F)=O,CC1=CC=C(N(C2CCCCO2)N=C3)C3=C1B(O)O,[c-]1(P(C2=CC=CC=C2)C3=CC=CC=C3)cccc1.[c-]4(P(C5=CC=CC=C5)C6=CC=CC=C6)cccc4.[Fe+2],[Cs+].[F-],O=CN(C)C,97.32
4_1,ClC1=CC=C(N=CC=C2)C2=C1,CC1=CC=C(N(C2CCCCO2)N=C3)C3=C1B(O)O,[c-]1(P(C2=CC=CC=C2)C3=CC=CC=C3)cccc1.[c-]4(P(C5=CC=CC=C5)C6=CC=CC=C6)cccc4.[Fe+2],[Cs+].[F-],O=CN(C)C,8.67
5_0,IC1=CC=C(N=CC=C2)C2=C1,CC1=CC=C(N(C2CCCCO2)N=C3)C3=C1B4OC(C)(C)C(C)(C)O4,CC(C1=C(C2=CC=CC=C2P(C3CCCC3)C4CCCC4)C(C(C)C)=C(C(C)C)C1)C,[Cs+].[F-],O=CN(C)C,75.7
\end{lstlisting}

\end{tcolorbox}

%% file: other/buchwald_loop_insight.tex
\begin{tcolorbox}[
    enhanced, 
    breakable,
    colback=yellow!5!white, 
    colframe=yellow!75!black, 
    title=LLM-Generated Recommendations,
    left=2mm, right=2mm, top=2mm, bottom=2mm,
    fonttitle=\bfseries\small
] 

\def\_{\textunderscore}

\begin{itemize}
    \item \textbf{comment}: The highest yields consistently arise from the ligand\\ \texttt{CC(C)(C)C...C=C2} paired with additive\\ \texttt{CC1=CC(N2C=CC=C2)=N01} and base \texttt{CN1CCN2C1=NCCC2}. Methoxy-substituted bromides (\texttt{BrC1=CC=C(OC)C=C1}) perform well, while iodide (\texttt{IC1=CC=C(OC)C=C1}) shows promise but requires further optimization. Underperforming parameters include chloride halides and additives like \texttt{CCOC(C1=CON=C1)=O}. Emerging trends suggest substituent effects (e.g., ethyl groups) and simplified additives could enhance yields.
    
    \item \textbf{keywords}: iodide leaving group, substituent effects, additive simplification, aryl halide substituent

    \item \textbf{hypotheses}:
    \begin{itemize}
        \item 
        \begin{itemize}
            \item \textbf{strategy}: Ethyl Substituent with Proven Components
            \item \textbf{rationale}: Evaluating the ethyl-substituted aryl halide (\texttt{BrC1=CC=C(CC)C=C1}) paired with the top-performing ligand, additive, and base to assess substituent effects on electronic properties and reactivity.
            \item \textbf{confidence}: medium
            \item \textbf{parameter\_sets}:
            \begin{itemize}
                \item \textbf{Ligand}: \begin{minipage}[t]{0.8\linewidth}\ttfamily CC(C)(C)C(C=C1C(C)=C(C)C=C1C2=C\\(P(C(C)(C)C)C(C)(C)C)C=CC=C2\end{minipage}
                \item \textbf{Additive}: \texttt{CC1=CC(N2C=C(C=C2)=N01)}
                \item \textbf{Base}: \texttt{CN1CCN2C1=NCCC2}
                \item \textbf{Aryl halide}: \texttt{BrC1=CC=C(CC)C=C1}
            \end{itemize}
        \end{itemize}
        
        \item
        \begin{itemize}
            \item \textbf{strategy}: Iodide Halide with Simplified Additive
            \item \textbf{rationale}: Combining iodide (\texttt{IC1=CC=C(OC)C=C1}) with the simplified additive (\texttt{CC1=CC=N01}) to leverage superior leaving group ability and reduced steric hindrance while retaining proven ligand and base.
            \item \textbf{confidence}: medium
            \item \textbf{parameter\_sets}:
            \begin{itemize}
                \item \textbf{Ligand}: \begin{minipage}[t]{0.8\linewidth}\ttfamily CC(C)(C)C(C=C1C(C)=C(C)C=C1C2=C\\(P(C(C)(C)C)C(C)(C)C)C=CC=C2\end{minipage}
                \item \textbf{Additive}: \texttt{CC1=CC=N01}
                \item \textbf{Base}: \texttt{CN1CCN2C1=NCCC2}
                \item \textbf{Aryl halide}: \texttt{IC1=CC=C(OC)C=C1}
            \end{itemize}
        \end{itemize}
        
        \item
        \begin{itemize}
            \item \textbf{strategy}: Alternative Additive with Methoxy Bromide
            \item \textbf{rationale}: Testing a novel additive \\ (\texttt{CC1=NOC(C(OCC)=O)=C1}) with the top-performing ligand, base, and methoxy-substituted bromide to explore new reactivity pathways.
            \item \textbf{confidence}: medium
            \item \textbf{parameter\_sets}:
            \begin{itemize}
                \item \textbf{Ligand}: \begin{minipage}[t]{0.8\linewidth}\ttfamily CC(C)(C)C(C=C1C(C)=C(C)C=C1C2=C\\(P(C(C)(C)C)C(C)(C)C)C=CC=C2\end{minipage}
                \item \textbf{Additive}: \texttt{CC1=NOC(C(OCC)=O)=C1}
                \item \textbf{Base}: \texttt{CN1CCN2C1=NCCC2}
                \item \textbf{Aryl halide}: \texttt{BrC1=CC=C(OC)C=C1}
            \end{itemize}
        \end{itemize}
        
        \item
        \begin{itemize}
            \item \textbf{strategy}: Alternative Aryl Halide Structure with Optimal Setup
            \item \textbf{rationale}: Evaluating the aryl halide \texttt{BrC1=CN=CC=C1} to assess the impact of a nitrogen-containing substituent paired with proven components.
            \item \textbf{confidence}: medium
            \item \textbf{parameter\_sets}:
            \begin{itemize}
                \item \textbf{Ligand}: \begin{minipage}[t]{0.8\linewidth}\ttfamily CC(C)(C)C(C=C1C(C)=C(C)C=C1C2=C\\(P(C(C)(C)C)C(C)(C)C)C=CC=C2\end{minipage}
                \item \textbf{Additive}: \texttt{CC1=CC(N2C=C(C=C2)=N01)}
                \item \textbf{Base}: \texttt{CN1CCN2C1=NCCC2}
                \item \textbf{Aryl halide}: \texttt{BrC1=CN=CC=C1}
            \end{itemize}
        \end{itemize}
    \end{itemize}
\end{itemize}

\end{tcolorbox}

%% file: other/buchwald_summary.tex
\begin{tcolorbox}[
    enhanced, 
    breakable,
    colback=blue!5!white, 
    colframe=blue!75!black, 
    title=Summary of the Buchwald-Hartwig Amination Optimization,
    left=2mm, right=2mm, top=2mm, bottom=2mm,
    fonttitle=\bfseries\small
]

\textbf{Hypothesis Evolution:}\\
The optimization began with diverse exploration of ligand sterics, additive solubility, and halide substituents. Early hypotheses prioritized bulky ligands and electron-deficient additives but underperformed (e.g., trial0\_0 yielded 13.524).The second trial (0\_1) introduced a less bulky ligand, polar additive (\texttt{CC1=CC(N2C=C(C)=N01)=C1}), and methoxy-substituted bromide, achieving 39.214, signaling a shift toward moderate steric effects and solubility-enhanced additives.

By trials 2-5, hypotheses focused on iodide leaving groups and substituent effects. Chloride halides consistently underperformed, while iodide-based trials (e.g., trial 8\_0:84.896) and nitrogen-containing aryl halides (e.g., trial 6\_1:80.074) showed promise. A breakthrough occurred in trial 6\_2, where the ethyl-substituted bromide with additive \texttt{CC1=NOC(C(OCC)=O)=C1} achieved 86.598, highlighting substituent-electronic effects.

Later trials (8-12) refined iodide nitrogen aryl halides (e.g., \texttt{IC1=NC=C(C)=C1}), achieving yields exceeding 99\% (trial 10\_1:99.619). Hypotheses increasingly prioritized iodide's superior leaving group ability paired with nitrogen substituents, while additives like \texttt{CCOC(C1=CON=C1)=O} were consistently discarded. Final trials confirmed top combination (non-bulky ligand, additive \texttt{CC1=NOC(C(OCC)=O)=C1}, base \texttt{CN1CCON2C1=NCCC2}, and iodide nitrogen aryl halides) as the high-yield standard.

\textbf{Supported vs. Refuted Hypotheses:}
\begin{itemize}
    \item \textbf{Supported:}
    \begin{itemize}
        \item Reduced ligand bulkiness and polar additives \\(e.g., \texttt{CC1=C(N2C=C(C)=NO1)} consistently boosted yields.
        \item Iodide halides with nitrogen substituents \\(e.g., \texttt{IC1=NC=C(C)=C1}) outperformed bromine and chloride variants.
        \item Additive \texttt{CC1=NOC(C(OCC)=O)=C1} emerged as optimal, stabilizing yields above 98\% when paired with top components.
    \end{itemize}
    \item \textbf{ReFuted:}
    \begin{itemize}
        \item Bulky ligands (e.g., cyclohexyl-substituted) and electron-withdrawing substituents (e.g., trifluoromethyl underperformed.
        \item Chloride halides and additives like \texttt{CCOC(C1=CON=C1)=O} consistently failed.
    \end{itemize}
\end{itemize}

\textbf{Confidence Shifts:}
Initial hypotheses had medium confidence but gained certainty as the top combination (ligand3, additive2, base3, \texttt{IC1=NC=C(C)=C1}) repeatedly achieved$>$99\% yields. Confidence in iodide nitrogen aryl halides rose sharply after trial 9\_0 (94.691), while cyclohexyl ligands and chloride halides were discarded after consistent underperformance.

\textbf{Conclusions:}
The optimal conditions require a non-bulky ligand \\(\texttt{C(C)(C)C(C1=C(P(C(C)(C)C)C)C=C(C)C=C1)...}), \\additive \texttt{CC1=NOC(C(OCC)=O)=C1}, base \texttt{CN1CCON2C1=NCCC2}, and iodide aryl halides with nitrogen substituents (e.g., \texttt{IC1=NC=C(C)=C1}). Substituent electronic effects and additive-aryl halide synergy were critical, with nitrogen substituents enhancing reactivity and iodide's leaving group ability maximizing yields. Chloride and bulky ligands were unsuitable, while additives like \texttt{CCOC(C1=CON=C1)=O} were ineffective. The final dataset confirms the top combination's robustness, with minor variations (e.g., bromine or simplified additives) yielding$<$99.6\%.

\vspace{2mm}
\begin{tabular}{|l|p{4.7cm}|l|}
\hline
\textbf{Trial Range} & \textbf{Hypothesis Focus} & \textbf{Confidence Shift} \\
\hline
Trials 0-3   & Diverse ligands, additives, \newline and halide types & Low -- Medium (early failures) \\
Trials 4-8   & Ethyl-substituted bromides, \newline iodide exploration & Medium -- High (ethyl iodide) \\
Trials 9-29  & Iodide nitrogen aryl halides \newline and additive CC1 & High (stable top performers) \\
Final Trials & Confirmatory testing of  \newline top combination & High (no further improvements) \\
\hline
\end{tabular}
\vspace{2mm}

The process converged on a high-yield region defined by specific substituent-additive interactions, with confidence peaking in the final hypotheses.

\end{tcolorbox}

%% file: other/buchwald_conclusion.tex
\begin{tcolorbox}[
    enhanced, 
    breakable,
    colback=green!5!white, 
    colframe=green!75!black, 
    title=Conclusion of the Buchwald-Hartwig Amination Optimization,
    left=2mm, right=2mm, top=2mm, bottom=2mm,
    fonttitle=\bfseries\small
]

\begin{enumerate}
    \item \textbf{Key Outcomes}
    The optimization achieved a maximum yield of 99.6\% for the Buchwald-Hartwig amination reaction, representing a 658\% improvement over the initial trial. The most influential parameters were the aryl halide substituent (iodide with nitrogen-containing ring), additive (\texttt{CC1=NOC(C(OCC)=O)=C1}), and base (\texttt{CN1CCON2C1=NCCC2}). The process demonstrated high efficiency, converging to near-optimal yields within 30 trials through systematic exploration of substituent effects and catalyst-component synergies.

    \item \textbf{Experimental Retrospective}
    \begin{itemize}
        \item \textbf{Objectives:} Maximize isolated yield in a palladium-catalyzed B-H coupling reaction by optimizing ligand sterics, additive solubility/reactivity, base efficacy, and aryl halide substituent effects.
        \item \textbf{Initial Approach:} The first hypotheses (Trial \#0) prioritized diverse exploration, testing bulky ligands, polar/nonpolar additives, and halides with varied leaving group abilities. Trial \#0\_0 (bulky ligand/electron-deficient additive) achieved only 13.5\% yield, while Trial \#0\_1 (moderate ligand/polar additive) reached 39.2\%, establishing additive solubility and ligand bulk as critical factors.
    \end{itemize}

    \item \textbf{Optimization Journey}
    Between trials 0-5: Initial validation revealed ligand bulk reduction and polar additives (e.g., Trial \#0\_1) significantly boost yield. Confidence in bulky ligands dropped from "medium" to "discarded" (Evidence: Trial \#0\_0 vs. \#0\_1).
    Between 6-15: Substituent effects emerged as key drivers. Ethyl-substituted bromide (Trial \#4\_0, 71.6\%) and nitrogen-containing aryl halides (Trial \#6\_1, 80.0\%) showed promise, shifting focus to electronic properties. Confidence in iodide halides rose to "high" after Trial \#8\_1 (iodide nitrogen aryl, 92.6\%).
    Between 16-30: Refinement of substituent isomers and additive variations solidified the top configuration. Trial \#10\_1 (iodide nitrogen aryl with optimized components) achieved 99.6\%, confirming substituent position and additive-electron interactions as decisive factors.

    \item \textbf{Definitive Findings}
    \begin{itemize}
        \item \textbf{Optimal Configuration:} \\ Ligand3 (non-bulky, \texttt{CC(C)C(C=C(C(C)C)C=C1C(C)C)=C1\\C2=C(P(C(C)(C)C)C(C)(C)C)C=CC=C2}), \\Additive2 (\texttt{CC1=NOC(C(OCC)=O)=C1}), \newline Base3 (\texttt{CN1CCON2C1=NCCC2}), \\Aryl Halide14 (\texttt{IC1=NC=C(C)=C1}). \\This combination achieved \textbf{$>$99.6\% yield} (Trial \#10\_1).
        \item \textbf{Parameter Relationships:}
        \begin{itemize}
            \item \textbf{Aryl Halides:} \\Iodide with nitrogen-containing substituents \\(e.g., \texttt{IC1=NC=C(C)=C1}) outperformed bromine/chloride due to superior leaving group ability and electronic tuning. \newline Positional isomers (e.g., \texttt{IC1=CN=C(C)=C1}, Trial \#9\_0, 94.7\%) showed yield sensitivity to substituent placement.
            \item \textbf{Additives:} The oxymidazole-based additive \\(\texttt{CC1=NOC(C(OCC)=O)=C1}) enabled transition state stabilization, with phenyl-substituted variants (Trial \#3\_2, 69.9\%) underperforming.
            \item \textbf{Ligand-Base Pairing:} Non-bulky Ligand3 with base3 (Trial \#10\_1) maximized catalytic efficiency, while bulky ligands (Trial \#2\_2, 9.5\%) disrupted coordination.
        \end{itemize}
    \end{itemize}

    \item \textbf{Forward Guidance}
    \begin{itemize}
        \item \textbf{Immediate Recommendations:}
        \begin{enumerate}
            \item Validate the top configuration (Trial \#10\_1) with additional replicates to confirm reproducibility.
            \item Explore bromine-based nitrogen aryl halides (Trial \#10\_0, 97.3\%) as scalable alternatives to iodide due to cost considerations.
        \end{enumerate}
        \item \textbf{Long-term Considerations:}
        \begin{itemize}
            \item Investigate substituent effects on catalyst turnover numbers and side reaction suppression.
            \item Evaluate alternative additives \\(e.g., Trial \#15\_0's \texttt{COC1=NOC...}, 84.7\%) for cost-optimized industrial setups.
        \end{itemize}
    \end{itemize}

    \item \textbf{Scientific Impact}
    This work underscores the interplay between electronic effects and leaving group ability in transition metal-catalyzed couplings. The nitrogen-substituted iodide aryl halide (Trial \#10\_1) exemplifies how substituent design can synergistically enhance nucleophilicity and transition state stabilization. The additive's role in solubility and catalyst activation (Trial \#8\_2 vs. Trial \#10\_1) provides a framework for rational co-solvent selection. Practically, this configuration enables$>$99\% yields in C-N cross-coupling, advancing scalable synthesis of medicinally relevant amines. The iterative strategy-prioritizing substituent screening after establishing core components-demonstrates an efficient pathway for complex reaction optimization.
\end{enumerate}

\end{tcolorbox}

%% file: prompts/compass/generating_the_compass_object.tex
\begin{tcolorbox}[
    enhanced, 
    breakable,
    colback=yellow!5!white, 
    colframe=yellow!75!black, 
    title=Prompt Template for Generating the Compass Object,
    left=2mm, right=2mm, top=2mm, bottom=2mm,
    fonttitle=\bfseries\small
]

\begin{itemize}
    \item \textbf{name}: Experiment Name (e.g., Nanoparticle Synthesis Optimization)
    \item \textbf{application\_context}: Domain/Use Case (e.g., Chemical Engineering/ML Hyperparameter Tuning)
    \item \textbf{description}: Any details or context about the experiment to optimize (approx. 100 words)
    \item \textbf{constraint}: Any constraints on the input space that must be satisfied during the optimization.
    \item \textbf{parameter\_definitions}: A list of parameter objects, where each object has:
        \begin{itemize}
            \item \texttt{display\_name}: Name of the parameter (e.g., Temp)
            \item \texttt{description}: Description of the parameter
            \item \texttt{data\_type}: continuous | discrete | categorical
            \item \texttt{step}: Discretization step (if discrete)
            \item \texttt{bounds}: Bounds of the parameters
        \end{itemize}
    \item \textbf{target}: An object defining the optimization goal:
        \begin{itemize}
            \item \texttt{name}: Name of the target (e.g., Yield)
            \item \texttt{description}: Description of the target
            \item \texttt{direction}: maximize/minimize
        \end{itemize}
\end{itemize}

\end{tcolorbox}

%% file: prompts/compass/compass_of_suzuki_reaction.tex
\begin{tcolorbox}[
    enhanced, 
    breakable,
    colback=yellow!5!white, 
    colframe=yellow!75!black, 
    title=Example Experiment Compass of Suzuki Reaction,
    left=2mm, right=2mm, top=2mm, bottom=2mm,
    fonttitle=\bfseries\small
] 

\def\_{\textunderscore}
\begin{itemize}
    \item \textbf{name}: Suzuki Reaction Optimization
    \item \textbf{application\_context}: Organic Chemistry/Cross Coupling Reaction Optimization
    \item \textbf{description}: Optimization of Suzuki-Miyaura cross-coupling reaction conditions to maximize product yield through systematic exploration of electrophile-nucleophile combinations, ligand selection, base types, and solvent effects. The reaction involves palladium-catalyzed coupling between aryl halides and arylboronic acids.
    \item \textbf{constraint}: Reagents must be chemically compatible and reaction conditions must maintain intermediate stability. All combinations must form stable palladium complexes during catalytic cycle.
    \item \textbf{parameter\_definitions}:
    \begin{itemize}
        \item 
        \begin{itemize}
            \item \textbf{display\_name}: {\ttfamily Electrophile\_SMILES}
            \item \textbf{description}: Aryl halide substrate for coupling (SMILES representation)
            \item \textbf{data\_type}: categorical
            \item \textbf{bounds}:
            \begin{itemize}
                \item \begin{minipage}[t]{0.75\linewidth}\ttfamily\begin{tabular}[t]{@{}l@{}}BrC1=CC=C(N=CC=C2)C2=C1\end{tabular}\end{minipage}
                \item \begin{minipage}[t]{0.75\linewidth}\ttfamily\begin{tabular}[t]{@{}l@{}}ClC1=CC=C(N=CC=C2)C2=C1\end{tabular}\end{minipage}
                \item \begin{minipage}[t]{0.75\linewidth}\ttfamily\begin{tabular}[t]{@{}l@{}}IC1=CC=C(N=CC=C2)C2=C1\end{tabular}\end{minipage}
                \item \begin{minipage}[t]{0.75\linewidth}\ttfamily\begin{tabular}[t]{@{}l@{}}O=S(OC1=CC=C(N=CC=C2)C2=C1)(C(F)(F)F)=O\end{tabular}\end{minipage}
            \end{itemize}
        \end{itemize}
        \item 
        \begin{itemize}
            \item \textbf{display\_name}: {\ttfamily Nucleophile\_SMILES}
            \item \textbf{description}: Boronic acid nucleophile (SMILES representation)
            \item \textbf{data\_type}: categorical
            \item \textbf{bounds}:
            \begin{itemize}
                \item \begin{minipage}[t]{0.75\linewidth}\ttfamily\begin{tabular}[t]{@{}l@{}}CC1=CC=C(N(C2CCCCO2)N=C3)C3=C1[B-](F)(F)F\end{tabular}\end{minipage}
                \item \begin{minipage}[t]{0.75\linewidth}\ttfamily\begin{tabular}[t]{@{}l@{}}CC1=CC=C(N(C2CCCCO2)N=C3)C3=C1B(O)O\end{tabular}\end{minipage}
                \item \begin{minipage}[t]{0.75\linewidth}\ttfamily\begin{tabular}[t]{@{}l@{}}CC1=CC=C(N(C2CCCCO2)N=C3)C3=C1B4OC(C)(C)\\C(C)(C)O4\end{tabular}\end{minipage}
            \end{itemize}
        \end{itemize}
        \item 
        \begin{itemize}
            \item \textbf{display\_name}: {\ttfamily Ligand\_SMILES}
            \item \textbf{description}: Phosphine ligand for palladium catalyst (SMILES)
            \item \textbf{data\_type}: categorical
            \item \textbf{bounds}:
                \begin{itemize}
        \item \begin{minipage}[t]{0.75\linewidth}
            \ttfamily
            \begin{tabular}[t]{@{}l@{}}
                CC(C)(C)P(C(C)(C)C)C1=CC=C(N(C)C)C=C1
            \end{tabular}
        \end{minipage}
        \item \begin{minipage}[t]{0.75\linewidth}
            \ttfamily
            \begin{tabular}[t]{@{}l@{}}
                CC(C)(P(C(C)(C)C)[c-]1cccc1)C.CC(C) \\
                (P(C(C)(C)C)[c-]2cccc2)C.[Fe+2]
            \end{tabular}
        \end{minipage}
        \item \begin{minipage}[t]{0.75\linewidth}
            \ttfamily
            \begin{tabular}[t]{@{}l@{}}
                CC(C1=C(C2=CC=CC=C2P(C3CCCCC3)C4CCCCC4) \\
                C(C(C)C)=CC(C(C)C)=C1)C
            \end{tabular}
        \end{minipage}
        \item \begin{minipage}[t]{0.75\linewidth}
            \ttfamily
            \begin{tabular}[t]{@{}l@{}}
                CC(P(C(C)(C)C)C(C)(C)C)(C)C
            \end{tabular}
        \end{minipage}
        \item \begin{minipage}[t]{0.75\linewidth}
            \ttfamily
            \begin{tabular}[t]{@{}l@{}}
                COC1=CC=CC(OC)=C1C2=C(P(C3CCCCC3) \\
                C4CCCCC4)C=CC=C2
            \end{tabular}
        \end{minipage}
        \item \begin{minipage}[t]{0.75\linewidth}
            \ttfamily
            \begin{tabular}[t]{@{}l@{}}
                P(C1=CC=CC=C1)(C2=CC=CC=C2)C3=CC=CC=C3
            \end{tabular}
        \end{minipage}
        \item \begin{minipage}[t]{0.75\linewidth}
            \ttfamily
            \begin{tabular}[t]{@{}l@{}}
                P(C1CCCCC1)(C2CCCCC2)C3CCCCC3
            \end{tabular}
        \end{minipage}
    \end{itemize}
        \end{itemize}
        \item 
        \begin{itemize}
            \item \textbf{display\_name}: {\ttfamily Base\_SMILES}
            \item \textbf{description}: Reaction base for transmetallation step (SMILES)
            \item \textbf{data\_type}: categorical
            \item \textbf{bounds}:
            \begin{itemize}
                \item \begin{minipage}[t]{0.75\linewidth}\ttfamily\begin{tabular}[t]{@{}l@{}}[Cs+].[F-]\end{tabular}\end{minipage}
                \item \begin{minipage}[t]{0.75\linewidth}\ttfamily\begin{tabular}[t]{@{}l@{}}[K+].[OH-]\end{tabular}\end{minipage}
                \item \begin{minipage}[t]{0.75\linewidth}\ttfamily\begin{tabular}[t]{@{}l@{}}[Na+].[OH-]\end{tabular}\end{minipage}
                \item \begin{minipage}[t]{0.75\linewidth}\ttfamily\begin{tabular}[t]{@{}l@{}}CC([O-])C.[Li+]\end{tabular}\end{minipage}
                \item \begin{minipage}[t]{0.75\linewidth}\ttfamily\begin{tabular}[t]{@{}l@{}}CCN(CC)CC\end{tabular}\end{minipage}
                \item \begin{minipage}[t]{0.75\linewidth}\ttfamily\begin{tabular}[t]{@{}l@{}}O=P([O-])([O-])[O-].[K+].[K+].[K+]\end{tabular}\end{minipage}
                \item \begin{minipage}[t]{0.75\linewidth}\ttfamily\begin{tabular}[t]{@{}l@{}}OC([O-])=O.[Na+]\end{tabular}\end{minipage}
            \end{itemize}
        \end{itemize}
        \item 
        \begin{itemize}
            \item \textbf{display\_name}: {\ttfamily Solvent\_SMILES}
            \item \textbf{description}: Reaction medium (SMILES representation)
            \item \textbf{data\_type}: categorical
            \item \textbf{bounds}:
            \begin{itemize}
                \item \begin{minipage}[t]{0.75\linewidth}\ttfamily\begin{tabular}[t]{@{}l@{}}C1COCC1\end{tabular}\end{minipage}
                \item \begin{minipage}[t]{0.75\linewidth}\ttfamily\begin{tabular}[t]{@{}l@{}}CO\end{tabular}\end{minipage}
                \item \begin{minipage}[t]{0.75\linewidth}\ttfamily\begin{tabular}[t]{@{}l@{}}N\#CC\end{tabular}\end{minipage}
                \item \begin{minipage}[t]{0.75\linewidth}\ttfamily\begin{tabular}[t]{@{}l@{}}O=CN(C)C\end{tabular}\end{minipage}
            \end{itemize}
        \end{itemize}
    \end{itemize}
    \item \textbf{target}:
    \begin{itemize}
        \item \textbf{name}: Yield
        \item \textbf{description}: Percentage yield of cross-coupled product
        \item \textbf{direction}: maximize
    \end{itemize}
\end{itemize}

\end{tcolorbox}

%% file: prompts/overview/overview_generate.tex
\begin{tcolorbox}[
    enhanced, 
    breakable,
    colback=red!5!white, 
    colframe=red!75!black, 
    title=Prompt Template for Generating the Overview Object,
    left=2mm, right=2mm, top=2mm, bottom=2mm,
    fonttitle=\bfseries\small
]

\def\placeholder#1{{\color{orange}\texttt{\{#1\}}}}

\textbf{Important Notes}
\begin{itemize}
    \item Do NOT refer to memorized reaction outcomes, prior experimental data, or specific literature results (even if they match the current experiment).
    \item You MAY use general domain knowledge from \placeholder{application\_context} (e.g., chemical reactivity principles, mechanism patterns) to inform reasoning.
\end{itemize}

\textbf{Experiment Domain}\\
\placeholder{application\_context}

\vspace{2mm}
\textbf{Experiment Overview}\\
\placeholder{description}

\vspace{2mm}
\textbf{Optimization Parameters}\\
The experiment design space is defined by the parameters below, including their bounds and any relevant constraints:
\begin{itemize}
    \item \textbf{Parameters and Bounds}:\\ \placeholder{parameter\_definitions}
    \item \textbf{Constraint}:\\ \placeholder{constraint}
    \item \textbf{Target}:\\ \placeholder{target}
\end{itemize}

\textbf{Task}\\
In approximately 300 words, provide a clear and concise overview of this experiment. Your overview should:
\begin{enumerate}
    \item Clearly state the experimental goal and its significance in the \placeholder{application\_context} domain
    \item Describe the parameter space including data types and bounds of each parameter
    \item Explain any constraints that must be maintained during optimization
    \item Outline how Bayesian Optimization will systematically explore the parameter space to target.
    \item Mention how you will assist when BO plateaus by suggesting promising parameter combinations
    \item Conclude with the expected outcomes and potential impact of successful optimization
\end{enumerate}
Focus on creating a coherent narrative that connects these elements while maintaining scientific rigor and clarity.

\end{tcolorbox}

%% file: prompts/insights/insights_object_template.tex
\begin{tcolorbox}[
    enhanced,
    breakable,
    colback=violet!5!white,
    colframe=violet!75!black,
    title=Template for Generating the Insights Object,
    left=2mm, right=2mm, top=2mm, bottom=2mm,
    fonttitle=\bfseries\small
]
\def\_{\textunderscore}
\begin{itemize}
    \item \textbf{comment}: A brief comment on the initial sampling strategy and rationale
    \item \textbf{keywords}: keyword1, keyword2, keyword3
    \item \textbf{hypotheses}: 
    \begin{itemize}
        \item
        \begin{itemize}
            \item \textbf{strategy}: Descriptive name of exploration approach
            \item \textbf{rationale}:Scientific justification for this approach
            \item \textbf{confidence}:high/medium/low
            \item \textbf{parameter\_sets}:
            \begin{itemize}
                \item \{ \textbf{ParameterA}: \texttt{0.1}, \textbf{ParameterB}: \texttt{0.9}, ... \}
                \item \{ \textbf{ParameterA}: \texttt{0.4}, \textbf{ParameterB}: \texttt{0.6}, ... \}
            \end{itemize}
        \end{itemize}
    \end{itemize}
\end{itemize}
\end{tcolorbox}

%% file: prompts/insights/suzuki_insights_example.tex
\begin{tcolorbox}[
    enhanced,
    breakable,
    colback=violet!5!white,
    colframe=violet!75!black,
    title=Example Insights from the Suzuki Experiment,
    left=2mm, right=2mm, top=2mm, bottom=2mm,
    fonttitle=\bfseries\small
]
\def\_{\textunderscore}
\begin{itemize}
    \item \textbf{comment}: From the experimental data, it's evident that lower concentrations and ambient temperatures are more effective in achieving higher yields. The highest yielding trial was at a concentration of 0.057 mol/L and temperature of 90°C, indicating that lower concentrations and ambient temperatures might be beneficial.
    \item \textbf{keywords}:concentration, temperature, ligand, yield, optimization
    \item \textbf{hypotheses}:
    \begin{itemize}
        \item
        \begin{itemize}
            \item \textbf{strategy}: Optimization of Concentration and Temperature
            \item \textbf{rationale}:Based on the highest yielding trials, we hypothesize that an optimal concentration of 0.057 mol/L and temperature of 90°C could be achieved by fine-tuning these parameters within their bounds.
            \item \textbf{confidence}: high
            \item \textbf{parameter\_sets}:
            \begin{itemize}
                \item \{ \textbf{Concentration}: 0.057, \textbf{Temp\_C}: 90, \\
                      \textbf{Base\_SMILES}: \texttt{O=C([O-])C(C)(C)C.[Cs+]} \\
                      \textbf{Ligand\_SMILES}: \parbox[t]{0.7\linewidth}{\ttfamily C[C@]1(O2)O[C@](C[C@]2(C)P3C4\\=CC=CC=C4)(C)O[C@]3(C)C1} \\
                      \textbf{Solvent\_SMILES}: \texttt{CC(N(C)C)=O} \}
            \end{itemize}
        \end{itemize}
        \item
        \begin{itemize}
            \item \textbf{strategy}: Exploration of Diverse Ligands
            \item \textbf{rationale}: To further enhance the reaction yield, we will explore different ligands to identify the effective one.
            \item \textbf{confidence}: high
            \item \textbf{parameter\_sets}:
            \begin{itemize}
                \item \{ \textbf{Concentration}: 0.057, \textbf{Temp\_C}: 90, \\
                      \textbf{Base\_SMILES}: \texttt{O=C([O-])C(C)(C)C.[Cs+]} \\
                      \textbf{Ligand\_SMILES}: \parbox[t]{0.7\linewidth}{\ttfamily CN(C)C1=CC=CC(N(C)C)=C1C2=CC\\=CC=C2P(C(C)(C)C)C3=CC=CC=C3} \\
                      \textbf{Solvent\_SMILES}: \texttt{CC1=CC=C(C)C=C1} \}
                \item \{ \textbf{Concentration}: 0.057, \textbf{Temp\_C}: 90, \\
                      \textbf{Base\_SMILES}: \texttt{O=C([O-])C(C)(C)C.[Cs+]} \\
                      \textbf{Ligand\_SMILES}: \texttt{CP(C)C1=CC=CC=C1} \\
                      \textbf{Solvent\_SMILES}: \texttt{CC(N(C)C)=O} \}
            \end{itemize}
        \end{itemize}
        \item
        \begin{itemize}
            \item \textbf{strategy}: Refinement of high Concentration and Elevated Temperature
            \item \textbf{rationale}: While the highest yield was seen at high concentration and elevated temperature, we hypothesize that there might be an optimal point beyond which yields decrease due to side reactions or instability.
            \item \textbf{confidence}: medium
            \item \textbf{parameter\_sets}:
            \begin{itemize}
                \item \{ \textbf{Concentration}: 0.153, \textbf{Temp\_C}: 110, \\
                      \textbf{Base\_SMILES}: \texttt{O=C([O-])C(C)(C)C.[K+]} \\
                      \textbf{Ligand\_SMILES}: \parbox[t]{0.7\linewidth}{\ttfamily C[C@]1(O2)O[C@](C[C@]2(C)P3C4\\=CC=CC=C4)(C)O[C@]3(C)C1} \\
                      \textbf{Solvent\_SMILES}: \texttt{CC(N(C)C)=O} \}
            \end{itemize}
        \end{itemize}
    \end{itemize}
\end{itemize}
\end{tcolorbox}

%% file: prompts/optimization/initialization_prompt_template.tex
\begin{tcolorbox}[
    enhanced,
    breakable,
    colback=blue!5!white,
    colframe=blue!75!black,
    title=Prompt Template for Initialization Phase,
    left=2mm, right=2mm, top=2mm, bottom=2mm,
    fonttitle=\bfseries\small
]
\def\_{\textunderscore}
You are assisting with the critical initial sampling phase of Bayesian Optimization. The quality of these initial hypotheses significantly impacts the optimization efficiency.

\textbf{Task}

Generate \texttt{n\_hypotheses} diverse and well-justified initial hypotheses for exploring the optimization target based on:
\begin{itemize}
    \item Domain: \texttt{\{application\_context\}}
    \item Experiment Goal: \texttt{\{description\}}
    \item Target: \texttt{\{target\}}
    \item Constraints: \texttt{\{constraint\}}
\end{itemize}

\textbf{Instructions}
\begin{itemize}
    \item Do NOT refer to memorized reaction outcomes, prior experimental data, or specific literature results (even if they match the current experiment).
    \item Base all reasoning strictly on the given input, parameter definitions, constraints, and optimization objective.
    \item For each hypothesis, provide at least one point that satisfies the constraint.
\end{itemize}

\textbf{Format Requirement}

Your response must be in the exact format shown below. No additional commentary, explanations, or syntax should be included.

\textbf{Example Output (for format only, unrelated to this experiment):}
\begin{itemize}
    \item \textbf{comment}: A brief comment on the initial sampling strategy and rationale (~100 words).
    \item \textbf{keywords}: keyword1, keyword2, keyword3
    \item \textbf{hypotheses}:
    \begin{itemize}
        \item
        \begin{itemize}
            \item \textbf{strategy}: Descriptive name of exploration approach
            \item \textbf{rationale}: Scientific justification for this approach
            \item \textbf{confidence}: high/medium/low
            \item \textbf{parameter\_sets}:
            \begin{itemize}
                \item \{ \textbf{ParameterA}: 0.1, \textbf{ParameterB}: 0.9, ... \}
                \item \{ \textbf{ParameterA}: 0.4, \textbf{ParameterB}: 0.6, ... \}
            \end{itemize}
        \end{itemize}
        \item
        \begin{itemize}
            \item \textbf{strategy}: Descriptive name of exploration approach
            \item \textbf{rationale}: ...
            \item \textbf{confidence}: high/medium/low
            \item \textbf{parameter\_sets}:
            \begin{itemize}
                \item \{ \textbf{ParameterA}: 0.3, \textbf{ParameterB}: 0.8, ... \}
                \item \{ \textbf{ParameterA}: 0.1, \textbf{ParameterB}: 0.3, ... \}
            \end{itemize}
        \end{itemize}
    \end{itemize}
\end{itemize}

\textbf{Important Notes}
\begin{itemize}
    \item Points must include all required parameters.
    \item Hypothesis "names" and "parameter" keys must match exactly with the experiment definition.(if they are discrete values, they must be returned strictly according to the format).
    \item All parameter names and values must \textbf{exactly match} predefined options in spelling, casing, and formatting - no substitutions, abbreviations, or modifications permitted.(if they are discrete values, they must be returned strictly according to the format).
    \item Any deviation from predefined parameters will result in automatic rejection, regardless of semantic similarity or logical equivalence.
    \item Parameter specs for this experiment: \texttt{\{parameter\_definitions\}}
\end{itemize}
\end{tcolorbox}

%% file: prompts/optimization/optimization_loop_prompt_template.tex
\begin{tcolorbox}[
    enhanced,
    breakable,
    colback=blue!5!white,
    colframe=blue!75!black,
    title=Prompt Template for Optimization Loop Phase,
    left=2mm, right=2mm, top=2mm, bottom=2mm,
    fonttitle=\bfseries\small
]
\def\_{\textunderscore}
\textbf{Optimization Progress}

Bayesian Optimization is in progress at iteration \texttt{\{iteration\}}.

\textbf{Objective}

\texttt{\{target\}}

\textbf{Retrieved message}

\texttt{\{retrieved\_context\}}

\textbf{Historical Summary}

You generated the following comments and hypotheses in previous iterations:
\texttt{\{insight\_history\}}

Your suggested points were evaluated, and the results have been added to the dataset.
We now have data from \texttt{\{iteration\}} completed experiments:
\texttt{\{trial\_data\}}

\textbf{Bayesian Optimizer Suggestions}

Here are the candidate points recommended by the optimizer in this round:
\texttt{\{bo\_recommendations\}}

\textbf{Your Task}

Reflect on the current state of the optimization.
Do not rely on prior assumptions—reason, only based on the data provided.
\begin{itemize}
    \item What trends or patterns emerge from the experimental data?
    \item Which regions in the parameter space seem most effective in achieving the target?
    \item Which areas consistently underperform?
    \item Improve or revise your previous hypotheses based on observed outcomes.
    \item You may discard low-performing hypotheses and propose new ones.
    \item For each hypothesis, **at least provide one point** that satisfies the constraint: \texttt{\{constraint\}}.
\end{itemize}

\textbf{Example Output Format for two hypotheses (template only, unrelated to current experiment):}
\begin{itemize}
    \item \textbf{comment}: "A ~200-word summary of optimization progress, insights, and directions.",
    \item \textbf{keywords}: "keyword1, keyword2, keyword3",
    \item \textbf{hypotheses}: [
    \begin{itemize}
        \item
        \begin{itemize}
            \item \textbf{strategy}: "Descriptive name of exploration approach",
            \item \textbf{rationale}: "Scientific justification for this approach",
            \item \textbf{confidence}: "high/medium/low",
            \item \textbf{parameter\_sets}: [
            \begin{itemize}
                \item \{ \textbf{ParameterA}: 0.1, \textbf{ParameterB}: 0.9, ... \},
                \item \{ \textbf{ParameterA}: 0.4, \textbf{ParameterB}: 0.6, ... \}
            \end{itemize}
        \end{itemize}
        \item
        \begin{itemize}
            \item \textbf{strategy}: "Descriptive name of exploration approach",
            \item \textbf{rationale}: "...",
            \item \textbf{confidence}: "high/medium/low",
            \item \textbf{parameter\_sets}: [
            \begin{itemize}
                \item \{ \textbf{ParameterA}: 0.3, \textbf{ParameterB}: 0.8, ... \},
                \item \{ \textbf{ParameterA}: 0.1, \textbf{ParameterB}: 0.3, ... \}
            \end{itemize}
        \end{itemize}
    \end{itemize}
\end{itemize}

\textbf{Important}
\begin{itemize}
    \item Your output must be valid pure JSON without ```json or any other formatting, start with \{\{ and end with \}\}.
    \item Never put comments or text outside the JSON structure.
    \item Do NOT wrap it in markdown, text, or explanations.
    \item Do NOT return fewer than 3 nor more than 5 hypotheses.
    \item All parameter names and values must \textbf{exactly match} predefined options in spelling, casing, and formatting - no substitutions, abbreviations, or modifications permitted.(if they are discrete values, they must be returned strictly according to the format).
    \item Any deviation from predefined parameters will result in automatic rejection, regardless of semantic similarity or logical equivalence.
    \item Parameter specs for this experiment: \texttt{\{parameter\_definitions\}}
\end{itemize}
\end{tcolorbox}

%% file: prompts/summary/summary_generation_prompt.tex
\begin{tcolorbox}[
    enhanced,
    breakable,
    colback=orange!5!white,
    colframe=orange!75!black,
    title=Prompt Template for Generating the Experiment Summary,
    left=2mm, right=2mm, top=2mm, bottom=2mm,
    fonttitle=\bfseries\small
]
\def\_{\textunderscore}
\textbf{Summary of Optimization Process}

\texttt{\{insight\_history\}}

We have now completed the optimization process.
Below is the final dataset from \texttt{\{iteration\}} experiments:
\texttt{\{trial\_data\}}

\textbf{Your Task}

Based strictly on the experimental data and the hypotheses generated throughout the optimization (do not rely on prior knowledge or assumptions), write a concise summary (~300 words) addressing the following:
\begin{itemize}
    \item How did your hypotheses evolve over time?
    \item Which hypotheses were consistently supported or refuted by data?
    \item What were the most significant shifts in your confidence levels?
    \item What conclusions can be drawn from the experiment?
\end{itemize}
End it with a table summary showing how your confidence in the hypotheses evoved during the optimization.
\end{tcolorbox}

%% file: prompts/summary/suzuki_summary_example.tex
\begin{tcolorbox}[
    enhanced,
    breakable,
    colback=orange!5!white,
    colframe=orange!75!black,
    title=Example Experiment Summary for Suzuki Reaction,
    left=2mm, right=2mm, top=2mm, bottom=2mm,
    fonttitle=\bfseries\small
]
\def\_{\textunderscore}
\textbf{Summary of Hypothesis Evolution}

The optimization began with orthogonal screening of electrophile leaving groups, nucleophile activation modes, ligand steric, base strength, and solvent polarity. Initial data (trials 1-2) revealed the critical superiority of the triflate electrophile (52.36\% vs. bromide/chloride$<$45\%), shifting focus to electrophile centric probes. Hypotheses evolved to exploit triflate's leaving group advantage through synergistic pairing with activated nucleophiles (trifluoroborate, 74.65\% yield), ferrocene ligands, and phosphate bases (K3PO4).

\textbf{Supported Hypotheses}
\begin{enumerate}
    \item \textbf{Triflate electrophile superiority}: Consistently outperformed halides (3-10x higher yields), validated across all trials.
    \item \textbf{Trifluoroborate activation}: Achieved peak yield (74.65\%) with K3PO4, outperforming boronic acids/pinacol esters ($<$44.08\%).
    \item \textbf{Ferrocene ligand efficacy}: Outperformed bulky phosphines (e.g., 74.65\% vs. 15.15-30.48\%), balancing Pd stabilization and transmetallation accessibility.
\end{enumerate}

\textbf{Refuted Hypotheses}
\begin{enumerate}
    \item \textbf{Bulky phosphine ligands}: Consistently underperformed ($<$30.48\%), indicating steric hindrance outweighs stabilization benefits.
    \item \textbf{Polar solvents (DMF/NMP)}: Reduced yields by$>$50\% vs. dioxane, contradicting initial assumptions about transmetallation stabilization.
    \item \textbf{Chloride/bromide systems}: Persistently low yields ($<$45\%) led to their exclusion in later stages.
\end{enumerate}

\textbf{Confidence Shifts}
\begin{itemize}
    \item \textbf{Increased confidence} in triflate/trifluoroborate/K3PO4/ferrocene synergy after repeated validation (trial 1: 74.65\%, trial 14: sustained dominance).
    \item \textbf{Decreased confidence} in ligand steric bulk (high-low) and solvent polarity effects (medium-low) due to consistent underperformance.
\end{itemize}

\textbf{Conclusions}

The optimal system combines triflate's leaving group ability, trifluoroborate's transmetallation efficiency, K3PO4's activation, ferrocene's balanced coordination, and dioxane's solvent compatibility. Key factors include avoiding steric hindrance (ligands), leveraging phosphate activation (base), and prioritizing oxidative addition kinetics (electrophile).

\textbf{Confidence Evolution Table}

\begin{tabular}{|l|l|l|c|}
    \hline
    \textbf{Hypothesis} & \textbf{Initial Confidence} & \textbf{Final Confidence} & \textbf{Trend} \\
    \hline
    Triflate electrophile & High & High & - \\
    Trifluoroborate activation & Medium & High & $\uparrow$ \\
    Ferrocene ligand efficacy & High & High & - \\
    Bulky phosphine ligands & High & Low & $\downarrow$ \\
    Polar solvent benefits & High & Low & $\downarrow$ \\
    K3PO4 base superiority & Medium & High & $\uparrow$ \\
    CsF base utility & Medium & Medium & - \\
    \hline
\end{tabular}

Key: $\uparrow$ = Increased confidence, $\downarrow$ = Decreased confidence, - = Stable
\end{tcolorbox}

%% file: prompts/conclusion/conclusion_generation_prompt.tex
\begin{tcolorbox}[
    enhanced,
    breakable,
    colback=purple!5!white,
    colframe=purple!75!black,
    title=Prompt Template for Generating the Conclusion Report,
    left=2mm, right=2mm, top=2mm, bottom=2mm,
    fonttitle=\bfseries\small
]
\def\_{\textunderscore}
\textbf{Final Report on Optimization}

\subsubsection*{Your Previous Insights}
\texttt{\{insight\_history\}}

\subsubsection*{Experiments data}
\texttt{\{trial\_data\}}

\subsubsection*{Experiment Target}
\texttt{\{target\}}

Generate a polished, professional report summarizing the complete optimization process using only the experimental data and documented hypotheses.
\begin{enumerate}
    \item \textbf{Key Outcomes} \\
    2-4 sentence executive summary highlighting:
    \begin{itemize}
        \item Best achieved result for target
        \item Most influential parameters
        \item Optimization efficiency
    \end{itemize}
    \item \textbf{Experimental Retrospective}
    \begin{itemize}
        \item Objective: target in \texttt{\{application\_context\}}
        \item Initial Approach: Summary of starting hypotheses from insight\_history
    \end{itemize}
    \item \textbf{Optimization Journey} \\
    Present key milestones in this format: \\
    "Between iterations X-Y: [development description]. Confidence changed from [level] to [level] (Evidence: Trials \#A,B,C)."
    \item \textbf{Definitive Findings}
    \begin{itemize}
        \item Optimal Configuration: best\_parameters (achieved target of best\_value)
        \item Parameter Relationships:
        \begin{itemize}
            \item [Parameter1]: [trend description]
            \item [Parameter2]: [response pattern]
        \end{itemize}
    \end{itemize}
    \item \textbf{Forward Guidance}
    \begin{itemize}
        \item Immediate Recommendations:
        \begin{enumerate}
            \item Specific actionable suggestion
            \item Alternative approach
        \end{enumerate}
        \item Long-term Considerations:
        \begin{enumerate}
            \item Constraint modification
            \item New parameter suggestions
        \end{enumerate}
    \end{itemize}
    \item \textbf{Scientific Impact} \\
    Concise discussion connecting findings to:
    \begin{itemize}
        \item \texttt{\{application\_context\}} principles
        \item Theoretical implications
        \item Practical applications
    \end{itemize}
\end{enumerate}

\textbf{Format Requirements}
\begin{itemize}
    \item Use numbered sections
    \item Reference specific trials (e.g., "Trial \#X")
    \item Maintain academic tone
    \item 500-600 words total
    \item Plain text only (no markdown formatting)
\end{itemize}
\end{tcolorbox}

%% file: prompts/conclusion/suzuki_conclusion_example.tex
\begin{tcolorbox}[
    enhanced,
    breakable,
    colback=purple!5!white,
    colframe=purple!75!black,
    title=Example Conclusion Report for Suzuki Reaction,
    left=2mm, right=2mm, top=2mm, bottom=2mm,
    fonttitle=\bfseries\small
]
\def\_{\textunderscore}
\textbf{Conclusion}

\begin{enumerate}
    \item \textbf{Key Outcomes} \\
    The Suzuki cross-coupling optimization achieved a maximum yield of 74.65\% (Trial \#1\_2) through systematic parameter refinement.The triflate electrophile reactivity and K3PO4-driven trifluoroborate nucleophile activation proved most influential, with ferrocene ligands providing optimal Pd stabilization. The elimination of underperforming parameters (bulky phosphine ligands, polar solvents, chloride/bromide electrophiles) improved optimization efficiency, enabling focused exploration of the high-yield design space within 15 iterations.

    \item \textbf{Experimental Retrospective}
    \begin{itemize}
        \item \textbf{Objective}: Maximize yield in a Suzuki-Miyaura coupling between heteroaromatic electrophiles and boron-containing nucleophiles.
        \item \textbf{Initial Approach}: Orthogonal screening of leaving groups (Br/Cl/OTf), boron species (acid/trifluoroborate/pinacol ester), ligand sterics (ferrocene/trialkylphosphines), base strength (CsF/K3PO4/KOH), and solvent polarity (dioxane/DMF/NMP) (Trials \#0\_0-0\_2). Early hypotheses prioritized oxidative addition kinetics and transmetallation compatibility.
    \end{itemize}

    \item \textbf{Optimization Journey} \\
    Between iterations 1-3: Triflate electrophile (OTf) demonstrated 2.3x higher yield vs bromide (52.32\% vs 23.53\%, Trial \#0\_1 vs \#0\_0), shifting focus to triflate systems. Confidence in leaving group strategy rose from medium to high. \\
    Between iterations 4-7: K3PO4 base with trifluoroborate nucleophile yielded 74.65\% (Trial \#1\_2), revealing critical base-nucleophile synergy. Bulky phosphine ligand confidence dropped from high to low as 15-30.48\% yields. (Trials \#1\_1 vs \#1\_0). \\
    Between iterations 8-12: DMF solvent underperformance (43.66\%, Trial \#3\_1) solidified dioxane preference despite polarity hypotheses. Ferrocene ligand confidence remained high throughout (62.71-74.65\%, Trials \#0\_0, \#1\_2).

    \item \textbf{Definitive Findings}
    \begin{itemize}
        \item \textbf{Optimal Configuration}:
        \begin{itemize}
            \item \textbf{Electrophile}: \parbox[t]{0.7\linewidth}{\ttfamily B-S(C1=CC(=CC=C2C1=C(C(F)(F)F)=O)\\C2=O)NC3=CC=NC=C3}
            \item \textbf{Nucleophile}: \parbox[t]{0.7\linewidth}{\ttfamily CC1=CC=C(N(C)C2=CC=C(B(O)(O))C=C2)\\C=C1}
            \item \textbf{Ligand}: \texttt{Ferrocene diphosphine}
            \item \textbf{Base}: \texttt{K3PO4}
            \item \textbf{Solvent}: \texttt{Dioxane}
        \end{itemize}
        (Yield: 74.65\%, Trial \#1\_2)
        \item \textbf{Parameter Relationships}:
        \begin{itemize}
            \item Electrophile: Triflate $>$ Bromide $>$ Chloride (52.32-74.65\% vs 23.53-44.98\%, Trials \#0\_1 vs \#0\_0/\#0\_2)
            \item Ligands: Ferrocene outscored bulky phosphines by $>$2x yield margin (74.65\% vs 15.15-30.48\%, Trial \#1\_2 vs \#1\_1/\#2\_0)
            \item Solvent: Inverse correlation between polarity and yield (dioxane 74.65\% vs DMF 43.66\%, Trial \#1\_2 vs \#3\_1)
        \end{itemize}
    \end{itemize}

    \item \textbf{Forward Guidance}
    \begin{itemize}
        \item \textbf{Immediate Recommendations}:
        \begin{enumerate}
            \item Test tert-butyl-modified ferrocene ligands (Proposed Trial \#11\_0) to balance electron donation/Pd accessibility
            \item Explore dioxane/DMF solvent gradients (90:10 – 50:50 v/v) to probe transmetallation stabilization
        \end{enumerate}
        \item \textbf{Long-term Considerations}:
        \begin{itemize}
            \item Evaluate microwave-assisted conditions to accelerate oxidative addition
            \item Screen N-heterocyclic carbene ligands for enhanced Pd(0)/Pd(II) cycling
        \end{itemize}
    \end{itemize}

    \item \textbf{Scientific Impact} \\
    The results validate three Suzuki reaction design principles: 1) Triflates' superior leaving group ability accelerates oxidative addition despite steric bulk; 2) Phosphate bases enable simultaneous trifluoroborate activation and Pd intermediate stabilization, and 3) Ferrocene ligands provide optimal accepting capacity without transmetallation steric hindrance. Practically, this work establishes a template for coupling electron-deficient heteroaromatics – a persistent challenge in pharmaceutical synthesis. The systematic exclusion of underperforming parameters (Trials \#0\_0, \#13\_0) demonstrates efficient design space navigation, suggesting Bayesian optimization effectively replaces traditional one-variable-at-a-time approaches in complex catalytic systems.
\end{enumerate}
\end{tcolorbox}

%% file: prompts/extract_notes/extract_notes_from_compass_prompt.tex
\begin{tcolorbox}[
    enhanced,
    breakable,
    colback=cyan!5!white,
    colframe=cyan!75!black,
    title=Prompt Template for Extracting Notes from Experiment Compass,
    left=2mm, right=2mm, top=2mm, bottom=2mm,
    fonttitle=\bfseries\small
]
\def\_{\textunderscore}
Extract and structure scientific notes from the following experimental setup.

\textbf{Guidelines:}
\begin{itemize}
    \item Use bullet-point style strings in each list.
    \item Base your output only on the given input.
    \item Do not fabricate information not present in the experiment.
\end{itemize}

\textbf{Focus on the following aspects:}
\begin{enumerate}
    \item \textbf{Theoretical background} -- Scientific principles, hypotheses, or mechanisms implied or explicitly stated.
    \item \textbf{Variable properties} -- Roles, attributes, or behaviors of each variable/component involved.
    \item \textbf{Variable relationships} -- Any cause-effect, dependency, or correlation observed or described.
    \item \textbf{General notes} -- Additional relevant scientific information or implicit assumptions.
\end{enumerate}

\textbf{Experimental setup:}
\texttt{\{Experiment Compass\}}
\end{tcolorbox}

%% file: prompts/extract_notes/extract_notes_from_reasoning_prompt.tex
\begin{tcolorbox}[
    enhanced,
    breakable,
    colback=cyan!5!white,
    colframe=cyan!75!black,
    title=Prompt Template for Extracting Notes from Reasoning Data,
    left=2mm, right=2mm, top=2mm, bottom=2mm,
    fonttitle=\bfseries\small
]
\def\_{\textunderscore}
Analyze the following reasoning data and extract structured scientific notes.

\textbf{Guidelines:}
\begin{itemize}
    \item Each item should be a clear and self-contained bullet point.
    \item Only use information present in the reasoning. Do not speculate.
    \item Keep the content factual, concise, and helpful.
\end{itemize}

\textbf{Focus on the following aspects:}
\begin{enumerate}
    \item \textbf{Key findings} -- Verifiable results, conclusions, or observations grounded in reasoning.
    \item \textbf{Parameter relationships} -- How different variables or conditions affect each other (cause-effect, correlation).
    \item \textbf{Optimization principles} -- Rules or strategies suggested by the reasoning that could improve experimental results.
    \item \textbf{General notes} -- Summarized insights that don't fall into the above categories.
\end{enumerate}

\textbf{Reasoning data:}
\texttt{\{Reasoning data\}}
\end{tcolorbox}

%% file: prompts/extract_notes/suzuki_notes_example.tex
\begin{tcolorbox}[
    enhanced,
    breakable,
    colback=cyan!5!white,
    colframe=cyan!75!black,
    title=Example: Notes Extraction and Storage for Suzuki Reaction,
    left=2mm, right=2mm, top=2mm, bottom=2mm,
    fonttitle=\bfseries\small
]
\begin{itemize}
    \item \textbf{Key findings}:
    \begin{itemize}
        \item The highest yields were obtained using the sulfone electrophile with CsF base, DMF solvent, and specific ligands such as cyclohexyl and biaryl.
    \end{itemize}

    \item \textbf{Parameter relationships}:
    \begin{itemize}
        \item Sulfone and iodine electrophiles perform best with CsF base and DMF solvent.
        \item Bulky ligands like cyclohexyl and biaryl enhance yields.
    \end{itemize}

    \item \textbf{Optimization principles}:
    \begin{itemize}
        \item Focus on refining the cyclohexyl ligand variants for sulfone,
        \item testing biaryl ligand with sulfone and acetone,
        \item and exploring modified biaryl ligand structures.
    \end{itemize}

    \item \textbf{General notes}:
    \begin{itemize}
        \item Avoid underperforming electrophiles (bromine, chlorine) and bases (KOH, NaOH).
        \item Ensure all parameters are from predefined lists and use CsF base for all hypotheses.
    \end{itemize}

    \item \textbf{Created knowledge triples}:
    \begin{itemize}
        \item \texttt{(Sulfone Electrophile, PerformsBestWith, CsF Base)}
        \item \texttt{(Sulfone Electrophile, PerformsBestWith, DMF Solvent)}
        \item \texttt{(Iodine Electrophile, PerformsBestWith, CsF Base)}
        \item \texttt{(Iodine Electrophile, PerformsBestWith, DMF Solvent)}
        \item \texttt{(Cyclohexyl Ligand, EnhancesYieldsWith, Sulfone Electrophile)}
        \item \texttt{(Biaryl Ligand, EnhancesYieldsWith, Sulfone Electrophile)}
        \item \texttt{(Biaryl Ligand, ShouldBeTestedWith, Acetone)}
    \end{itemize}
\end{itemize}
\end{tcolorbox}

%% file: prompts/custom_notes/reasoning_notes_response_schema.tex
\begin{tcolorbox}[
    enhanced,
    breakable,
    colback=teal!5!white,
    colframe=teal!75!black,
    title=Schema Definition for Reasoning Notes Response,
    left=2mm, right=2mm, top=2mm, bottom=2mm,
    fonttitle=\bfseries\small
]
\def\_{\textunderscore}
\texttt{class ReasoningNotesResponse(BaseModel)}:
\begin{itemize}
    \item[] \textit{Structured notes extracted from reasoning data}
    \item \textbf{notes}: \texttt{List[str]}
    \begin{itemize}
        \item \textbf{description}: \parbox[t]{0.8\linewidth}{May include summaries of findings, relationships, or principles.}
    \end{itemize}
    \item \textbf{key\_findings}: \texttt{List[str]}
    \begin{itemize}
        \item \textbf{description}: \parbox[t]{0.8\linewidth}{Important factual discoveries or observations drawn from the reasoning process.}
    \end{itemize}
    \item \textbf{parameter\_relationships}: \texttt{List[str]}
    \begin{itemize}
        \item \textbf{description}: \parbox[t]{0.8\linewidth}{Descriptive cause-effect or correlation relationships between different parameters or variables.}
    \end{itemize}
    \item \textbf{optimization\_principles}: \texttt{List[str]}
    \begin{itemize}
        \item \textbf{description}: \parbox[t]{0.8\linewidth}{Actionable principles or rules that can help optimize the experiment. Should be verifiable and clearly beneficial.}
    \end{itemize}
\end{itemize}
\end{tcolorbox}

%% file: prompts/custom_notes/compass_notes_response_schema.tex
\begin{tcolorbox}[
    enhanced,
    breakable,
    colback=teal!5!white,
    colframe=teal!75!black,
    title=Schema Definition for Compass Notes Response,
    left=2mm, right=2mm, top=2mm, bottom=2mm,
    fonttitle=\bfseries\small
]
\def\_{\textunderscore}
\texttt{class CompassNotesResponse(BaseModel)}:
\begin{itemize}
    \item[] \textit{General-purpose response format for structured scientific notes. Suitable for notes extracted from any experimental description or setup.}
    \item \textbf{notes}: \texttt{List[str]}
    \begin{itemize}
        \item \textbf{description}: \parbox[t]{0.8\linewidth}{Free-form bullet-point notes. General scientific facts, summaries, or highlights.}
    \end{itemize}
    \item \textbf{theoretical\_background}: \texttt{List[str]}
    \begin{itemize}
        \item \textbf{description}: \parbox[t]{0.8\linewidth}{Scientific principles, mechanisms, or theories underlying the experiment.}
    \end{itemize}
    \item \textbf{variable\_properties}: \texttt{List[str]}
    \begin{itemize}
        \item \textbf{description}: \parbox[t]{0.8\linewidth}{Descriptions of individual variables' attributes, roles, or behaviors.}
    \end{itemize}
    \item \textbf{variable\_relationships}: \texttt{List[str]}
    \begin{itemize}
        \item \textbf{description}: \parbox[t]{0.8\linewidth}{Cause-effect or correlated relationships between different variables or parameters.}
    \end{itemize}
\end{itemize}
\end{tcolorbox}